\documentclass{article}

\usepackage{arxiv}

\usepackage[utf8]{inputenc}
\usepackage[T1]{fontenc}

\usepackage{amsmath,amssymb}

\usepackage{graphicx}
\usepackage{subcaption}
\usepackage{booktabs}
\usepackage{multirow}

\usepackage{algorithm}
\usepackage{algpseudocode}

\usepackage[numbers,sort&compress]{natbib}

\usepackage{microtype}

\usepackage{xcolor}
\usepackage{hyperref}
\usepackage{doi}

\hypersetup{
    colorlinks=true,
    linkcolor=blue!50!black,
    citecolor=blue!50!black,
    urlcolor=blue!50!black,
    pdftitle={Interpretable MEG Decoding of Perceived Speech: Source-Space Patterns and the Stimulus Features That Drive Retrieval},
    pdfauthor={Ilia~Semenkov, Daria~Kleeva, Ivan~Dakhtin, Zarina~Maksudova, Alex~Ossadtchi},
    pdfkeywords={Magnetoencephalography, Speech perception, Neural decoding, Interpretable deep learning, Spatial--temporal decomposition, Source localization, Occlusion analysis, Naturalistic stimuli}
}

\newcommand{\E}{\mathbb{E}}
\DeclareMathOperator{\subcorr}{subcorr}
\newcommand{\Defossez}{D{\'e}fossez}

\title{Interpretable MEG Decoding of Perceived Speech: Source-Space Patterns and the Stimulus Features That Drive Retrieval}

\date{}

\author{ \href{https://orcid.org/0000-0003-1515-7062}{\includegraphics[scale=0.06]{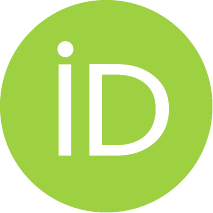}\hspace{1mm}Ilia~Semenkov}\\
	AXXX, Moscow, Russia\\
	HSE University, Moscow, Russia\\
	\texttt{iliasemenkov@gmail.com} \\
	\And
	\href{https://orcid.org/0000-0002-6040-2154}{\includegraphics[scale=0.06]{orcid.pdf}\hspace{1mm}Daria~Kleeva} \\
	HSE University, Moscow, Russia\\
	\\
	\And
	\href{https://orcid.org/0000-0003-1667-9006}{\includegraphics[scale=0.06]{orcid.pdf}\hspace{1mm}Ivan~Dakhtin} \\
	HSE University, Moscow, Russia\\
	\AND
	\href{https://orcid.org/0009-0004-8681-5848}{\includegraphics[scale=0.06]{orcid.pdf}\hspace{1mm}Zarina~Maksudova} \\
	ITMO University, St. Petersburg, Russia\\
	\And
	\href{https://orcid.org/0000-0001-8827-9429}{\includegraphics[scale=0.06]{orcid.pdf}\hspace{1mm}Alex~Ossadtchi} \\
	HSE University, Moscow, Russia\\
    AXXX, Moscow, Russia\\
	\texttt{ossadtchi@gmail.com} \\
}

\renewcommand{\headeright}{}
\renewcommand{\undertitle}{}
\renewcommand{\shorttitle}{Interpretable MEG Decoding of Perceived Speech}

\begin{document}

\maketitle

\begingroup
\renewcommand{\thefootnote}{}
\footnotetext{%
Project page:
\url{https://ivsemenkov.github.io/LISA/}

\par\smallskip
This is the author accepted manuscript of an article published in
\textit{NeuroImage}, 344 (2026), 122355.
The version of record is available at
\url{https://doi.org/10.1016/j.neuroimage.2026.122355}.

\par\smallskip
\textcopyright{} 2026.
This manuscript version is made available under the CC-BY-NC-ND 4.0 license (\url{https://creativecommons.org/licenses/by-nc-nd/4.0/}).
}
\endgroup

\begin{abstract}
Short segments of perceived speech can be retrieved from non-invasive
magnetoencephalographic (MEG) recordings by a deep network trained with a
CLIP-style objective against wav2vec 2.0 audio
embeddings~\cite{Defossez2023,Baevski2020,Radford2021}. Such studies do not convert their weights into the notions used in classical electrophysiology. Although the spatial filters in the front end of~\cite{Defossez2023} could in principle be mapped to source topographies, the dynamic properties of those sources remain out of reach. Which properties of the speech stream contribute to the decoding is equally unclear.

Here we scrutinize a decoder with a physiologically motivated,
factorized spatial--temporal front end that permits joint interpretation
of learned spatial topographies and temporal patterns in relation to
the MEG forward model. 
Building on the framework of Petrosyan et al.~\cite{Petrosyan2021,Petrosyan2022}, we replace the 2D Fourier spatial attention of~\cite{Defossez2023} with a layer parameterized by spherical harmonics~\cite{Sivakumar2016}, reduce the
subject-specific representation from 270 to $K=25$ branches, and add
trainable temporal filters so that each branch has an explicit learned spatial
filter and temporal response. Ocular and cardiac components are removed before training, since either could supply stimulus-locked information that 
would otherwise be mistaken for cortical.

On the cleaned MEG-MASC dataset~\cite{Gwilliams2023}  the model reaches $39.75\pm0.34\%$ Top-1
accuracy among 1005 candidates across six model initializations
(mean \(\pm\) SD), with
approximately $20\times$ fewer trainable parameters in the
decoder. Its learned spatial patterns can be mapped through a cortical
inverse model~\cite{Petrosyan2021,Haufe2014}, yielding approximate
source-magnitude projections with recurrent bilateral structure over lateral
cortex, concentrated around temporal and adjacent peri-Sylvian regions.
Paired MEG occlusion, replacing feature-marked speech segments
with matched donors from feature-present and feature-absent intervals, shows
significant positive effects for 15 of 19 tested stimulus features, the largest effects being silence,
sound intensity, vowels and acoustic onsets. Intriguingly, the randomly ordered word lists behave
oppositely: replacing word-list MEG with MEG recorded during coherent
narrative improves retrieval relative to word-list donor MEG. 
The wav2vec target can be reduced to about twelve learned feature dimensions
with only a small reduction in retrieval accuracy, 
whereas strong temporal compression causes
a clear performance loss.

Decoders with physiologically interpretable spatial--temporal front ends can thus serve as
knowledge-discovery tools: their learned weights can be converted into source-space and temporal patterns, while input interventions reveal what the
decision rule relies on.
\end{abstract}


\section{Introduction}
MEG signals contain sufficient information to discriminate and
identify perceived speech at several levels. Theta-band response patterns in
auditory cortex were shown to discriminate spoken sentences
\cite{LuoPoeppel2007}, and subsequent work identified previously unseen
fragments of continuous natural speech from listeners' MEG by reconstructing
speech-envelope-related activity~\cite{Koskinen2013SpeechFragments}.
Beyond stimulus identification, analyses of continuous speech have resolved
acoustic, phonetic, and lexical representations
\cite{DingSimon2012Continuous,BrodbeckPresaccoSimon2018,BrodbeckHongSimon2018},
while multivariate decoding has recovered phonetic features, phoneme identity
and serial position, and statistical properties of speech from naturalistic
MEG~\cite{Gwilliams2022}. More recently, deep neural decoders have substantially
expanded the scale and content of non-invasive speech decoding. Contrastive
models can retrieve the corresponding unseen speech segment from more than a
thousand candidates using only three seconds of MEG~\cite{Defossez2023}, and
large-scale models trained across multiple EEG and MEG datasets can decode
individual words, including words absent from the training
set~\cite{dAscoli2025}.
 
Accuracy alone, though, does not tell us what the network detected. These
models turn sensor signals into speech embeddings through a chain of
transformations, and the resulting weights match nothing an
electrophysiologist can name: not a cortical location, not a rhythm, not a
time course. The measurement and its interpretation therefore describe
different things, and no gain in accuracy will connect them. The concern
is not ours alone. Writing about their own earlier work, d'Ascoli et
al.~\cite{dAscoli2025} say that ``it is unclear whether the decoder relies
on the perceptual characteristics of the speech segment or the semantic
features of the underlying words''. 
Recent work has analyzed the performance in this retrieval setting~\cite{Zhang2026}, but it did not
identify which properties of the speech stream the decoder uses. That
question can be settled by intervening directly on the decoder's MEG input,
which requires no particular architecture.
Where in cortex that information arises, and with what dynamics, is a different question, and answering it does require a decoder whose weights can be read.

This matters because we increasingly ask decoding models to do science
rather than only to score well. Classical studies of speech perception
compare a handful of carefully chosen conditions and fit simple
statistical models to the differences between them. The control is
excellent, but the design suits continuous natural speech poorly. Machine
learning offers another route: set a hard decoding task on natural
recordings, then examine the rule the network learned. This reverses the
usual order of an experiment. Instead of deciding in advance which
stimulus features ought to matter and then testing them, we let the
network use whatever helps and afterward ask what it used.
 
The phenomenon of speech perception suits this approach well. 
MEG provides millisecond temporal resolution and whole-head coverage~\cite{Cohen1968,Baillet2017}.
Fifty years of invasive and non-invasive work have described the cortical networks involved~\cite{Penfield1959,Geschwind1970,Friederici2011}. 
And because speech production and perception share circuitry~\cite{Fadiga2002,Wilson2004,Liu2023,Eliades2008}, the applications are concrete --- speech neuroprostheses~\cite{Card2024,Petrosyan2022}, imagined-speech interfaces~\cite{Bisla2025}, and passive language mapping during surgery~\cite{Nourmohammadi2023,Taplin2016,Voskoboynikov_2025}.
 
Explainable AI has begun to reach neuroimaging. Recent work has audited
EEG foundation models with sparse autoencoders~\cite{LehnSchioler2024},
tracked how functionally specialized units emerge in networks trained on
brain signals~\cite{Hammer2022}, and used causal tracing to find the
pathways inside brain-to-speech models~\cite{Maghsoudi2025}. These methods
say useful things about what a network represents internally. What they do
not do is connect those representations to the physics that produced the
data --- the forward model that carries cortical currents to the sensors.
The studies that tackle speech or complex movement also work on
intracranial recordings~\cite{Hammer2022,Maghsoudi2025}, while the
non-invasive work targets broad clinical categories rather than
language~\cite{LehnSchioler2024}. Nobody has yet combined strong decoding
of complex natural behavior from non-invasive data with an interpretation that 
links learned patterns to cortical inverse-model projections and temporal structure. 
Without that link, the move from an accuracy figure to a neuroscientific claim stays risky: we get
high numbers and little idea what they mean.
 
A second line of work avoids the problem by building interpretability into
the architecture. Compact models such as EEGNet~\cite{Lawhern2016},
ShallowNet~\cite{Schirrmeister2017} and LF-CNN~\cite{Zubarev2019} separate
spatial from temporal filtering, which mirrors how the data arise:
band-limited activity in a neural population reaches the sensors through
the forward operator. Those models were built for simpler classification
and regression problems, and earlier attempts to interpret them missed one
point --- the spatial and temporal filters adapt at the same time, and
each changes what the other sees. Petrosyan et al.~\cite{Petrosyan2021}
corrected this and derived how to map jointly trained weights onto source
locations and their dynamics. The same front end later drove a decoder for
invasive ECoG and sEEG speech decoding~\cite{Petrosyan2022}. Its advantage
over post-hoc explanation is that the interpretation refers to the forward
model of the MEG signal, not to latent features and a heuristic for
ranking them. We extend it here to whole-head MEG from many subjects
during a demanding language task.
 
We start from D{\'e}fossez et al.~\cite{Defossez2023}, who paired a
spatial-attention layer with subject-specific spatial layers so that
spatial filters could be trained across participants, then passed the
result through convolutional layers to approximate wav2vec 2.0 audio
embeddings~\cite{Baevski2020} under a contrastive
objective~\cite{Radford2021}. We change four things. First, their spatial
attention uses 2D Fourier functions over a flattened projection of the
sensor array; we use real spherical harmonics
instead~\cite{Sivakumar2016}. 
MEG sensors occupy an approximately spherical arrangement,
so spherical harmonics provide a parameterization adapted to their
three-dimensional geometry.
Second, we add trainable depthwise temporal filters after the subject block,
giving each branch both a participant-specific spatial filter and a
learned temporal response. Third,
we cut the latent representation to $K = 25$ branches, which imposes a
small and physiologically plausible bottleneck. Finally, one change
concerns the data rather than the model: we remove ocular and cardiac
components before training. 
These artifacts are a plausible route to
shortcut learning here, because eye activity tracks linguistic structure and attended speech
even without corresponding visual input~\cite{Jin2018,Gehmacher2024,Braga2016}, while
cardiac dynamics vary with narrative intensity and conscious narrative
processing~\cite{Wallentin2011,Perez2021}. 
Left in the data, they
could supply the same stimulus-locked acoustic or linguistic information that our occlusion analysis later probes.
 
On MEG-MASC~\cite{Gwilliams2023} the model retrieves the correct
three-second segment among 1005 candidates with $39.8\%$ Top-1 and $70.4\%$
Top-10 accuracy, averaged over six independently trained solutions. The
complete MEG decoder contains 486,619 trainable parameters, approximately
$20\times$ fewer than the brain decoder of D{\'e}fossez et
al.~\cite{Defossez2023} (Appendix~\ref{app:model-size}).
These figures cannot be compared directly with those reported by
D{\'e}fossez et al.~\cite{Defossez2023}, whose test segments are aligned to
word onsets and whose reported preprocessing does not
explicitly remove ocular or cardiac components.
Within our own ablation grid, where the data and test set are held fixed,
the $K=270$, five-block configuration---the closest tested model in branch
count and decoder depth---is 14.8 times larger than the main model and
scores 3.60 percentage points lower in Top-1 accuracy
(Appendix~\ref{app:model-size}).

We first ask where in cortex the information supporting retrieval arises, by mapping the trained spatial and temporal weights to source space. We then ask which properties of the speech stream that information corresponds to, by intervening on the decoder's input
with paired MEG substitutions in which feature-marked intervals are replaced
by matched real-MEG donors. The Methods develop the interpretation framework
and the substitution procedure. The Results report 
the estimated cortical projections, the stimulus features the decoder relies on, 
and a set of compression and ablation experiments that bound both.

With this we hope to demonstrate that an architecture equipped with explicit physical and physiological priors loses no accuracy and can still be inspected once trained. 
Its weights yield source-space and temporal patterns, while paired input interventions reveal which stimulus-linked information the trained decoder
uses. Decoders built this way can serve as instruments for studying the
brain, not only as benchmarks for decoding it.

\section{Theoretical background}
\label{sec:background}

This section establishes the theoretical and methodological framework for modeling MEG data and identifying task-related neural sources. We first introduce the biophysical generative model of MEG signals and review classical spatial filtering techniques used to estimate source activity from sensor measurements. Following a discussion on the trade-offs between traditional approaches and modern deep learning methods, we briefly outline the mathematically grounded framework of Petrosyan
et al.~\cite{Petrosyan2021} for interpreting deep neural networks with
factorized spatial-temporal layers. By linking learned network weights back to the spatial topographies and temporal dynamics of the underlying neural populations, this methodology enables the extraction of physiologically meaningful features during complex decoding tasks, which we subsequently apply to the decoding of perceived speech.

\subsection{MEG generative equation}
\label{ssec:meg-generative}

MEG sensor measurements can be modeled as linear superpositions of
cortical current-source contributions. Synchronous currents in a cortical
patch are commonly approximated by an equivalent current dipole
(ECD)~\cite{Murakami2006,Baillet2017}, with location
\(\mathbf r_n\), orientation \(\boldsymbol{\theta}_n\), and time-varying
amplitude \(s_n(t)\). For an array of \(M\) sensors, the contribution of the
\(n\)-th source is \(\mathbf g_n s_n(t)\), where
\(\mathbf g_n=\mathbf g(\mathbf r_n,\boldsymbol{\theta}_n)\) is its
\(M\times1\) gain vector, or sensor topography. Gain vectors are obtained
from the electromagnetic forward model using the head, cortical, and sensor
geometry derived from MRI and MEG metadata with standard anatomical modeling
tools~\cite{Fischl2012}.

When multiple sources are active, linearity of Maxwell's equations gives
\begin{equation}
\mathbf{x}(t)
   = \sum_{n=1}^{N}\mathbf{g}_n s_n(t)+\sum_{u=1}^{N_{\mathrm{u}}}\mathbf{q}_u p_u(t) + \mathbf{n}(t),
\label{eq:meg-forward}
\end{equation}
 The first summation in (\ref{eq:meg-forward}) corresponds to the task-related sources whose activity via some potentially complex non-linear mapping predicts our behavioral variable $z(t)$. 
 The second sum describes the contribution of the $N_{\mathrm{u}}$ task-unrelated sources to the measured data $\mathbf{x}(t)$, where $p_u(t)$ denotes the time course of the $u$-th task-unrelated source and $\mathbf{q}_u$ its sensor topography.
 The latter can be regarded as the spatially-correlated brain noise term. The last element in this equation $\mathbf{n}(t)$ is the spatially-white sensor noise.  

A typical brain decoding task involves designing an algorithm that operates
on the sensor measurements $\mathbf{x}(t)$ to infer our target behavioral
variable $z(t)$. We assume $z(t)$ depends (possibly non-linearly) only
on the task-related activity, i.e.\ on a subset of source activations
$s_i(t),\,i=1,\ldots,N$ whose contributions are mediated through the
corresponding topographies $\mathbf{g}_i$. 
The contribution of these sources to $\mathbf{x}(t)$ is therefore confined to the subspace $\mathcal{S}=\mathrm{span}\{\mathbf{g}_1,\ldots,\mathbf{g}_N\}$ spanned by the topographies $\mathbf{g}_n$ of these task-related sources. 
Crucially, since $\mathcal{S}$ is in general not orthogonal to the brain-noise subspace
$\mathcal{I}=\mathrm{span}\{\mathbf{q}_1,\ldots,\mathbf{q}_{N_{\mathrm{u}}}\}$, an effective
decoder must not only tune to $\mathcal{S}$ but also tune away from
$\mathcal{I}$.

\subsection{From sensor signals to source activity}
\label{ssec:sensor-to-source}

The standard tool used in classical MEG analysis to access the activity of
neuronal sources $s_n(t)$ is a spatial filter $\mathbf{w}_k^\top$ applied to
the channel time series:
\begin{equation}
\hat{s}_k(t)
  = \mathbf{w}_k^\top\mathbf{x}(t)
  = \sum_{n=1}^{N}\mathbf{w}_k^\top\mathbf{g}_n s_n(t) 
   +\sum_{u=1}^{N_{\mathrm{u}}}
\mathbf{w}_k^\top\mathbf{q}_u p_u(t) + \mathbf{w}_k^\top\mathbf{n}(t).
\label{eq:beamformer}
\end{equation}

A spatial filter for source \(k\) should retain sensitivity to its
topography \(\mathbf g_k\) while suppressing other task-related and
task-unrelated topographies. Because multichannel MEG has finite rank and a
limited number of sensor-space degrees of freedom, these objectives cannot in
general be satisfied simultaneously, producing a trade-off between target
sensitivity and interference rejection.

For the linear minimum-mean-square-error estimator, the
orthogonality principle~\cite{Kay1993} gives the following relation between
the optimal spatial-filter matrix and the source topographies when additive
noise and task-unrelated activity are uncorrelated with the task-related
signals (Appendix~\ref{app:spatial-filter-background}):

\begin{equation}
\mathbf{W} = \mathbf{R}_{\mathbf{x}}^{-1}\mathbf{G}\mathbf{R}_{\mathbf{s}},
\label{eq:w_vs_g}
\end{equation}

\noindent where $\mathbf{R}_{\mathbf{x}} = \mathbf{E}\left\{\mathbf{x}(t)\mathbf{x}(t)^\top\right\}$ is the
$M \times M$ data covariance matrix and $\mathbf{R}_{\mathbf{s}}$ is the
$N \times N$ covariance matrix of the source signals $s_n(t)$, $n = 1,\ldots,N$.

In practice, the task-relevant source topographies are unknown.
Classical inverse modeling, task-driven spatial filtering
\cite{Nikulin2011,Dahne2014,ossadtchi2024representational}, and deep neural
decoding provide complementary trade-offs between anatomical explicitness,
task sensitivity, and expressive power. These considerations motivate a
task-driven but physiologically interpretable front end; a more detailed
comparison of the three approaches is provided in
Appendix~\ref{app:spatial-filter-background}.

\subsection{Interpretable neural architectures with factorized spatial-temporal layers}
\label{ssec:petrosyan-framework}
Several compact deep neural network architectures have been developed to decode 
EEG and MEG activity by employing factorized spatial and temporal filtering 
operations. Notable examples include EEGNet~\cite{Lawhern2016}, DeepConvNet and 
its compact variant ShallowConvNet~\cite{Schirrmeister2017}, the separable CNN 
proposed by Waytowich et al.~\cite{Waytowich2018}, as well as LF-CNN and 
VAR-CNN~\cite{Zubarev2019}. 

It is noteworthy that all these architectures contain not only the trainable spatial filters considered above but also the temporal filters that can be tuned to specific dynamical properties of the activity of neuronal sources. While these models successfully leverage separated spatial and temporal processing stages to perform automatic feature extraction with a minimal number of parameters, none of these earlier studies provided a rigorous interpretation of the learned weights. Specifically, the previous interpretation attempts overlooked the fact that in such 
factorized architectures, both the spatial and temporal filters adapt concurrently 
during the training procedure. 

\begin{figure}[t]
\centering
\includegraphics[width=\linewidth]{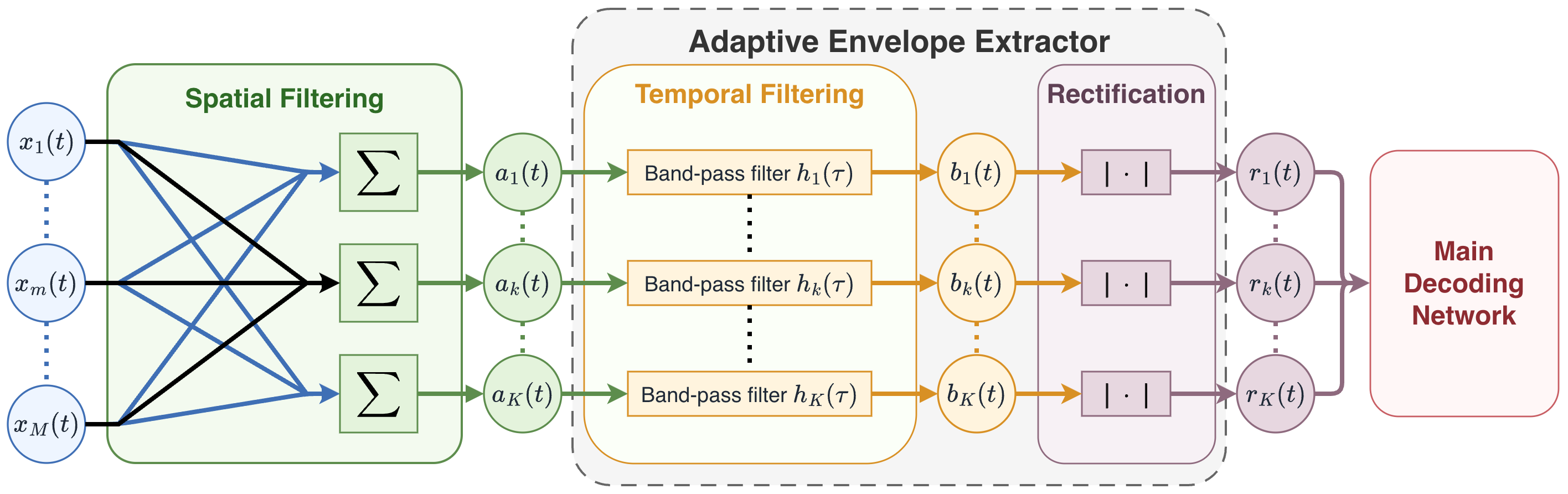}
\caption{
Schematic of the factorized spatio-temporal front end and branch-wise interpretation framework of Petrosyan et al.~\cite{Petrosyan2021}, with the $k$-th branch highlighted in blue. This schematic illustrates the reference architecture underlying the interpretation framework rather than the LISA architecture introduced in Section~\ref{sec:methods}.
}
\label{fig:collection_of_branches}
\end{figure}

To address this limitation, Petrosyan et al.~\cite{Petrosyan2021} introduced a
theoretically justified framework for interpreting jointly trained spatial and
temporal filters while accounting for their mutual dependence.
In this formulation, the front end is treated as a collection of
jointly learned spatial--temporal branches, see
Figure~\ref{fig:collection_of_branches}. The $k$-th branch performs the
following operation

\begin{equation}
r_k(t)
   = f\!\left(\mathbf{w}_k^\top \mathbf{x}(t)\ast h_k(t)\right),
\label{eq:branch}
\end{equation}
where $\ast$ denotes 1D temporal convolution, $h_k(t)$ is a $T$-tap impulse response of the $k$-th branch temporal filter and $f(\cdot)$ is an optional element-wise non-linearity. 
Equation~\eqref{eq:branch} describes the general branch formulation of Petrosyan et al.~\cite{Petrosyan2021}; in LISA, \(f\) is the identity, so no gating or pointwise non-linearity is applied after the temporal-filtering layer.

Applying equation \eqref{eq:w_vs_g} to the branch operation equation \eqref{eq:branch}, taking into account the fact that the spatial and temporal processing occur within a mutual context and stating that $\mathbf{R}_s$ is unknown in practice, we can arrive at the following recipe for deriving the corresponding spatial and temporal patterns from the learned filter weights. 

The branch-specific spatial pattern is computed as

\begin{equation}
\hat{\mathbf{g}}_k\;\propto\;\mathbf{R}_{h_k}\mathbf{w}_k,
\label{eq:haufe-temporal}
\end{equation}
where
$\mathbf{R}_{h_k}=\E\{(\mathbf{x}(t).\!\ast h_k(\tau))(\mathbf{x}(t).\!\ast h_k(\tau))^\top\}$
is the covariance of the multichannel data filtered with the temporal filter
of the $k$-th branch and $.\!\ast$ denotes per-channel convolution. The
intuition here is that the spatial filter, while tuning away from the interfering sources, takes into account the extent to which the contribution of these interfering sources is eliminated by the temporal filter operating within the same branch and therefore uses the accordingly filtered  multichannel data to compute the spatial covariance matrix. 

To map the branch-specific topographies $\hat{\mathbf{g}}_k$ onto the
cortex, one can choose from a variety of inverse-modeling
approaches~\cite{hecker2026invertmeeg}. In this work, we used the Minimum
Norm Estimate (MNE)~\cite{Hamalainen1994}. In its basic form, the
corresponding linear inverse operator can be written as

\begin{equation}
\mathbf{W}_{\mathrm{MNE}}
=
\mathbf{G}_{\mathrm{all}}^\top
\left(
\mathbf{G}_{\mathrm{all}}\mathbf{G}_{\mathrm{all}}^\top
+\lambda^2\mathbf{I}
\right)^{-1},
\label{eq:mne_all}
\end{equation}

where $\mathbf{G}_{\mathrm{all}}\in\mathbb{R}^{M\times P}$ contains the
sensor topographies of all source components included in the forward
model, $P$ depends on the cortical source space and orientation model,
$\lambda^2$ is the regularization coefficient, and $\mathbf{I}$ is the $M\times M$ identity matrix.
The specific noise covariance, depth weighting, orientation constraint,
and regularization used for the reported cortical maps are described in
Section~\ref{ssec:weight-interpretation}.

Analysis of each branch also yields a temporal pattern describing the
second-order temporal structure selected by that branch. Let
$a_k(t)=\mathbf{w}_k^\top\mathbf{x}(t)+b_k$ be the spatially filtered branch
signal and let
$\mathbf{a}_k(t)=[a_k(t),a_k(t-1),\ldots,a_k(t-T+1)]^\top$ denote its
lag-embedded representation. Let $\mathbf h_k=[h_k(0),\ldots,h_k(T-1)]^\top$ denote the vector of the $T$ learned temporal-filter coefficients, ordered consistently with this lag vector. The temporal pattern is then computed as
\begin{equation}
\hat{\mathbf p}_k\;\propto\;\mathbf R_{a_k}\mathbf h_k,
\qquad
\mathbf R_{a_k}=\operatorname{Cov}\!\left[\mathbf a_k(t)\right].
\label{eq:temporal-pattern}
\end{equation}
In the reported analyses, $\mathbf R_{a_k}$ was estimated with Ledoit--Wolf
shrinkage and $T=15$. The magnitude spectrum
$|\mathcal F\{\hat{\mathbf p}_k\}|$ was used as a descriptive frequency-domain
representation of this temporal pattern. Thus, each branch is characterized
by a spatial pattern $\hat{\mathbf g}_k$ and a temporal pattern
$\hat{\mathbf p}_k$ derived while accounting for the covariance structure of
the corresponding filtered signals.

The number of branches \(K\) sets the dimensionality of the learned
spatial--temporal representation passed from the interpretable front end to
the downstream decoder. Sweeping \(K\) therefore probes how many separately
parameterized branch components are needed to preserve retrieval-relevant
information. A performance plateau provides a practical upper bound on this
representational dimensionality for the present task.

Historically, the majority of interpretable, compact deep neural network architectures utilizing factorized spatial-temporal filtering---such as those proposed by Lawhern et al. \cite{Lawhern2016}, DeepConvNet \cite{Schirrmeister2017}, separable CNNs \cite{Waytowich2018}, and LF-CNN/VAR-CNN \cite{Zubarev2019}---have been applied to relatively simple classification or regression tasks. These models typically lacked sophisticated decoder heads beyond their factorized blocks; at most, architectures like DeepConvNet employed a simple succession of convolutional layers. However, as demonstrated by Petrosyan et al.~\cite{Petrosyan2022}, such interpretable front-ends can be successfully adapted for highly complex tasks like overt speech decoding. In their approach, a recurrent neural network processed the outputs from the spatial-temporal filtering block to predict speech mel-spectrograms from ongoing brain activity immediately preceding an utterance. Subsequent localization of the pivotal neural sources---derived from the individual branches of the network's front-end---successfully identified language-related cortical sites 
whose locations showed correspondence with regions where direct electrical
stimulation induced speech arrest and production errors, supporting
the physiological relevance of the extracted features. Furthermore, the analysis of temporal patterns revealed a relationship between compact spatial patterns and high central frequencies. 
Together with the correspondence to stimulation-identified
language sites, this spatial--spectral relationship
supports neural activity as the primary source of the
information driving the decoder, rather than muscular artifacts.

In the current work we apply the methodology of Petrosyan et al.~\cite{Petrosyan2021} to the task of decoding perceived speech from the concurrently recorded MEG data.

\section{Methods}
\label{sec:methods}

\begin{figure}[t]
\centering
\includegraphics[width=\linewidth]{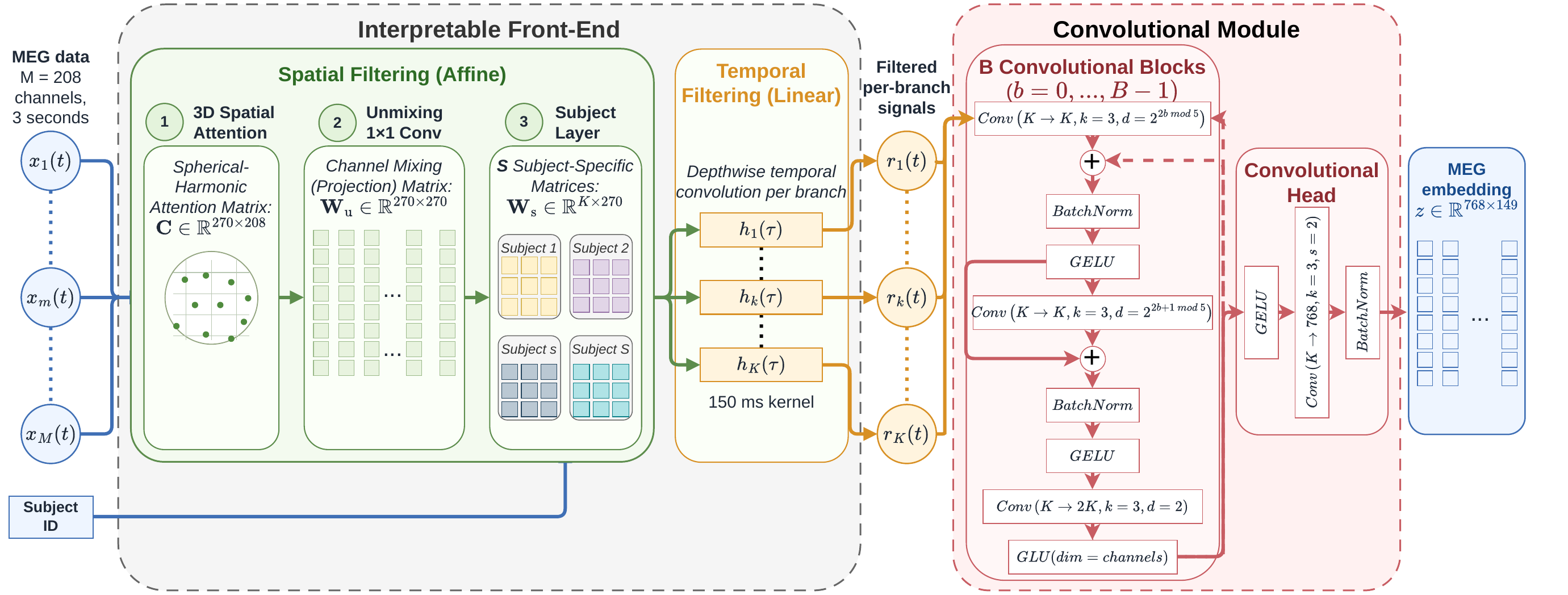}
\caption{
Our network's architecture for MEG-to-audio embedding alignment.
A 3-second, 208-channel MEG segment is processed by an interpretable front end: spherical-harmonic 3D spatial attention maps the sensor signals to 270 geometry-parameterized virtual channels; a $1 \times 1$ unmixing convolution applies a learned affine transformation in this channel space; the subject-specific layer then projects the representation to $K$ interpretable branches selected by the subject ID.
Each branch is passed through a depthwise temporal convolution with a 150 ms kernel, producing filtered branch-wise signals.
These signals are processed by a convolutional module with $B$ residual convolutional blocks, where we evaluate $B \in \{0,\ldots,5\}$ and use $B=2$ in the main architecture, followed by a convolutional head that outputs the MEG embedding aligned with the wav2vec audio embedding.
}
\label{fig:lisa_architecture}
\end{figure}

As stated earlier, in this work we solve the retrieval task in which the network learns to construct
embeddings for MEG data similar to the audio embeddings produced by
wav2vec 2.0~\cite{Baevski2020}, with a CLIP-style
objective~\cite{Radford2021}, following~\cite{Defossez2023}. Our architecture, see Figure \ref{fig:lisa_architecture}
inherits the overall block structure of~\cite{Defossez2023} but replaces
its spatial-attention layer with a geometry-adapted spherical-harmonic parameterization, augments it with a temporal-filtering layer in the spirit of
\cite{Petrosyan2021, Petrosyan2022} and modifies a convolutional decoder.
We refer to this architecture as LISA (Linear, Interpretable, Slim and Aware). ``Linear'' refers to the explicitly linear spatial--temporal front end used for physiological interpretation; ``Interpretable'' to the recovery of branch-specific spatial and temporal patterns; ``Slim'' to the compact parameterization of the decoder; and ``Aware'' to its explicit awareness of sensor geometry and participant identity through spherical-harmonic spatial attention and participant-specific projections.
The decoding setting is participant-dependent: test participants are
not held out by identity, and each MEG segment is processed using the
participant-specific spatial projection learned for that participant during
training. The present study therefore does not evaluate decoding in previously
unseen participants.

\subsection{3D spatial attention layer}
\label{ssec:3d-attention}

\Defossez{} et al.~\cite{Defossez2023} parameterized the spatial-attention
layer through 2-D Fourier functions defined over a planar projection of the
sensor layout. Because the MEG sensors occupy a three-dimensional,
approximately spherical arrangement, we instead parameterize each of the
$J=270$ virtual channels by real spherical harmonics. For virtual channel
$j$ and sensor $m$, the unnormalized coefficient is
\begin{equation}
 c_{jm}
 = \sum_{\ell=0}^{L-1}\sum_{q=-\ell}^{\ell}
   \gamma_{j}^{q,\ell}Y_{\ell}^{q}(\theta_m,\varphi_m),
\label{eq:3d-attention}
\end{equation}
where $(\theta_m,\varphi_m)$ are the polar and azimuthal angles of sensor
$m$, $Y_{\ell}^{q}$ is a real spherical-harmonic basis function, and
$\gamma_{j}^{q,\ell}$ is learned. We set $L=24$, so the implementation uses
degrees $\ell=0,\ldots,23$ and therefore $L^2=576$ basis functions for each
virtual channel.

Equivalently, let
$\mathbf B\in\mathbb R^{M\times L^2}$ contain the spherical-harmonic basis
functions evaluated at the $M$ sensor locations and let
$\boldsymbol{\gamma}_j\in\mathbb R^{L^2}$ collect the coefficients of virtual
channel $j$. Equation~\eqref{eq:3d-attention} can then be written as
$\mathbf c_j=\mathbf B\boldsymbol{\gamma}_j$. For the present
$M=208$, $L=24$ sensor layout,
$\mathbf B\in\mathbb R^{208\times576}$ has numerical rank 208 and spectral
condition number $\kappa_2=1.81$. The harmonic representation is therefore
overcomplete at the sampled sensor locations but spans the complete
208-dimensional sensor space at this order.
Thus, spherical harmonics provide a geometry-adapted
parameterization of the shared attention stage, not a physiological
restriction on the represented cortical sources. Physiological
interpretation instead uses the complete effective spatial filters
after the attention, unmixing, and participant-specific projections,
together with the learned branch-wise temporal patterns.

The coefficients are normalized across the $M$ sensors separately for each
virtual channel,
$\widetilde{\mathbf c}_j=\operatorname{softmax}(\mathbf c_j)$, and applied as
\begin{equation}
 \mathrm{SA}_j(\mathbf{x}(t))
 = \widetilde{\mathbf c}_j^{\top}\mathbf{x}(t),
 \qquad j=1,\ldots,J.
\label{eq:sa-output}
\end{equation}

\subsection{Interpretable front-end}
\label{ssec:temporal-filtering}

The interpretable subject block follows the factorized spatial--temporal
structure introduced in Eq.~\eqref{eq:branch}.
Parameter $K$ controls the number of branches in the block.
Each branch defines a learned spatial--temporal component through its
effective participant-specific spatial filter and branch-wise temporal filter;
these components form the units of the subsequent pattern and source-space
analyses.
For participant $s$, let
$\mathbf C\in\mathbb R^{270\times208}$ denote the softmax-normalized
3-D attention matrix whose $j$-th row is
$\widetilde{\mathbf c}_j^\top$ from Eq.~\eqref{eq:sa-output},
$\mathbf W_u\in\mathbb R^{270\times270}$ and
$\mathbf b_u\in\mathbb R^{270}$ the weights and bias of the shared
$1\times1$ unmixing convolution, and
$\mathbf W_s\in\mathbb R^{K\times270}$ the participant-specific
projection. The effective participant-specific spatial filtering matrix is
\begin{equation}
 \mathbf W^{(s)}
 = \mathbf W_s\mathbf W_u\mathbf C
 \in\mathbb R^{K\times208},
 \label{eq:w-whole}
\end{equation}
and the corresponding branch-wise bias is
\begin{equation}
 \mathbf b^{(s)} = \mathbf W_s\mathbf b_u
 \in\mathbb R^K.
 \label{eq:branch-bias}
\end{equation}
The branch signals before temporal filtering are therefore
\begin{equation}
 \mathbf a_s(t)
 = \mathbf W^{(s)}\mathbf x_s(t) + \mathbf b^{(s)}.
 \label{eq:branch-spatial-output}
\end{equation}
The $k$-th row $\mathbf w_{s,k}^{\top}$ of $\mathbf W^{(s)}$
defines the spatial filter used by branch $k$, while $b^{(s)}_k$
is its learned offset.

Brain rhythms are key components of non-invasively measured electrical neuronal
activity, reflecting different aspects of how the cortex processes incoming
information~\cite{Buzsaki2006}.
To capture branch-specific temporal structure, we applied one trainable
1-D depthwise temporal filter to each of the $K$ branch signals.
Each filter had 15 samples, corresponding to 150\,ms at the MEG sampling
rate of $f_s=100$\,Hz. The layer used same-length zero padding,
$K$ groups, and no bias term. The temporal filters were shared across
participants, whereas the preceding spatial projection was
participant-specific. No gating or pointwise non-linearity was applied
after this layer. Therefore, the output of branch $k$ is then obtained by applying its temporal
filter to $a_{s,k}(t)$:
\begin{equation}
 r_{s,k}(t) = \bigl(a_{s,k} * h_k\bigr)(t).
 \label{eq:implemented-branch}
\end{equation}

\subsection{Non-linear decoder}
\label{ssec:nonlinear-decoder}

The $K$ branch-wise signals produced by the interpretable front end were
passed to a non-linear temporal decoder adapted from the convolutional brain
decoder of Défossez et al.~\cite{Defossez2023}. The decoder comprised $B$
temporal convolutional blocks followed by a convolutional head. We evaluated
$B\in\{0,\ldots,5\}$; when $B=0$, the branch signals were passed directly to
the convolutional head.

Each temporal block contained three one-dimensional convolutions with kernel
size 3 and stride 1. For block $b=0,\ldots,B-1$, the dilation factors of the
first two convolutions were
\begin{equation}
d_{b,1}=2^{(2b\bmod 5)},
\qquad
d_{b,2}=2^{((2b+1)\bmod 5)}.
\end{equation}
Both convolutions preserved the temporal length through padding and were
followed by batch normalization and a GELU activation. The second convolution
used a residual connection in every block, while the first convolution used
a residual connection in all blocks except the first. The third convolution
used kernel size 3, stride 1, and dilation 2, and increased the channel
dimension from $K$ to $2K$. A gated linear unit then reduced it back to $K$
channels.

The convolutional head first applied a GELU activation and then projected the
$K$ decoder channels to the $F=768$ wav2vec feature channels using a
one-dimensional convolution with kernel size 3, stride 2, and no padding,
followed by batch normalization. Consequently, an input containing $T$ MEG
samples produced
\begin{equation}
T'=\left\lfloor\frac{T-3}{2}\right\rfloor+1
\end{equation}
output time points. For the 3\,s inputs sampled at 100\,Hz used in the main
experiment, this transformed the $K\times300$ branch representation into a
$768\times149$ MEG-derived embedding. The main model used $K=25$ branches and
$B=2$ temporal convolutional blocks.

\subsection{Spatial weights interpretation and source localization}
\label{ssec:weight-interpretation}

For participant $s$, the $k$-th row
$\mathbf w_{s,k}^{\top}$ of the effective spatial-filter matrix
$\mathbf W^{(s)}$ defined in Eq.~\eqref{eq:w-whole} is the spatial
filter of branch $k$. We compute the corresponding spatial pattern
$\hat{\mathbf g}_{s,k}$ using Eq.~\eqref{eq:haufe-temporal}.

All interpretation analyses used the temporal kernels \(h_k\) learned by the
trained decoder directly. Before pattern estimation, the channel-wise training
scaling was analytically inverted. Writing the combined robust and standard
scaling for channel \(m\) as
\(z_m=(x_m-o_m)/s_m\), where \(x_m\) is the baseline-corrected MEG signal in
physical units, the stored MEG was restored as \(x_m=s_m z_m+o_m\).
The effective spatial affine transformation was converted consistently:
\(w^{\mathrm{phys}}_{k,m}=w^{z}_{k,m}/s_m\) and
\(b^{\mathrm{phys}}_k=b^{z}_k-\sum_m w^{\mathrm{phys}}_{k,m}o_m\).
Thus, the branch activity represented by the transformed filter is identical
to that of the trained model, while the sensor amplitudes used for pattern
estimation are expressed in physical MEG units. The covariance matrix in
Eq.~\eqref{eq:haufe-temporal} was estimated with Ledoit--Wolf shrinkage after
filtering the physical-unit MEG with the corresponding learned kernel \(h_k\).

For cortical localization, we applied the MNE inverse solution using MNE-Python~\cite{Gramfort2013} to the branch-specific sensor topographies. For each participant and recording, KIT device-to-head geometry was reconstructed from the corresponding digitization and marker files, a participant-specific head-to-\texttt{fsaverage} transform was fitted, and a forward solution was constructed on the common \texttt{fsaverage} cortical anatomy using an \texttt{ico4} source space and boundary-element head model. We used an ad hoc sensor-noise covariance, a loose orientation constraint of 0.5, depth weighting of 0.5, and a regularization parameter of \(\lambda^2=1/3\), following the MNE-Python \texttt{lambda2} convention. 
For each branch, we retained the absolute source amplitudes and normalized the cortical map to unit \(L_2\) norm before visualisation and clustering. Thus, participants share a common cortical mesh for comparison, while their sensor-to-head geometry and forward solutions remain participant-specific.
We additionally assessed the sensitivity of this template-anatomy localization
using the subset of participants with usable individual structural MRI
(Appendix~\ref{app:individual-mri-validation}).

\subsection{Clustering branch patterns across participants}
\label{ssec:clustering}

For the participant--branch analysis, we computed the spatial and temporal patterns described in Section~\ref{ssec:petrosyan-framework} for each of the \(K=25\) branches and all 27 participants, using the \emph{Cable Spool Fort} recordings from the first session. Sensor-space spatial patterns were computed after restoring the MEG and effective spatial weights to physical units and were projected using each participant's forward model onto the common \texttt{fsaverage} cortical surface; absolute source amplitudes were then normalized to unit \(L_2\) norm. The temporal representation was the 15-sample pattern of Eq.~\eqref{eq:temporal-pattern}. Its FFT magnitude was used for visualisation, whereas its log-magnitude spectrum was used to compare branches. This produced \(27\times25=675\) participant--branch items, each represented by a cortical source-magnitude map and a temporal spectrum.

For items $i$ and $j$, we computed the Pearson correlation
$r^{\mathrm{src}}_{ij}$ between their cortical source-magnitude maps and the
Pearson correlation $r^{\mathrm{temp}}_{ij}$ between their log-magnitude
spectra. We combined the two views through the distance
\begin{equation}
d_{ij}
=
1-
\min\left\{
\frac{r^{\mathrm{src}}_{ij}+1}{2},
\frac{r^{\mathrm{temp}}_{ij}+1}{2}
\right\}.
\label{eq:branch-clustering-distance}
\end{equation}
This weakest-view combination makes two branches close only when both their
cortical distributions and their temporal frequency profiles are similar.

Before clustering, we screened the sensor-space spatial patterns for excessive
local roughness. Roughness was defined as the mean squared difference between
each sensor value and the values at its six nearest sensors, divided by the
variance of the pattern across sensors. Patterns with roughness greater than
$1$ were excluded from clustering.

We applied complete-linkage agglomerative clustering to all remaining
participant--branch items and cut the hierarchy at a minimum raw Pearson
correlation of $\tau=0.25$ in both views, corresponding to a distance threshold
of $(1-\tau)/2=0.375$ in
Eq.~\eqref{eq:branch-clustering-distance}. Taking the minimum across views makes
the pairwise criterion strict: two branches can be close only if both their
source maps and temporal spectra are similar. Complete linkage then makes the
cluster-level criterion strict by enforcing this bound for the least similar
pair in a cluster. Because a single low-similarity pair can therefore block a
merge, we used $\tau=0.25$ as a deliberately permissive worst-case bound rather
than as the expected within-cluster similarity. Consequently, every pair
within a resulting cluster had both raw correlations of at least $0.25$.

Each cluster was represented by its medoid: the observed
participant--branch item with the highest mean joint spatial--temporal
similarity to the other members of that cluster. We used an observed medoid
rather than a pointwise cluster average because averaging spatially displaced
topographies or source maxima can blur their structure and produce a synthetic
pattern that is less representative of the actual cluster members.

\subsection{Paired MEG occlusion analysis of stimulus-feature use}
\label{ssec:occlusion-features}

To determine which stimulus-linked neural information supports the trained
decoder's retrieval decisions, we tested a battery of candidate features
spanning broad acoustic state, rapid spectral change, phonetic identity,
lexical timing, contextual predictability, and experimentally inserted
disruptions of the narrative. The analysis contrasts matched MEG substitutions
associated with the presence and absence of each feature. We therefore first
define the candidate features and their control states before describing the
substitution procedure.

\paragraph{Stimulus features.}
We evaluated 19 acoustic, phonetic, and linguistic features on the same
100\,Hz audio timeline used to construct the MEG--audio pairs. Four binary
features described complete silent periods, silence onsets corresponding to
speech cessation, silence offsets corresponding to speech resumption, and the
union of the two boundary masks. Seven further features marked vowels, stops,
fricatives, sibilants, nasals, liquids/glides, and schwa using Montreal Forced
Aligner TextGrid annotations. These phoneme groups were not mutually exclusive:
sibilants were also included among fricatives, and schwa among vowels.

A separate word-onset feature marked the beginning of every spoken word,
including words occurring within continuous speech rather than only after
silence. Two features captured the dataset's controlled departures from
natural narrative speech. Pseudowords were non-words inserted within
otherwise coherent sentences, providing spoken sequences without an
associated lexical meaning. 
Random word lists consisted of words taken from the preceding approximately
five minutes of the story and presented in random order. This preserved
individual words while removing normal sentence and narrative structure. 
The
corresponding masks covered the full spoken duration of each pseudoword or
word-list segment.

The continuous linguistic features were word surprisal, predictive entropy,
and word frequency. Surprisal was computed with GPT-2 as the summed negative
log$_2$ probability of the subword tokens forming each word. Predictive entropy
was the entropy of the model's next-token distribution immediately before the
word, and frequency was measured on the Zipf scale. The remaining continuous
features described basic acoustic structure. Loudness was an energy-based proxy
computed as the log mean-square waveform energy in non-overlapping 10\,ms
frames. Acoustic-onset strength was a spectral-flux envelope computed from a
centred log-power mel spectrogram using a 128\,ms analysis window and a 10\,ms
hop, and therefore measured rapid changes in the stimulus spectrum.

\paragraph{Feature states and controls.}
For the binary features, the annotated intervals constituted the
feature-present state. The continuous features were converted into binary
contrasts. For surprisal, predictive entropy, loudness, and acoustic-onset
strength, the upper and lower quartiles defined the feature-present and
feature-absent states, respectively. 
Accordingly, the upper-quartile loudness and acoustic-onset states are labelled ``high loudness'' and ``strong acoustic onset'', respectively, in Figure~\ref{fig:occlusion-main}.
For word frequency, the direction was
reversed so that rare and frequent words formed the feature-present and
feature-absent states. Quartile thresholds for the word-level features were estimated from ordinary
narrative words in the five non-test pieces of \emph{The Black Willow};
pseudowords and words presented in the random word-list segments were excluded
from threshold estimation.

The feature-absent state was chosen to provide a meaningful control rather
than a generic complement whenever possible. For each phoneme class, the
control comprised other labelled phonemes. For pseudoword and random word-list
insertions, it comprised unmanipulated narrative words. For the remaining
binary features, it comprised time outside the marked feature intervals.
Feature-present masks were expanded by 20\,ms on either side to accommodate
small timing errors. When a predefined control mask was available, it was
expanded likewise and any resulting overlap with the feature-present mask was
removed. Otherwise, the feature-absent donor pool was defined as all time
outside the expanded feature-present mask.

\paragraph{Paired MEG substitution.}
The 1005 test queries were the 3\,s windows from pieces 5--11 of
\emph{The Black Willow}. Real-MEG donor intervals were drawn from pieces
0--4 of the same story. For every participant and session, we
selected test windows that contained the feature but were not completely
covered by it. All feature-positive intervals within an eligible query were
replaced simultaneously in two conditions:
\begin{align}
X^{f\rightarrow 0} &: \text{replacement by MEG from a feature-absent interval},\\
X^{f\rightarrow f} &: \text{replacement by MEG from a feature-present interval}.
\end{align}
The feature-present substitution served as a matched control for the
replacement procedure itself. Comparing $X^{f\rightarrow 0}$ directly with
the unmodified query would conflate the effect of changing the feature state
with generic consequences of replacing the original MEG, including loss of
query-specific activity, donor mismatch, and blending at the replacement
boundaries. In contrast, both substitution arms replaced the same intervals
with real donor MEG under the same matching and tapering procedure, while
differing in whether the donor interval was associated with the feature.
The unmodified-query ranks were retained as a descriptive baseline for
quantifying the total retrieval damage produced by each substitution arm,
but were not used as the primary feature-specific contrast.

Donor intervals had exactly the same duration as the replaced interval and
came from the same participant and session as the target query. Within each
matched pair, the feature-present and feature-absent donors came from the
same audio file; the absent interval was selected as the nearest available
match to a randomly sampled present interval. 
No additional matching was imposed on lexical, phonetic, or acoustic
covariates; the resulting contrasts therefore retain naturally occurring
correlations among stimulus properties.
All 208 channels were copied
together, and a 20\,ms raised-cosine taper blended each replacement into the
original signal. Because multiple valid donor intervals were available, an
effect estimated from a single donor pair could depend substantially on that
particular random selection. We therefore evaluated each eligible query using
five matched donor pairs and averaged over these repetitions in the rank
contrast below, reducing variability attributable to donor selection. These
repetitions were used for averaging and were not treated as independent
observations. 
The same fixed 150\,ms MEG--audio offset used during
training and testing was retained for donor extraction.
This offset defines the MEG--audio alignment and is not an
estimate of feature-specific neural response latency.

Importantly, windows without the feature, windows fully covered by the feature, and
windows for which a complete set of matched donor pairs was unavailable were
excluded only from the feature-specific query average. They remained in the
unchanged 1005-candidate retrieval bank for every rank calculation.

\paragraph{Rank contrast.}
For eligible query $i$ and donor pair $j\in\{1,\ldots,5\}$, let
$r^{f\rightarrow 0}_{sqij}$ and $r^{f\rightarrow f}_{sqij}$ denote the rank of the correct audio candidate for participant $s$ and
session $q$. Rank was defined as one plus the number of candidates with a
cosine similarity strictly greater than that of the correct candidate. The
subject--session effect for feature $f$ was
\begin{equation}
E_{sqf}
=
\frac{1}{5|\mathcal I_f|}
\sum_{i\in\mathcal I_f}\sum_{j=1}^{5}
\left(
 r^{f\rightarrow 0}_{sqij}
 -
 r^{f\rightarrow f}_{sqij}
\right),
\label{eq:occlusion-effect-session}
\end{equation}
where $\mathcal I_f$ is the set of eligible test queries.
Because both the number of eligible queries and the total duration
of replaced MEG differ across features, \(D_{sf}\) is a within-feature
contrast between its matched substitution conditions; its magnitude is not
a calibrated effect-size scale for comparisons between different features.
Positive values indicate that replacing the marked intervals with feature-absent MEG worsened
retrieval more than the matched feature-present control. For participants
with two sessions, the two session effects were averaged so that every
participant contributed one value,
\begin{equation}
D_{sf}=\frac{1}{Q_s}\sum_{q=1}^{Q_s}E_{sqf}.
\end{equation}

\paragraph{Statistical inference.}
For each feature, the observed statistic was the studentized participant mean,
\begin{equation}
T_f=
\frac{\overline D_f}
{\operatorname{SD}(D_{1f},\ldots,D_{Sf})/\sqrt{S}},
\qquad S=27.
\end{equation}
We tested the two-sided alternative $\mathbb{E}[D_{sf}]\neq0$ using
$N_{\mathrm{perm}}=100{,}000$ participant sign-flip permutations. In each
permutation, one random sign was applied to the complete 19-feature vector of
each participant, thereby preserving dependencies among features within
participants. Familywise error was controlled with a two-sided single-step
max-$|T|$ correction. If $T_f^{(b)}$ is the permuted statistic and
$M^{(b)}=\max_g \left|T_g^{(b)}\right|$, the corrected value was

\begin{equation}
p_f^{\mathrm{FWER}}
=
\frac{
1+\sum_{b=1}^{N_{\mathrm{perm}}}
\mathbf{1}\!\left[M^{(b)}\geq \left|T_f\right|\right]
}{N_{\mathrm{perm}}+1}.
\end{equation}

\paragraph{Feature-present control and significance criterion.}
A difference between the two substitution conditions is interpretable only if
the feature-present substitution preserves meaningful retrieval. If replacing
the marked intervals with feature-present MEG already reduces retrieval to
random ranking, the contrast with the feature-absent substitution may reflect
different degrees of generic disruption rather than the decoder's use of the
tested feature. We therefore used the $f\rightarrow f$ condition as a validity
check for each feature.

After averaging donor pairs and query windows within each
participant--session and then sessions within each participant, we computed
the mean $f\rightarrow f$ rank across the 27 participants. Under random
ordering of 1005 candidates, the correct target is equally likely to occupy
any rank from 1 to 1005, so the expected mean rank is
\begin{equation}
\mathbb{E}[r_{\mathrm{random}}]
=
\frac{1+1005}{2}
=
503.
\end{equation}
The feature-present control was considered valid when its mean rank was below
503. This check was performed once per feature and was not used to exclude
individual query windows.

Therefore, a feature was marked as showing a significant positive occlusion effect only
when all three criteria were met: the mean participant-level contrast
$\overline{D}_f$ was positive, the two-sided participant sign-flip max-$|T|$
test yielded $p_f^{\mathrm{FWER}}<0.05$, and the mean rank in the
$f\rightarrow f$ condition was below 503.
Significant negative effects were defined analogously, with
$\overline{D}_f<0$.

\subsection{Dataset and preprocessing}
\label{ssec:dataset_prep}

We use the MEG-MASC dataset~\cite{Gwilliams2023}, comprising audio and MEG
signals from 49 approximately one-hour participant--session recordings
obtained from 27 English-speaking participants. Twenty-two participants
contributed two sessions and five contributed one. In each available session,
participants listened to four fictional stories from the MASC corpus. The
study was approved by the institutional review board ethics committee of New
York University Abu Dhabi~\cite{Gwilliams2023}. 

Audio was resampled to 16\,kHz and divided into 3\,s windows with a
1\,s stride. We discarded windows whose peak absolute waveform amplitude was below
\(10^{-4}\). A window was retained only if at least 50\% of the duration of
one or more annotated words fell within it. The target representation for
each retained window was obtained with the wav2vec 2.0 Base
model~\cite{Baevski2020} by averaging its final four hidden layers at each
model time step.

ICA was computed separately for each of the 196 recordings using
MNE-Python FastICA. The decomposition was fitted to the 208 magnetometers of
the complete recording after 40\,Hz low-pass filtering, with 60 components
estimated after PCA, and the resulting decomposition was applied to the
original 1000\,Hz data. Components were identified manually by inspecting
their sensor topography, spectrum, time course, and epoch image; components
corresponding to eye movements or blinks, cardiac activity, or broad
far-field environmental magnetic interference were marked for exclusion.
No automatic EOG- or ECG-based criteria were used because the MEG-MASC
recordings contain no EOG or ECG channels. A median of 5 components per
recording was removed (range 2--8).

The data were resampled from 1000 to
100\,Hz. Separately for each participant--session--story recording, we
subtracted the per-channel mean over the 0.5\,s immediately preceding the
first stimulus onset, robust-scaled each channel using its median and
interquartile range, standardized it to zero mean and unit variance, and
clipped the resulting values to \(\pm20\) standard deviations. 
Across all 196 ICA-cleaned recordings, clipping affected
408,775 of 4,288,793,600 channel--time samples (0.0095\%).
For the 27 first-session \emph{Cable Spool Fort} recordings used
to estimate the principal spatial--temporal patterns
(Section~\ref{ssec:clustering}), only 1,507 of 427,419,200 samples
(0.00035\%) were clipped, with a maximum of 0.0032\% within any
individual recording. Even the most affected sensor in these recordings had only 0.10\%
of its samples clipped.

Using zero-based piece indices throughout, the entire stories
\emph{LW1}, \emph{Cable Spool Fort}, and \emph{Easy Money}, together with
pieces 0--4 of \emph{The Black Willow}, formed a 2698-segment development
set. Within this set, piece 4 of \emph{The Black Willow} was held out for
validation and excluded from neural-network parameter fitting. Pieces 5--11
of \emph{The Black Willow} formed the test set of 1005 candidate segments.
Although the 3\,s windows overlap within each partition, the
parameter-fitting, validation, and test partitions share no audio or MEG
samples. An audit of the actual MEG intervals used by the model across all
49 participant--session recordings, including the 150\,ms MEG--audio offset,
confirmed that the intervals are disjoint across split boundaries.
For \emph{The Black Willow}, the scaling parameters were fitted only on samples
preceding the first test-piece onset to avoid test leakage.
Each 3\,s audio segment was paired with the 3\,s MEG segment beginning
150\,ms later to accommodate the general delay of auditory neural
responses, following~\cite{Defossez2023}.
This 150\,ms offset shifts the entire MEG interval
relative to audio, irrespective of the linguistic features
present. It neither estimates the response latency of any
individual feature nor restricts the decoder to neural activity
occurring at that latency.
In contrast to~\cite{Defossez2023}, we did not align test segments to word
onsets, resulting in a less structured and therefore more challenging
retrieval setting.

\subsection{Training and testing}
\label{ssec:training}

Models were trained with the AdamW optimizer~\cite{Loshchilov2019}, using a
learning rate of $3\times10^{-4}$ for the network parameters and
$10^{-3}$ for the learned temperature, zero weight decay, a batch size of
100, and at most 50 epochs.
Early stopping used a patience of seven epochs and
checkpoint selection was based on the lowest validation loss.
Unless stated otherwise, the reported main checkpoint, architectural sweeps,
ablations, and interpretation analyses use seed 42.
For uncertainty estimates of the six-seed reference model, binary
Top-1 and Top-10 outcomes were first averaged across the aligned model
initializations. Participant-balanced performance was then computed by
averaging within each participant and weighting all 27 participants equally;
95\% confidence intervals were obtained from 10,000 bootstrap resamples of
participants. Stimulus variability was summarized separately across the seven
contiguous test pieces after averaging across model initializations.
All experiments used the $L=24$ spherical-harmonic setting defined in
Section~\ref{ssec:3d-attention}, corresponding to degrees
$0,\ldots,23$ and 576 real basis functions per virtual channel.
The training objective was a one-directional MEG-to-audio contrastive
cross-entropy loss. MEG-derived embeddings were compared with the unique
audio targets represented in the current minibatch. 
Within each minibatch, when multiple MEG-derived embeddings corresponded
to the same audio-segment target, they were all assigned to the same target
embedding. 
The resulting similarity matrix therefore had one row per MEG-derived
embedding and $U$ columns, where $U$ is the number of unique audio targets
in the minibatch. Using this matrix instead of a square batch-by-batch
similarity matrix prevents duplicate instances of the same positive audio
target from being treated as negatives in the cross-entropy loss.
Similarities were computed after $L_2$ normalization over the
feature--time dimensions and divided by a learned temperature parameter.
Training was performed on a cluster node with a
single NVIDIA Tesla V100 32\,GB GPU and 8 cores of Intel Xeon Gold 6152
CPU. 

To estimate the effect of a window size, we also evaluated paired MEG and audio
segments of \(W\in\{1.5,2.25,3,4,5\}\)\,s. We retained the start positions
of the original 3\,s segments rather than constructing a new set of
segments at each duration. For an anchor at time \(t\), the audio target
was regenerated directly from the continuous recording over
\([t,t+W]\), and the corresponding MEG input covered
\([t+150\,\mathrm{ms},t+W+150\,\mathrm{ms}]\). wav2vec 2.0 
audio embeddings were then recomputed separately for each
window size.

As longer window sizes usually result in fewer windows, to keep the training-set size and retrieval-database size constant across durations, we used a common set of window starts containing only start positions for
which a complete 5\,s segment remained within the audio. This yielded
991 test candidates at every duration. Separate models with \(K=25\)
branches and \(B\in\{0,2,5\}\) convolutional blocks were trained for each
duration, with all other training and checkpoint-selection procedures kept
unchanged. The 3\,s ablation condition was regenerated using this same
pipeline and common anchor set rather than copied from the main
1005-candidate experiment.

To examine the temporal scale required by the interpretable front end, we trained otherwise identical two-block, \(K=25\) models with temporal-filter kernel sizes of \(1, 5, 9, 15, 19, 25, 29, 49, 65, 81, 101, 151, 201, 251,\) and \(299\) samples, corresponding to supports of 10\,ms--2.99\,s at the 100\,Hz MEG sampling rate. The sweep was repeated for three training seeds (42--44); within each seed, all conditions used the same data split, optimizer, early-stopping rule, and validation-based checkpoint selection. The one-sample condition contains no temporal context and acts only as a learnable scaling of each branch.

\paragraph{Branch-wise functional importance.}
We quantified the contribution of individual branches in the six
trained \(K=25\), two-block models (seeds 42--47) with all network weights
frozen. Each branch was perturbed separately on the final test set in three
ways: its output was set to zero; it was set to zero while the remaining
branches were rescaled by \(25/24\) similar to the dropout formulation; 
or its output was replaced by a
within-participant permutation across test queries. The latter preserves the
participant and the marginal distribution of that branch while disrupting its
query-specific information. The audio candidate bank and all other branch
outputs were left unchanged. Importance was measured as the change in Top-1
and Top-10 retrieval relative to the unmodified checkpoint of the same seed.
Branch identities were analysed within each independently trained model rather
than aligned across seeds.

\section{Results}
\label{sec:results}

\subsection{Decoding accuracy}
\label{ssec:decoding-accuracy}

\begin{figure}[t]
\centering
\includegraphics[width=\linewidth]{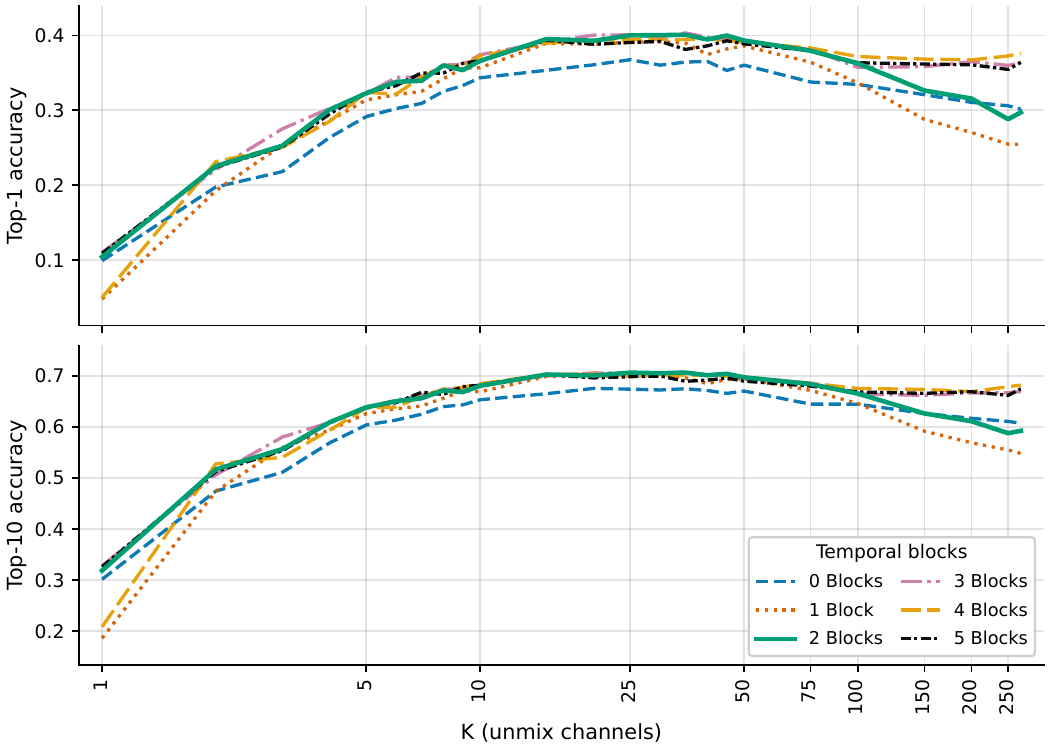}
\caption{
Retrieval accuracy as a function of the number of interpretable branches $K$ and the number of convolutional blocks in the decoder.
For each configuration, Top-1 and Top-10 test accuracy are reported at the checkpoint with the lowest validation loss.
Accuracy increases sharply from very small $K$ to approximately $K=10$--$25$, then enters a broad plateau; larger $K$ does not produce systematic gains and can mildly degrade performance.
Across decoder depths, the zero-block model is consistently weaker, while models with 2--5 convolutional blocks form a similar high-performing regime.
The main 2-conv, $K=25$ configuration lies on this compact high-accuracy plateau.
}
\label{fig:k_conv_blocks}
\end{figure}

We assess decoding efficacy on the held-out pieces 5--11 of \emph{The Black Willow}.
We evaluated the \(K\times B\) capacity grid on the validation
data (Appendix~\ref{app:architecture-validation}). Validation performance
shows a broad high-performing regime at intermediate \(K\), with little
systematic benefit from increasing decoder depth beyond a shallow nonlinear
decoder. Figure~\ref{fig:k_conv_blocks} reports the corresponding final-test
performance across the capacity grid.

On the final test set, both metrics show a sharp improvement when \(K\) increases from very small values to the intermediate regime. With only one or a few branches, the model is strongly bottlenecked: the branch space is too small to preserve the information needed to distinguish the correct audio segment from the candidate set. 
Performance then rises rapidly and reaches a broad high-accuracy plateau around
\(K=10\)--\(25\). Thus, retrieval-relevant MEG information can be compressed into a few
tens of learned branch components without sacrificing performance relative to
much larger branch spaces. 
In particular, the sweep identifies a compact high-performing regime
around \(K=10\)--\(25\), showing that substantially larger branch spaces are
not required for near-maximal retrieval in this architecture.
Increasing \(K\) toward hundreds does not produce systematic gains.

The behavior at large $K$ is also informative. If the branch dimension were simply acting as generic model capacity, one would expect performance to keep improving, or at least remain monotonic, as $K$ grows. Instead, several curves flatten and then mildly decline at larger $K$. This suggests that the useful task-related subspace is compact, and that adding many extra branch channels mostly increases redundant or harder-to-regularize degrees of freedom rather than adding new decodable information.

Convolutional depth modulates this pattern but does not change the overall conclusion. The model without convolutional blocks is consistently weaker once $K$ enters the useful range, showing that the compact branch representation still needs downstream temporal/contextual processing. Adding convolutional blocks substantially improves retrieval, but the gains saturate: models with two or more blocks occupy a similar high-performing regime over the plateau. Deeper variants can match or slightly exceed the two-block model at some values of $K$, but they do not produce a qualitatively different dependence on branch dimensionality.

We use \(K=25\), \(B=2\) as the fixed reference architecture for the
mechanistic and representation analyses below unless stated otherwise. This
configuration lies within the broad high-performing regime in both the
validation and final-test capacity landscapes.
The architecture-capacity sweep and the choice of this
reference configuration are exploratory: multiple configurations
were evaluated on the same final-test material, without a separate
confirmatory evaluation on an independently held-out narrative.
We therefore interpret these comparisons in terms of broad
capacity--performance trends rather than as independent validation
of a uniquely optimal architecture or fine-grained rankings
among configurations.

For this reference architecture, equal-participant averaging yielded
39.90\% Top-1 accuracy (95\% participant-bootstrap CI,
34.31--45.84\%) and 70.70\% Top-10 accuracy (95\% CI,
65.40--75.89\%). Across the seven contiguous test pieces, seed-averaged
accuracy was \(41.06\pm5.34\%\) Top-1 and \(71.47\pm3.77\%\) Top-10
(mean \(\pm\) SD across pieces).

\begin{figure}[t]
    \centering
    \includegraphics[width=0.8\linewidth]{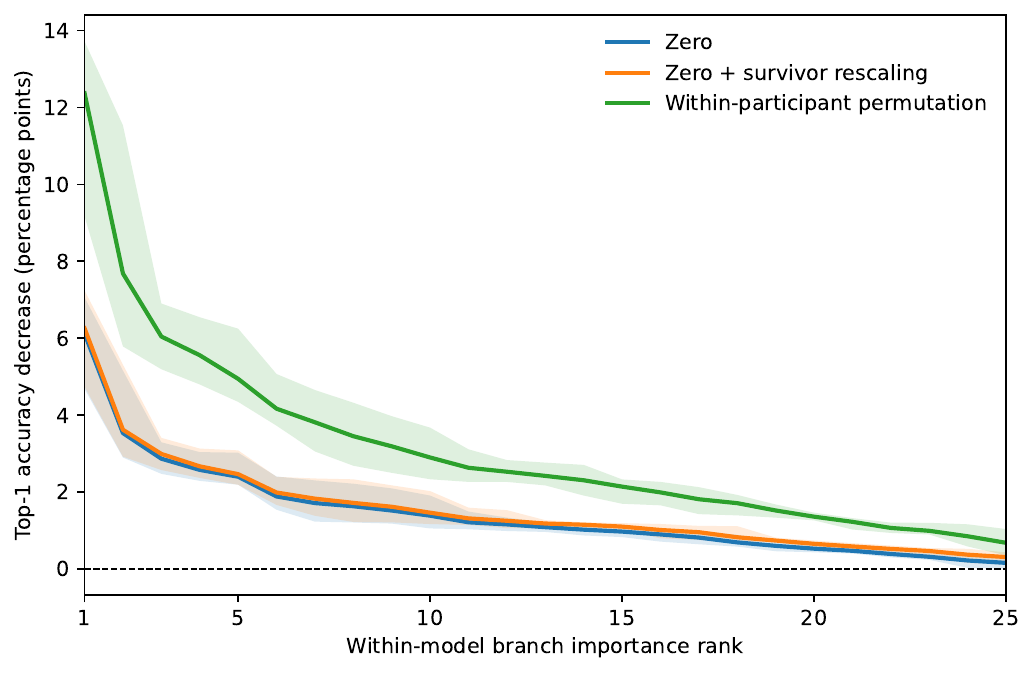}
    \caption{
    Branch-wise functional importance across six independently
    trained \(K=25\) models. Within each seed and intervention, branches are
    ranked by the decrease in Top-1 accuracy, with rank 1 denoting the largest
    decrease. Lines show the mean across seeds and shaded regions show the
    minimum--maximum range. Zeroing was evaluated both directly and with the
    remaining branches rescaled by \(25/24\); the permutation intervention
    replaced each branch by a within-participant permutation across test queries.
    Ranks are defined independently within each model and therefore do not imply
    branch correspondence across seeds.}
    \label{fig:branch-importance}
\end{figure}

\paragraph{Branch-wise functional importance.}
Individual branches contributed unequally to retrieval (see Fig.~\ref{fig:branch-importance}). Nevertheless,
the observed effect of zeroing any single branch was detrimental for both
Top-1 and Top-10 accuracy in all \(6\times25=150\) branch--seed cases.
Across the six models, the within-model median Top-1 decrease from branch
zeroing ranged from 0.96 to 1.21 percentage points; the weakest branch in each
model produced a decrease of 0.004--0.264 points, whereas the strongest
produced a decrease of 4.68--7.06 points. Rescaling the surviving branches by
\(25/24\) gave a similar distribution, with median decreases of
1.06--1.27 points. Disrupting query-specific branch information by
within-participant permutation produced larger effects: median Top-1 decreases
of 2.17--2.76 points and maximal decreases of 9.17--13.72 points across
seeds. Thus, the \(K=25\) representations contain branches with widely
different marginal contributions to retrieval, including a tail of relatively
weak components.

\subsection{Modified attention scores}
\label{ssec:attention-maps}

\begin{figure}[t]
\centering
\includegraphics[width=\linewidth]{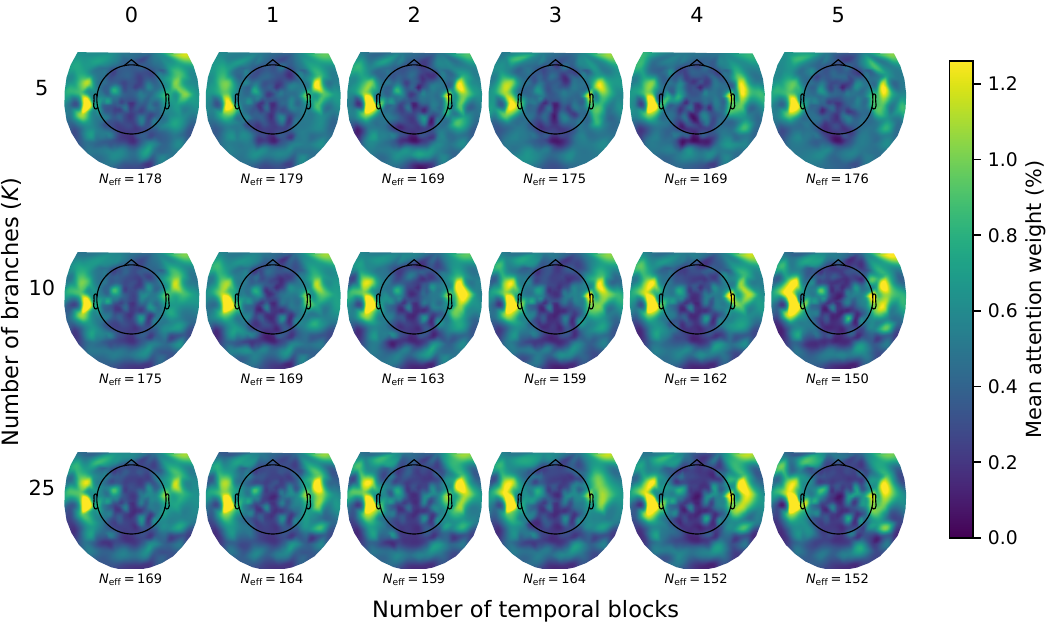}
\caption{3D spherical-harmonic attention learned by architectures with varying numbers of non-linear convolutional blocks ($B$) and branch counts ($K = 5, 10, 25$). $N_{\mathrm{eff}}=(\sum_m p_m^2)^{-1}$ is the inverse-Simpson effective
number of sensors, where
$p_m=\bar c_m/\sum_{m'}\bar c_{m'}$; smaller values indicate that attention is
concentrated on fewer sensors. For visualization maximal value was capped to 99-th percentile, but all $N_{\mathrm{eff}}$ are computed with full attention weights without clipping. 
These maps visualize first-stage attention weights only and are not
anatomical localizations or complete participant-specific spatial filters.
}
\label{fig:attention-maps-vs-blocks-branches}
\end{figure}

Figure~\ref{fig:attention-maps-vs-blocks-branches} compares the mean
softmax-normalized attention weights across the $J=270$ virtual channels,
\begin{equation}
    \bar c_m =
    \frac{100}{J}\sum_{j=1}^{J}\widetilde c_{jm},
\end{equation}
for architectures with varying numbers of non-linear convolutional blocks
($B$) and branch counts ($K=5,10,25$). Across the grid, increasing decoder
depth and branch count is generally accompanied by progressively more
spatially structured and concentrated attention. This trend is reflected in
the reduction of $N_{\mathrm{eff}}$, the effective number of sensors carrying
the attention mass. Because every map is normalized to the same total
attention mass, these differences represent a redistribution of attention
across sensors rather than a change in its overall magnitude.

These maps show only the first shared spatial-attention stage and are
not interpreted anatomically. Subsequent shared unmixing and
participant-specific linear transformations can reinforce, combine, or cancel
individual attention channels. Physiological interpretation therefore uses the
complete effective participant-specific spatial filters
\(\mathbf W^{(s)}\) and their corresponding spatial patterns in the subsequent
analyses.

\subsection{Cross-participant spatial structure}
\label{ssec:cross-subject}

\begin{figure}[!ht]
\centering
\begin{subfigure}[t]{\linewidth}
 \centering
  \includegraphics[width=0.9\linewidth]{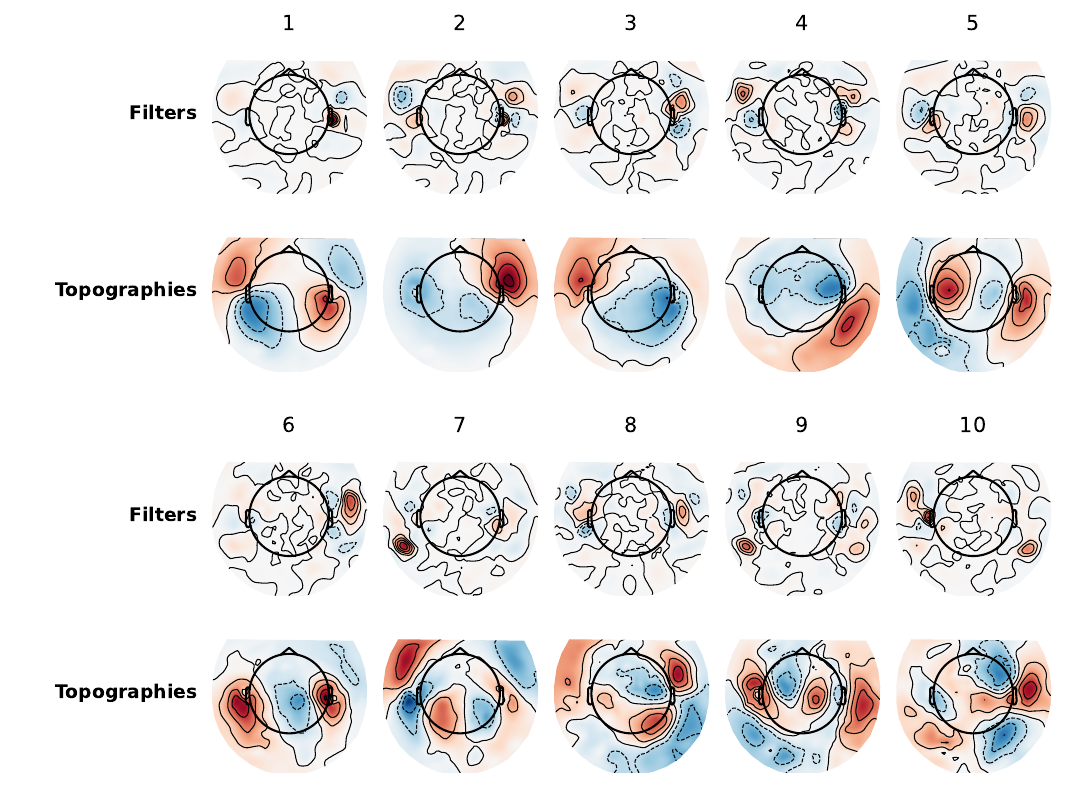}
  \caption{First ten singular vectors calculated separately for all $L_2$-normed spatial filters and topographies.}
 \label{fig:cross-subject-components}
\end{subfigure}
\vspace{0.02\textheight}
\begin{subfigure}[t]{\linewidth}
 \centering
  \includegraphics[width=0.8\linewidth]{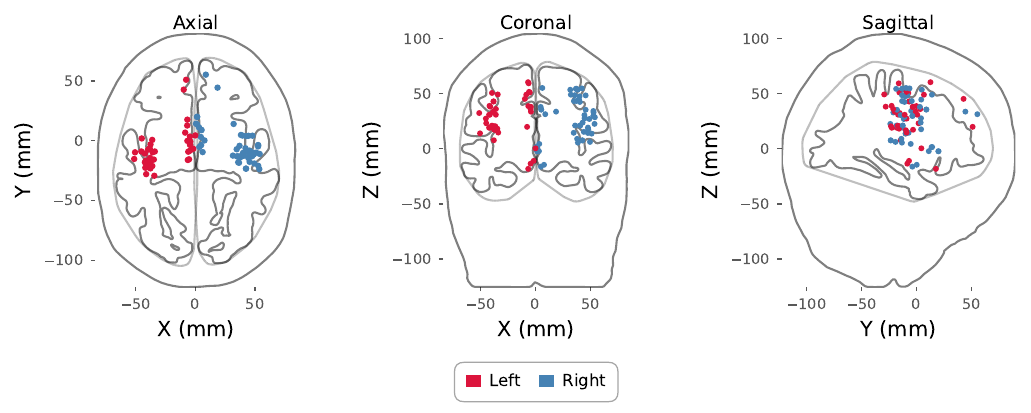}
  \caption{Participant-specific RAP-MUSIC candidate localizations of
    the shared cross-subject spatial-pattern subspace in
    Figure~\ref{fig:cross-subject-components}, pooled in \texttt{fsaverage}
    coordinates for visualization (subspace correlation threshold \(0.98\)).}
\label{fig:rap-music-dipoles}
\end{subfigure}
\caption{Leading ten singular vectors of the across-subject spatial
filter and spatial-pattern matrices aggregated from the interpretable branches
of all 27 participants, together with participant-specific RAP-MUSIC candidate
localizations pooled in the common \texttt{fsaverage} coordinate frame.}
\label{fig:cross-subject-patterns}
\end{figure}

Analysis of the spatial topographies aggregated over 27 participants
characterizes the cross-participant spatial structure of the learned branch
representation beyond the first-stage attention weights.
To probe the cross-subject structure of the learned representations we
aggregated the spatial filters from all interpretable branches across all 27 subjects: 
for each of the 25 branches in the subject-specific interpretable block,
we collected the corresponding $208 \times 1$ spatial filter for each
participant, yielding 675 spatial filters in total.
For this analysis, spatial patterns were computed from the MEG
recording obtained during the first session of \emph{Cable Spool Fort}. We
excluded 10\,s from each end of the recording. Before SVD, we applied the same
spatial-roughness screen as in Section~\ref{ssec:clustering}; the retained
spatial filters and spatial patterns were then normalized to unit \(L_2\)
norm. Consequently, this decomposition summarizes recurring spatial structure
across the learned branches rather than weighting patterns by their marginal
contribution to retrieval; branch-wise functional contribution is quantified
separately by the intervention analysis above.
We then computed spatial patterns following~\cite{Haufe2014} and performed SVD on both the filter and pattern matrices, extracting the leading ten singular vectors from each. 
The dominant spatial patterns in
Figure~\ref{fig:cross-subject-patterns} show several distinct bilateral and
lateralized sensor-space configurations, with prominent extrema over lateral
sensor regions.

Unlike the spatial attention weights in Figure \ref{fig:attention-maps-vs-blocks-branches} the spatial topography patterns from Figure~\ref{fig:cross-subject-patterns} result from the principled approach to jointly treat the spatial and temporal processing happening in each branch of the network's front-end \citep{Petrosyan2021}. This approach recognizes the role of both the spatial and the temporal filters in not only tuning to the target sources but also de-tuning from the interference. 
It also permits inverse-model localization of the corresponding
learned sensor topographies.

To roughly localize the sources underlying these dominant spatial patterns,
we applied a variant of RAP--MUSIC~\cite{Mosher2002} separately with each participant's forward model to the shared sensor-space subspace
spanned by the top ten singular vectors of the spatial-pattern matrix. 
At each recursion, we selected the cortical location with the highest subspace
correlation and projected the full local three-orientation lead-field
subspace from both the spatial-pattern subspace and the remaining candidate
lead fields, as described in Appendix~\ref{app:rap-music}. Using a
subspace-correlation threshold of $0.98$, the procedure identified the
candidate source locations shown on the \texttt{fsaverage} anatomy in
Figure~\ref{fig:rap-music-dipoles}. 
The resulting participant-specific source locations were transformed to the common \texttt{fsaverage} coordinate frame and pooled for visualization.
Across participants, the candidate locations were distributed
bilaterally over lateral cortex, with the densest concentrations around
temporal and adjacent peri-Sylvian regions and sparser locations elsewhere. Noteworthy is the distinct cluster in the right motor area, which likely corresponds to cluster 11 in Figure \ref{fig:patterns}.

\subsection{Clustered spatial and temporal patterns}
\label{ssec:patterns}

\begin{figure}[!ht]
\centering
\includegraphics[width=0.95\linewidth]{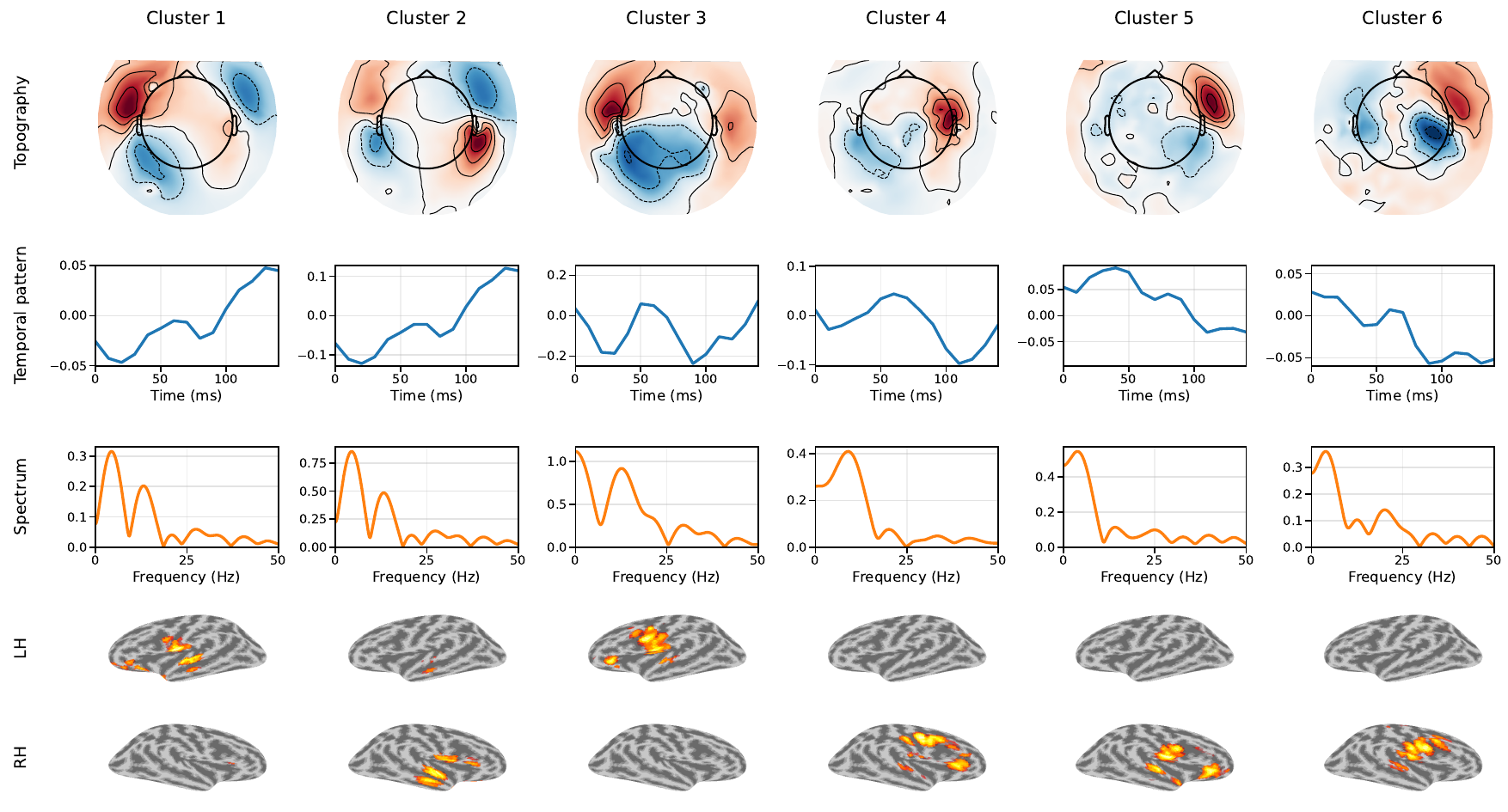}
\vspace{0.5em}
\includegraphics[width=0.95\linewidth]{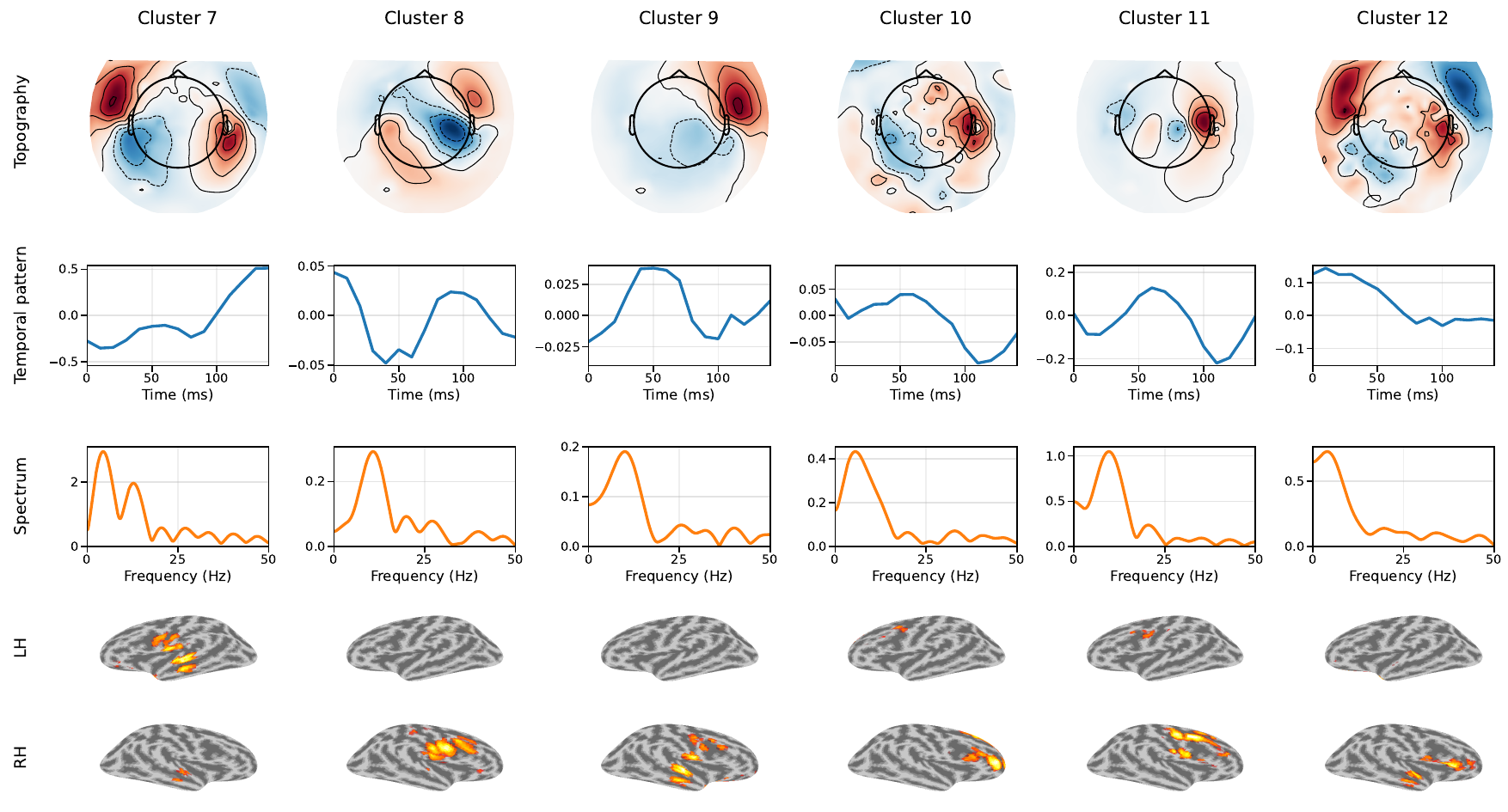}
\caption{The 12 largest clusters for the main \(K=25\) model, computed from the \emph{Cable Spool Fort} recordings in the first session after spatial-roughness screening. Each column shows the medoid of one cluster. Rows show, from top to bottom, the sensor-space spatial pattern, the 150\,ms temporal pattern, its magnitude spectrum, and the corresponding participant-specific MNE-Python~\cite{Gramfort2013} source-magnitude estimate on the common \texttt{fsaverage} surface in left- and right-hemisphere lateral views.}
\label{fig:patterns}
\end{figure}

Figure~\ref{fig:patterns} shows the medoids of the twelve largest of
the 57 spatial--spectral clusters obtained as described in
Section~\ref{ssec:clustering}. Each comprises the 150\,ms temporal pattern of
Eq.~\eqref{eq:temporal-pattern}, its magnitude spectrum, and the sensor-space
spatial pattern with the corresponding source-magnitude estimate.

The spatial-roughness screen excluded 104 of the 675 participant--branch
items, leaving 571 items for clustering. The excluded items were strongly
enriched for very-low-frequency temporal structure and had spatially irregular
sensor topographies, consistent with branches dominated primarily by nuisance
or background fluctuations rather than stable physiological source patterns.
Across the full 57-cluster partition, cluster size ranged from 1 to 39
participant--branch items (median 7), and participant representation ranged
from 1 to 18 participants (median 5); 10 clusters contained items from only
one participant. The twelve displayed clusters contain 14--39 items each.
Eleven include branches from 6--18 participants, whereas Cluster~10 contains
16 branches from a single participant and should therefore not be interpreted
as a population-recurrent motif. 
Because each medoid is an observed participant--branch item rather
than an average pattern, we quantified how well it represents the remaining
members of its cluster. Across the twelve displayed clusters, the mean raw
Pearson correlation between the medoid and the other cluster members ranged
from \(0.58\) to \(0.79\) for cortical source maps and from \(0.68\) to
\(0.79\) for temporal spectra. Pooling the 265 member--medoid comparisons,
\(99.2\%\) had correlations of at least \(0.40\) in both views and \(91.7\%\)
had correlations of at least \(0.50\) in both views.
Stability of the underlying
branch interpretations across stories, repeated sessions, and model
initializations is reported in
Appendix~\ref{app:interpretation-stability}. A separate descriptive
analysis in the same appendix examines sensitivity of cluster medoid
selection to exclusion of low-importance branches under the original
cluster assignments.
A comparison of the principal spatial--spectral clusters across
six independently trained models also reveals recurring broad lateral
temporal and peri-Sylvian source-space patterns, although individual
cortical foci and temporal spectra vary between initializations
(Figure~\ref{fig:cross-seed-motifs}).

The retained clusters fall into broad spatial groups whose
distributions are consistent with cortical systems implicated in speech
perception. The largest shows a bilateral projection along the
superior temporal gyrus and superior temporal plane, where continuous tracking
of the speech envelope has been localized with both invasive and non-invasive
recordings~\cite{Ding2012,Gross2013,Nourski2009,Kubanek2013,Brodbeck2020} and
where the acoustic-edge response has been reported~\cite{Oganian2019}. 
This interpretation is also consistent with the paired occlusion analysis in
Section~\ref{ssec:occlusion-results}, where
replacing silent, loud, vowel and acoustic-onset intervals
with feature-absent donor MEG degrades retrieval most. 

A second group extends posteriorly and superiorly toward the supramarginal
region and the bank of the central sulcus. 
These projected regions overlap dorsal-stream sites in the
dual-stream account~\cite{Hickok2007}, which assigns phonological and
sensorimotor functions to posterior temporal and inferior parietal cortex.
Their presence is consistent with the smaller but reliable effects for
fricatives, stops and sibilants, and with reports that passive speech
perception recruits articulatory
representations~\cite{Fadiga2002,Liu2023,Eliades2008}.

A third group combines more anterior middle and inferior temporal foci with additional frontal contributions.
Functionally, this distribution aligns with the
accounts in which temporal-lobe perceptual predictions are reconciled in
inferior frontal cortex~\cite{Cope2023}, and the temporal foci fit
auditory--conceptual integration~\cite{Bonilha2017}. 

The retained clusters also differ in their temporal spectra.
Several clusters show two components in their temporal patterns: a dominant
slow component below approximately \(10\)\,Hz and a second, weaker faster
component broadly expressed around \(10\)--\(15\)\,Hz. This faster structure
is present in clusters with both left- and right-lateralized source maps.
The slow component is common to many clusters and is consistent
with an evoked response.

The faster components of the learned temporal patterns
overlap the alpha and lower-beta frequency ranges, where neural
oscillations have been linked to predictive and top-down processing
during naturalistic speech comprehension~\cite{zioga2023naturalistic}.
Together with the common slow component, this suggests that different branches capture complementary temporal dynamics within the speech-responsive network.

The temporal filters are $15$ samples long at
$100$\,Hz, which gives a frequency resolution of roughly $6.7$\,Hz. 
A complementary analysis of a 1.51\,s model provides approximately
0.66\,Hz independent Fourier-bin spacing and recovers a similar broad
spatial/source organization while resolving repeated secondary maxima across
approximately \(8\)--\(17\)\,Hz, showing that the broad faster structure
observed with 150\,ms temporal support persists when the temporal support is
increased by an order of magnitude and the spectral resolution is substantially
refined; see
Appendix~\ref{app:long-temporal-patterns}.
The extended filter-length experiment in Figure~\ref{fig:temporal-filter-support} shows that retrieval remains strong over substantially longer supports, with mean Top-1 highest at 2.51\,s and mean Top-10 at 1.51\,s before both decline at 2.99\,s.

\subsection{Paired occlusion identifies stimulus-associated MEG information used for retrieval}
\label{ssec:occlusion-results}

\begin{figure}[t]
    \centering
    \includegraphics[width=\linewidth]{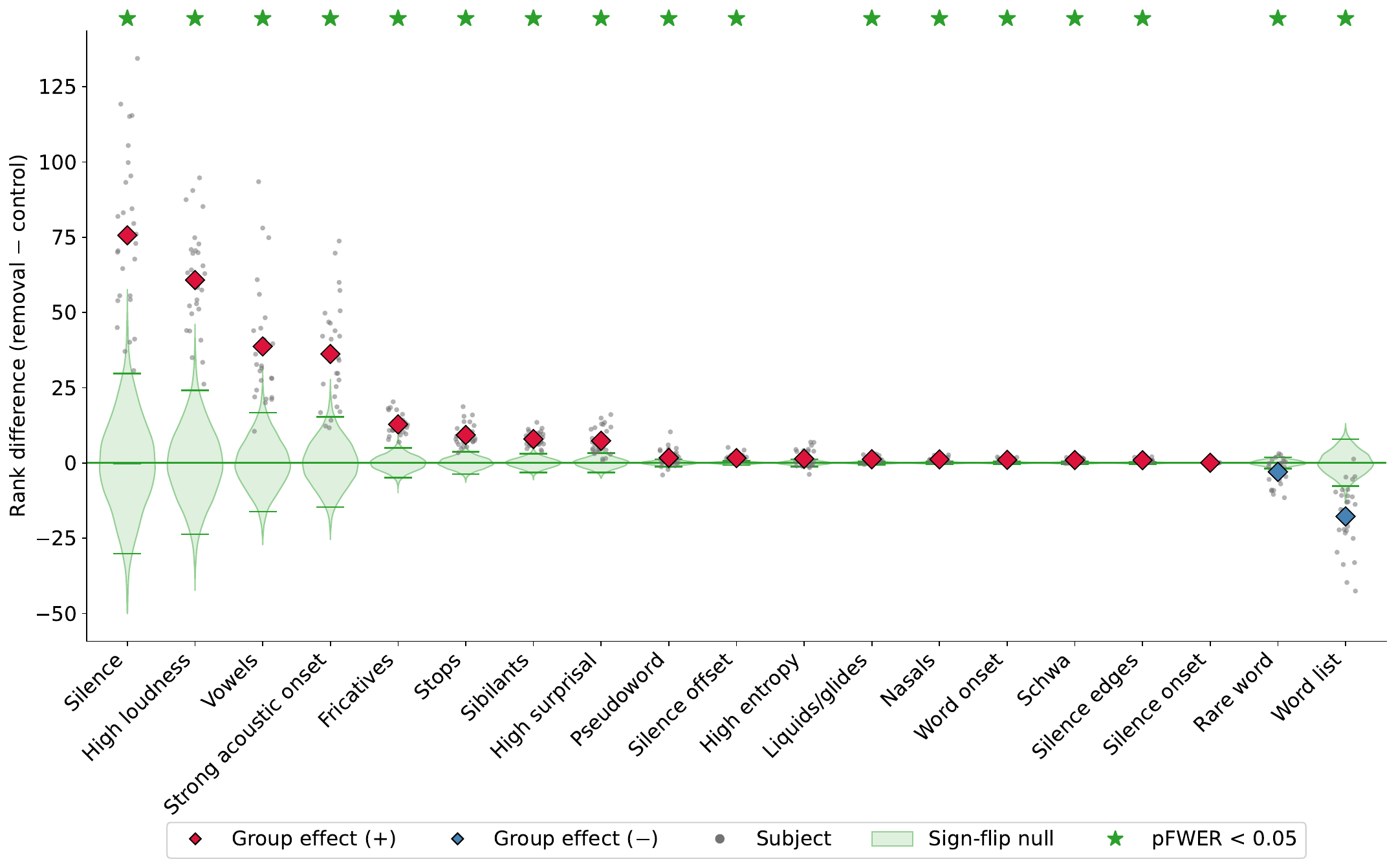}
    \caption{
    Paired MEG occlusion effects for 19 stimulus features. For each
    participant, the plotted effect is the retrieval-rank difference between
    feature-absent replacement (``removal'') and matched feature-present
    replacement (``control''), after averaging donor realisations, eligible
    queries, and multiple sessions. Positive values indicate that preserving
    feature-associated MEG information retained a better rank. Grey points show
    participant effects and diamonds show group means. Green violins show the
    feature-wise sign-flip null distributions in rank-difference units; stars
    mark two-sided single-step max-$|T|$ familywise-error-corrected
    $p<0.05$. Every rank was computed against the complete, unchanged bank of
    1005 candidates. Because the masks differed in duration and in their sets
    of eligible queries, effect magnitudes should not be read as a calibrated
    ranking of feature-encoding strength across features.
    }
    \label{fig:occlusion-main}
\end{figure}

The paired rank contrast
\(
\Delta r_f=r_{f\rightarrow 0}-r_{f\rightarrow f}
\)
was positive after familywise-error correction for 15 of the 19 tested
features (Figure~\ref{fig:occlusion-main}). Thus, for these features, real MEG
from feature-present intervals preserved the correct candidate rank better
than equally sized replacements from feature-absent intervals. None of the
feature-present controls approached random ranking: their participant-balanced
mean ranks ranged from 23.2 to 153.5, compared with an expected rank of 503
under random ordering.
Because the substitution replaces more MEG for long contiguous events such as
silence than for brief ones such as word onsets, the ordering of effect
magnitudes across features is confounded by the duration of the replaced
intervals, so the ranking should be read as identifying which features the
decoder uses rather than how much each contributes.
The feature masks also differed substantially in their mutual overlap
(Appendix~\ref{app:feature-mask-overlap}). In particular, high loudness and
strong acoustic onset had a Jaccard overlap of 0.39, whereas silence overlapped
little with either high loudness (0.008) or strong acoustic onset (0.032).
Sibilants and schwa were nested subclasses of fricatives and vowels,
respectively. Accordingly, the linguistic  contrasts are interpreted as 
stimulus-state-associated MEG differences rather than as
covariate-adjusted effects of individual lexical variables.

The largest observed contrasts were obtained for silence
(\(\Delta r=75.62\)), high loudness (\(60.77\)), vowels (\(38.70\)), and
strong acoustic onset (\(36.19\)); all had
\(p_{\mathrm{FWER}}<10^{-4}\), and the effect was positive in all 27
participants. These results show that the decoder uses MEG distinctions
associated with broad speech state, sound intensity, rapid acoustic change,
and phonetic content. 

Further corrected positive effects were observed for fricatives
(\(\Delta r=12.80\)), stops (\(9.27\)), sibilants (\(7.94\)), and high
surprisal (\(7.34\)). Smaller effects remained significant for pseudowords
(\(1.71\), \(p_{\mathrm{FWER}}=0.039\)), silence offsets (\(1.60\)),
liquids/glides (\(1.21\)), nasals (\(1.18\)), word onsets (\(1.02\)), schwa
(\(0.96\)), and the combined silence-edge mask (\(0.91\)); except for
pseudowords, these effects had \(p_{\mathrm{FWER}}<10^{-4}\).

High predictive entropy
(\(\Delta r=1.40\), \(p_{\mathrm{FWER}}=0.119\)) and silence onsets
(\(0.05\), \(p_{\mathrm{FWER}}=0.494\)) were inconclusive after correction.
Rare words (\(-2.98\), \(p_{\mathrm{FWER}}=0.011\)) and random word lists
(\(-17.77\), \(p_{\mathrm{FWER}}<10^{-4}\)) showed corrected negative effects.

The negative random-word-list effect indicates that replacing
word-list MEG with MEG recorded during coherent narrative improved retrieval.
This result contrasts with the smaller positive pseudoword effect. 
A plausible explanation for the opposite signs is that a
pseudoword, embedded in an otherwise coherent sentence, can evoke an
additional response to a local linguistic violation, as in the
N400~\cite{kutas1980reading,lau2008cortical}, whereas random word lists
lack the sentence- and narrative-level structure that normally supports
ongoing neural tracking~\cite{Gillis2023}.
Because the word-list and
coherent-narrative states also differ in acoustic, lexical, and phonetic
properties, however, this contrast does not isolate narrative structure alone.

As a robustness check across model initializations, all 15 corrected
positive effects from the seed-42 analysis remained positive and significant
in all six trained models, while the random-word-list effect remained negative
and significant in all six
(Appendix~\ref{app:occlusion-seed-robustness}).

\subsection{Effect of segment duration on retrieval}
\label{ssec:window-duration}

\begin{figure}[t]
    \centering
    \includegraphics[width=\linewidth]
    {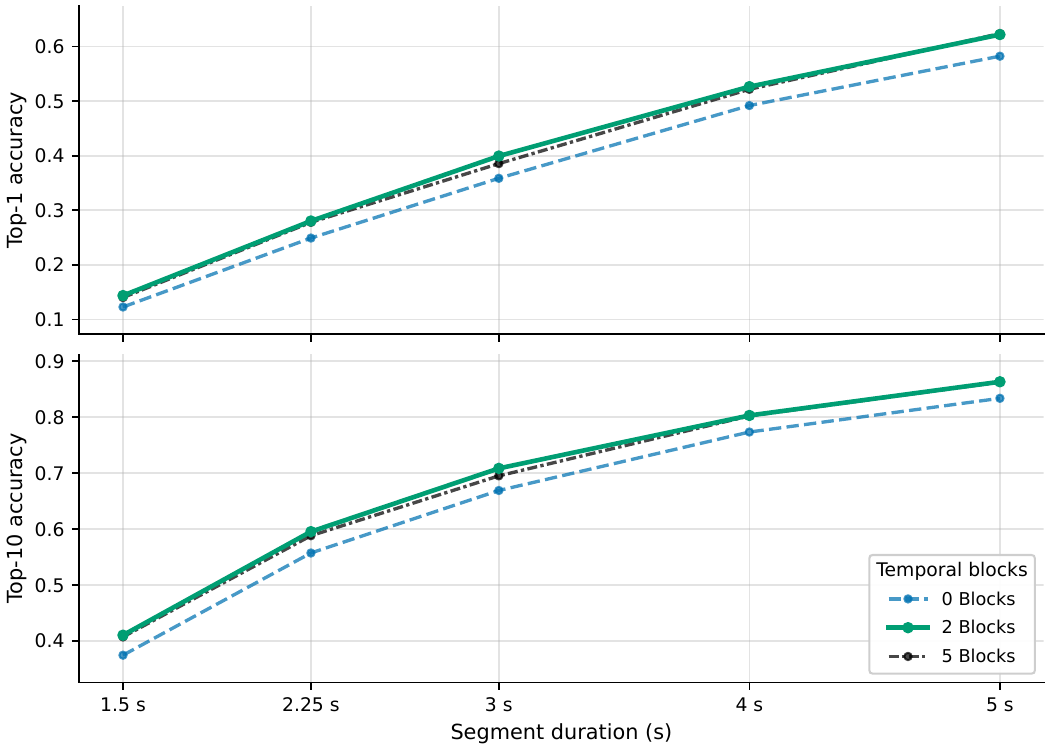}
    \caption{
    Test retrieval accuracy as a function of paired MEG--audio segment
    duration for models with \(K=25\) branches and \(B\in\{0,2,5\}\)
    convolutional blocks. Each point reports final-test accuracy from the
    checkpoint with the lowest validation loss. Audio embeddings were
    regenerated directly from the continuous sounds at every duration, and
    all conditions used the same 5\,s-feasible anchor set and the same
    991-candidate retrieval database. The 3\,s points belong to this
    regenerated ablation and are distinct from the main 1005-candidate
    evaluation.
    }
    \label{fig:window-duration-accuracy}
\end{figure}

Retrieval accuracy increased monotonically with paired MEG--audio segment
duration for all three decoder depths
(Figure~\ref{fig:window-duration-accuracy}). For the main two-block model,
Top-1 accuracy increased from \(14.37\%\) at 1.5\,s to \(28.03\%\) at
2.25\,s, \(39.94\%\) at 3\,s, \(52.66\%\) at 4\,s, and \(62.20\%\) at
5\,s. Top-10 accuracy followed the same progression, increasing from
\(41.00\%\) to \(59.52\%\), \(70.84\%\), \(80.31\%\), and \(86.33\%\),
respectively.

The same duration dependence was present with zero and five convolutional
blocks. The five-block model closely followed the two-block model, reaching
\(62.38\%\) Top-1 and \(86.27\%\) Top-10 accuracy at 5\,s. Removing the
convolutional blocks reduced performance at every duration, although the
increase with segment length remained monotonic. The zero-block model
reached \(58.24\%\) Top-1 and \(83.37\%\) Top-10 accuracy at 5\,s. Thus,
the benefit of longer segments was robust to decoder depth, while the
downstream convolutional module provided an additional but comparatively
modest improvement.

\begin{figure}[t]
    \centering
    \includegraphics[width=\linewidth]
    {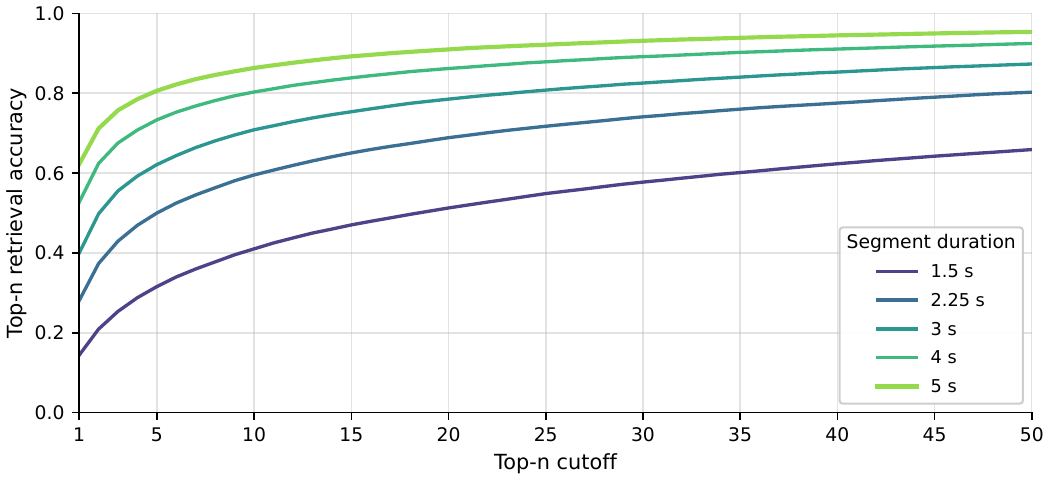}
    \caption{
    Top-\(n\) test retrieval accuracy for the main two-block,
    \(K=25\) model across paired MEG--audio segment durations. Curves were
    calculated using the similarity rank of the correct audio segment among the same
    991 candidates. The advantage of longer segments is
    present throughout the evaluated range \(n=1,\ldots,50\).
    }
    \label{fig:window-duration-topn}
\end{figure}

The complete rank curves showed that the improvement was not restricted to
Top-1 or Top-10 accuracy
(Figure~\ref{fig:window-duration-topn}). For the main two-block model, we estimated Top-\(n\) accuracy at different cutoffs from Top-1 to Top-50. The
ordering by segment duration was preserved at every retrieval cutoff. At \(n=50\), accuracy increased from \(65.88\%\) for
1.5\,s segments to \(80.26\%\), \(87.32\%\), \(92.47\%\), and \(95.37\%\)
for 2.25, 3, 4, and 5\,s segments, respectively. Longer paired segments
therefore improved both exact retrieval and the broader ranking of the
correct audio candidate.

Because the MEG input and the audio target were lengthened together, this
experiment measures the effect of paired segment duration rather than
isolating neural context integration alone. The gains may reflect both the
additional neural information available in a longer MEG segment and the
greater amount or distinctiveness of information in the corresponding
audio target.

\subsection{Feature-space compression}
\label{sec:feature_compression}

\begin{figure}[t]
    \centering
    \includegraphics[width=\linewidth]{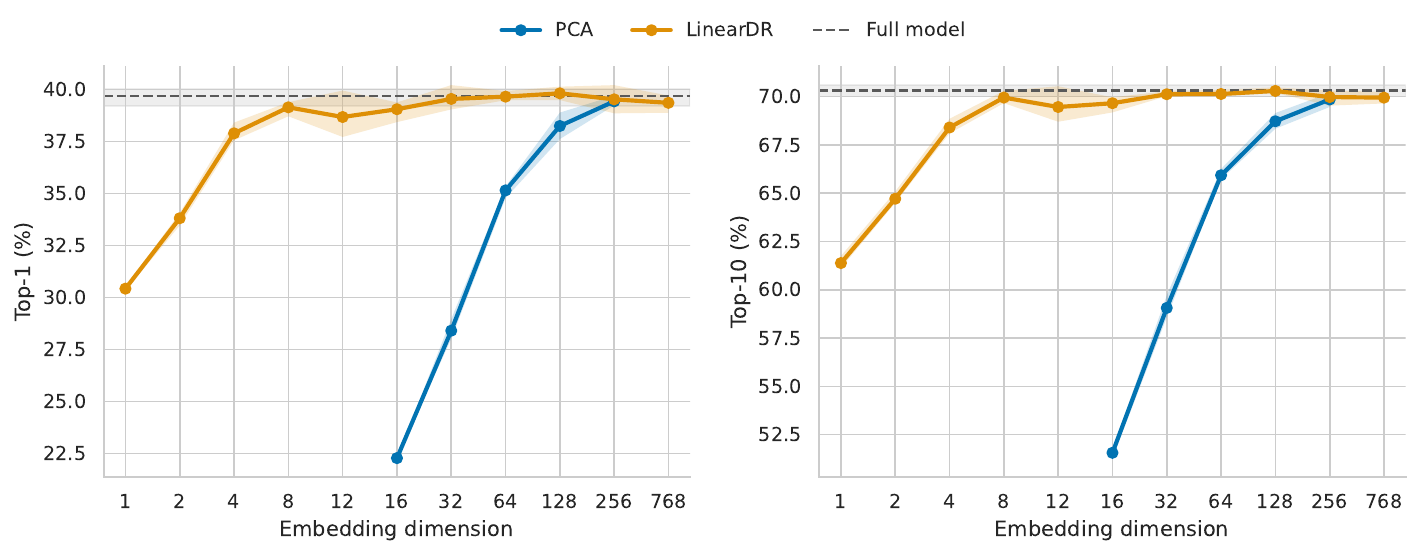}
    \caption{
    Feature-space compression of the target wav2vec representation.
    The feature dimension is reduced using either a fixed PCA projection or a trainable linear projection optimized with the retrieval loss.
    Curves show the mean across three matched training seeds (42--44), with shaded regions spanning the minimum and maximum across seeds. Learned feature reduction preserves retrieval accuracy over a broad low-dimensional range, whereas PCA degrades substantially faster as dimensionality is reduced.
    }
    \label{fig:feature_reduction}
\end{figure}

We first tested whether the full wav2vec feature dimension is necessary for retrieval, or whether the MEG-decodable part of the target representation lies in a much smaller feature subspace. We repeated the feature-reduction experiment across three matched training seeds (42--44). In all experiments in this section, the MEG encoder architecture was kept fixed to the same main configuration with 25 branches and 2 convolutional blocks and only the feature dimension of the audio target was changed (Fig.~\ref{fig:feature_reduction}).

We compare two types of feature reduction. In the PCA condition, we fit a fixed PCA projection on the training audio embeddings, treating all time points from all training segments as samples in the original wav2vec feature space. The MEG model is then trained to predict these PCA-reduced audio targets. This is an unsupervised reduction: it preserves high-variance directions in the audio embedding distribution, but it does not know which directions are recoverable from MEG or useful for retrieval. In the LinearDR condition, the audio embedding is passed through a trainable linear projection along the feature axis, implemented as a $1 \times 1$ convolution over feature channels. This projection is optimized jointly with the MEG decoder through the contrastive retrieval loss, so it is allowed to select a task-specific feature subspace.

The key result is that feature compression is highly effective when the projection is learned with the retrieval objective, and this pattern is preserved across the three training seeds. The learned projection shows a broad low-dimensional performance plateau rather than a sharply preferred dimensionality. Relative to the matched unreduced models, mean Top-1 accuracy changed by only $-0.15$, $-0.04$, and $+0.12$ percentage points at 32, 64, and 128 dimensions, respectively; even an 8-dimensional projection reduced mean Top-1 accuracy by only $0.55$ percentage points. Thus, a small learned linear subspace retains most of the retrieval-relevant information in the original 768-dimensional wav2vec 2.0 representation. For MEG-to-audio retrieval, the useful alignment is therefore concentrated in a compact, task-driven feature subspace.

PCA shows a different pattern. At 256 dimensions its mean performance remains close to the unreduced baseline, but degradation increases rapidly as the retained dimensionality is reduced: at 128 and 64 dimensions, mean Top-1 accuracy is lower by 1.45 and 4.55 percentage points, respectively. This separation between PCA and LinearDR is informative. It shows that the relevant feature directions are not simply the leading variance directions of wav2vec. High-variance audio-embedding components are not necessarily the components that are decodable from MEG, and conversely, MEG-relevant components may occupy directions that PCA does not preserve early. Thus, the feature-compression experiment supports a supervised subspace interpretation: the useful target space is low-dimensional, but it must be found through the neural decoding task rather than through variance preservation alone.

\subsection{Temporal-resolution compression}
\label{sec:time_compression}

We then performed an analogous experiment along the temporal axis of the target embedding. Here the question is different: can the model retrieve the correct 3-second audio segment from a coarse summary of its wav2vec trajectory, or does it need a temporally resolved target representation?

\begin{figure}[t]
    \centering
    \includegraphics[width=\linewidth]{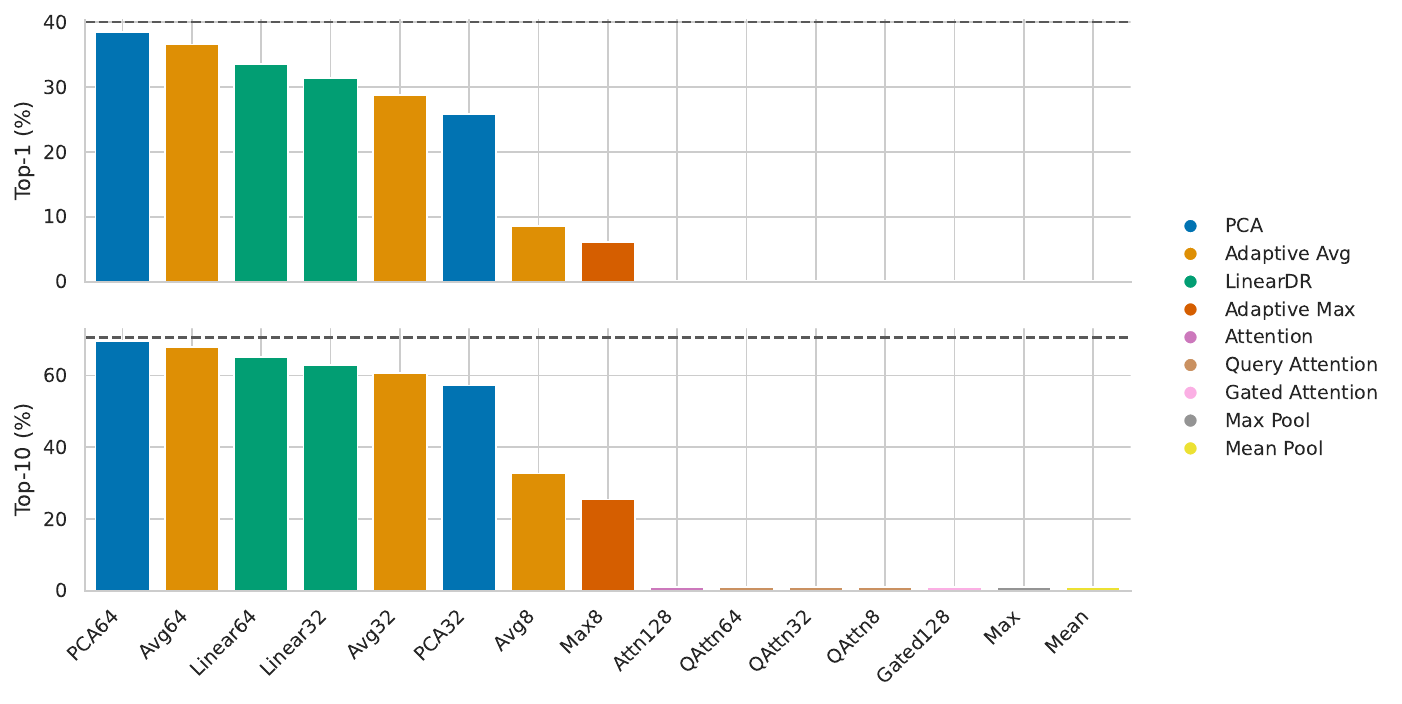}
    \caption{
    Temporal-resolution compression of the target wav2vec trajectory.
    Temporal PCA, trainable linear reduction, local pooling, and global pooling
    methods are compared. Each condition is a separately trained seed-42
    model. The comparison is descriptive: aggressive temporal compression causes
    a clear performance loss, and global pooling collapses to near-chance
    retrieval.
    }
    \label{fig:time_reduction}
\end{figure}

We evaluated several temporal reductions. Time PCA fits a fixed PCA
projection over the sequence length of the training audio embeddings, while
Time LinearDR uses separate trainable linear maps along the temporal axis
for the MEG-derived and audio embeddings. 
Adaptive
average and max pooling reduce the sequence to a fixed number of local bins.
Mean pooling, max pooling, additive attention pooling, and gated attention
pooling each produce one global vector. Query-attention pooling instead uses
a learned set of queries and produces a sequence of $Q$ outputs, with
$Q\in\{8,32,64\}$ in the evaluated conditions
(Fig.~\ref{fig:time_reduction}).

The temporal results differ sharply from the feature-compression results. Moderate temporal reduction is possible, especially with 64 temporal components, but there is no broad low-dimensional plateau comparable to the feature axis. Performance declines substantially as the temporal representation is shortened, and reducing the sequence to 32 or 8 time points causes a clear loss. This indicates that temporal structure is not merely redundant sampling of a static segment-level representation.

The clearest result is the failure of the single-vector reductions. Mean,
max, additive-attention, and gated-attention pooling all produced near-chance
retrieval. Query-attention pooling also performed poorly despite retaining a
learned sequence of 8, 32, or 64 query outputs. Thus, the result is not
limited to the crudeness of simple averaging: both global pooling and severe
learned temporal compression discard information needed to identify the
correct audio segment. The model benefits from preserving the
within-window trajectory of the speech representation.

Together, the feature and temporal compression experiments reveal an
asymmetry in the target representation. The feature axis contains a compact
task-specific subspace that can be learned efficiently, whereas aggressive
compression of the temporal wav2vec trajectory substantially degrades
retrieval. The temporal-reduction configurations in
Figure~\ref{fig:time_reduction} are separately trained seed-42 models, so we
treat this comparison descriptively and do not infer fine-grained differences
between reduction methods. These results characterize the dependence of
retrieval on temporal compression of the target representation; they do not
estimate the temporal resolution of neural processing.

\subsection{Architectural ablations}
\label{sec:architecture_ablations}

\begin{figure}[t]
    \centering
    \includegraphics[width=\linewidth]{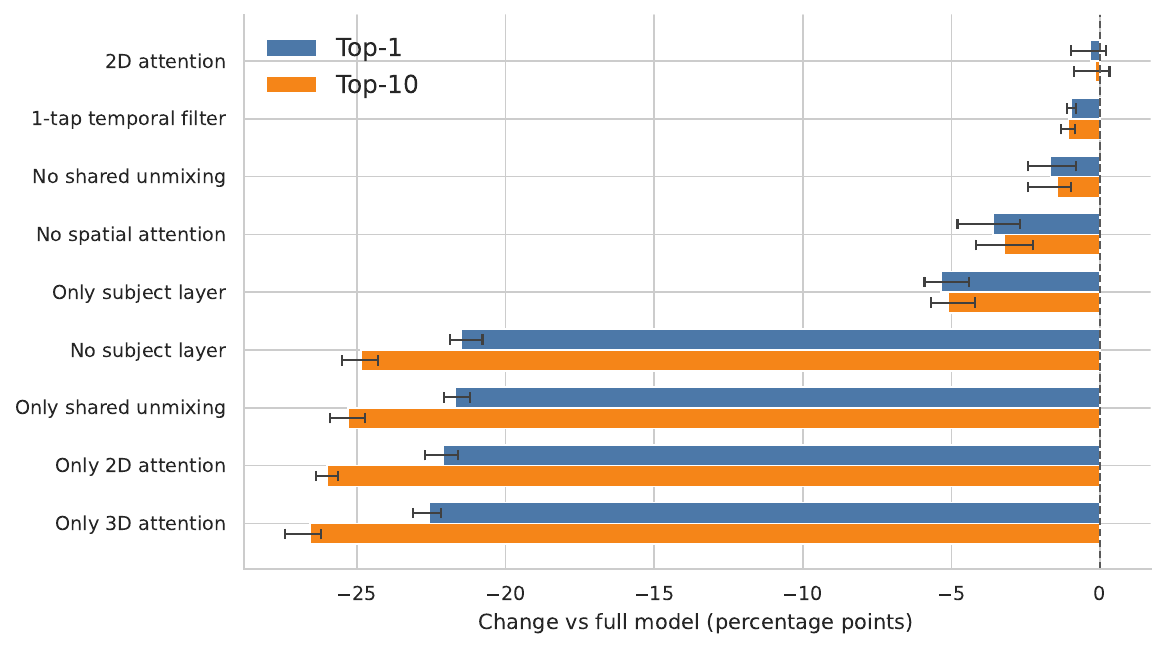}
    \caption{
    Architectural ablations of the network's front end.
    Bars show the mean change in Top-1 and Top-10 retrieval accuracy
    relative to the full model of the same seed, and whiskers span the minimum
    and maximum across the six matched seeds. Ablations that remove
    subject-conditioned spatial processing produce the largest degradations.
    }
    \label{fig:ablations}
\end{figure}

We ablated the main components of our network's front end to understand how its spatial and temporal factorization contributes to retrieval. The full model combines a geometry-aware 3D spatial attention layer, a shared unmixing stage, subject-conditioned spatial mappings, and temporal filtering in the compact branch space. The ablation results are summarized as changes relative to the full model in Fig.~\ref{fig:ablations}.
All architectural ablations shown in Figure~\ref{fig:ablations}
were repeated across six matched training seeds (42--47), with each effect
computed relative to the corresponding full model of the same seed.

The strongest separation is between models with and without
participant-specific spatial adaptation. Removing the subject-conditioned
projection from the otherwise complete front end reduced Top-1 accuracy by
\(21.50\pm0.39\) percentage points and Top-10 accuracy by
\(24.85\pm0.47\) percentage points across the six matched seeds
(mean \(\pm\) SD), leaving mean accuracies of 18.25\% and 45.55\%,
respectively. Conversely, retaining the subject-specific projection while
removing both shared spatial-attention and unmixing stages produced much
smaller reductions of \(5.35\pm0.51\) and \(5.11\pm0.51\) percentage
points. This asymmetry indicates that participant-specific spatial adaptation
is the dominant spatial component for multi-participant decoding in this
dataset. This is consistent with the fact that anatomy, head position, and
sensor--source geometry vary across participants, so a purely shared
sensor-space projection must simultaneously accommodate these differences
and extract task-relevant activity.

Within the subject-conditioned setting, the remaining ablations show
how the rest of the front end improves this aligned representation. Removing
spatial attention reduced Top-1 accuracy by \(3.59\pm0.87\) percentage
points and Top-10 accuracy by \(3.23\pm0.76\) percentage points across the
six matched seeds, indicating an additional benefit from the shared spatial
projection before participant-specific adaptation.
Replacing the 3D spherical-harmonic attention with the 2D Fourier
parameterization produced only a small change across six matched seeds:
\(-0.32\pm0.43\) percentage points in Top-1 and
\(-0.18\pm0.51\) in Top-10 accuracy relative to the 3D model
(mean \(\pm\) SD). At the common harmonic setting \(H=24\), the 3D layer
uses \(H^2=576\) trainable basis coefficients per attention channel, whereas
the 2D Fourier layer uses \(2H^2=1152\). The two parameterizations therefore
give comparable retrieval performance in this experiment, with the spherical
representation using half as many basis coefficients.

Temporal filtering provides a further gain when the spatial front end is kept fixed. Across six matched seeds, replacing the 15-sample temporal filters with learnable one-sample filters reduced Top-1 accuracy by \(0.98\pm0.14\) percentage points and Top-10 accuracy by \(1.06\pm0.17\) percentage points (mean \(\pm\) SD). The one-sample condition retains learnable instantaneous branch-wise scaling but removes temporal context, so this consistent reduction indicates that the compact branches benefit from local temporal transformations before nonlinear processing. In this sense, the temporal filters contribute to the intended spatial--temporal decomposition of the model: each branch can select not only where to look in sensor space, but also which local temporal dynamics are useful for retrieval.

The unmixing layer has a similar role in the spatial factorization.
Removing it reduced Top-1 accuracy by \(1.67\pm0.54\) percentage points and
Top-10 accuracy by \(1.43\pm0.54\) percentage points across the six matched
seeds (mean \(\pm\) SD), supporting the usefulness of an intermediate shared branch space before subject-conditioned adaptation and temporal filtering. Its effect is smaller than the losses associated with removing subject-conditioned processing, but is consistent across model initializations.

Overall, the ablations support the full network's front-end design. Subject-conditioned spatial mappings provide the final projection into the $K$ interpretable branches while accounting for cross-subject variability. The 3D spatial attention layer supplies a geometry-aware shared sensor projection before this subject-conditioned mapping, and the unmixing layer refines this shared representation by mixing the attention channels before they are projected into the branch space. Temporal filters then add branch-wise local temporal selectivity, turning the learned spatial branches into spatial--temporal components.

\subsection{Effect of temporal-filter support}
\label{ssec:temporal-filter-support}

\begin{figure}[t]
    \centering
    \includegraphics[width=0.92\linewidth]
    {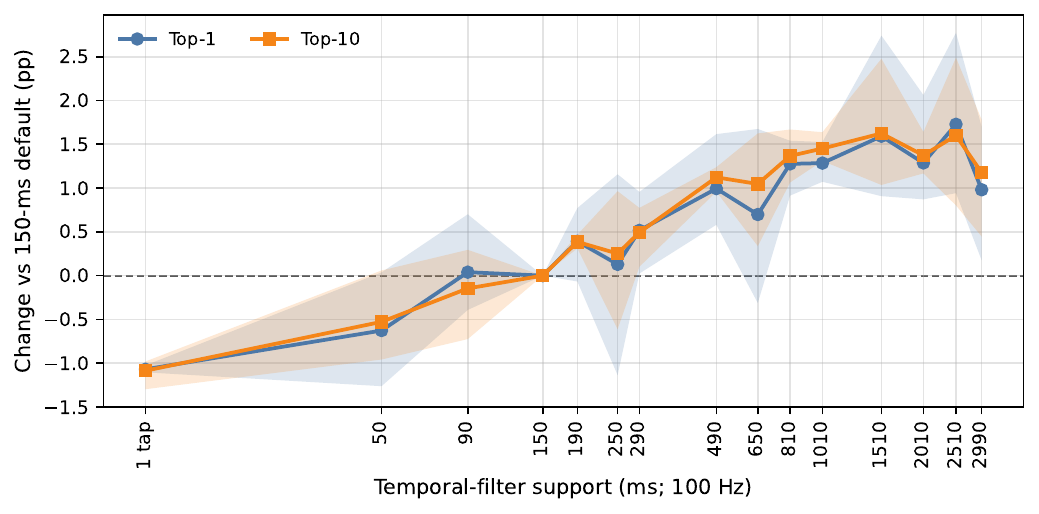}
    \caption{
        Effect of temporal-filter support on retrieval accuracy. Markers and lines show
        the mean change in Top-1 and Top-10 accuracy relative to the 150\,ms default
        across three training seeds (42--44); shaded bands show the seedwise range.
        The one-sample condition contains no temporal context and acts only as
        branch-wise scaling. Checkpoints were selected by validation loss.
    }
    \label{fig:temporal-filter-support}
\end{figure}

The one-sample ablation establishes that temporal context contributes to
retrieval, but does not show whether performance depends on a particular
temporal scale. 
We therefore varied the branch-wise filter support from one to
299 samples (10\,ms--2.99\,s at 100\,Hz), while keeping the two-block,
\(K=25\) architecture fixed and repeating the sweep across three training
seeds (42--44) (Figure~\ref{fig:temporal-filter-support}).

Retrieval improved overall as temporal-filter support increased beyond
the 150\,ms default, although the dependence was not monotonic. Across the
three seeds, the 150\,ms models averaged 39.70\% Top-1 and 70.33\% Top-10
accuracy. Removing temporal context with a one-sample filter reduced these
values by 1.07 and 1.09 percentage points, respectively. Performance
continued to improve over substantially longer supports: mean Top-1 was
highest at 2.51\,s (41.43\%, +1.73 percentage points relative to the
matched 150\,ms models), whereas mean Top-10 was highest at 1.51\,s
(71.95\%, +1.62 percentage points); the corresponding seed-42
long-support spatial--temporal analysis is reported in
Appendix~\ref{app:long-temporal-patterns}. 
At the longest tested support of
2.99\,s, both metrics declined to 40.68\% Top-1 and 71.51\% Top-10.
Thus, the dependence on temporal support is better described as a broad
long-support plateau than as a narrow optimum near 150\,ms.

\section{Discussion}
\label{sec:discussion}

We set out to investigate which functional systems may support
the decoding of perceived speech from MEG, and what in the speech stream
the decoder actually uses. 
The method links the front-end weights of a deep network
to three standard quantities: source topographies, temporal patterns,
and their spectra. 
We treat the front end as a set of branches. 
Each branch combines a participant-specific spatial filter with a
shared temporal filter, yielding a learned spatial--temporal component for
subsequent interpretation.
The outputs of these branches feed a stack of convolutional layers that turn them into a 
wav2vec 2.0 embedding for retrieval. 
Having characterized these spatial and temporal patterns, we
then intervened on the MEG input with paired feature-present and
feature-absent substitutions to determine which stimulus-associated
information supports retrieval.

\paragraph{Preprocessing.}
We removed ocular and cardiac components from the recordings before training.
This was not routine hygiene. A network trained to minimize a retrieval
loss is potentially opportunistic. 
The retrieval objective does not distinguish cortical from
stimulus-correlated peripheral signals, making the latter a possible
shortcut for speech retrieval~\cite{Geirhos2020}.
Eye activity is known to track linguistic structure and attended speech even
without corresponding visual input~\cite{Jin2018,Gehmacher2024,Braga2016},
while cardiac dynamics vary with narrative intensity and conscious narrative
processing~\cite{Wallentin2011,Perez2021}.
Left in the data, either
could have supplied stimulus-locked information about speech boundaries. 
The feature-specific occlusion effects reported above 
are therefore less susceptible to ocular and cardiac confounds.

\paragraph{A compact localizable front end.}
Our model combines competitive retrieval accuracy with a compact and
interpretable brain decoder. Across six seeds, the main configuration
($K{=}25$, two convolutional blocks) reaches $39.75 \pm 0.34\%$ Top-1
and $70.40 \pm 0.31\%$ Top-10 (mean \(\pm\) SD across model
initializations) among $1005$ candidates, while using
approximately twenty times fewer trainable parameters than the brain decoder
specified by D{\'e}fossez et al.~\cite{Defossez2023} (Appendix~\ref{app:model-size}).
D{\'e}fossez et al.
reported $41.3\%$ Top-1 and $70.7\%$ Top-10 among $1363$ candidates,
while Zhang et al.~\cite{Zhang2026} reported Top-1 accuracy between
$39.6\%$ and $41.9\%$ and Top-10 accuracy between $69.2\%$ and $71.1\%$
among $1464$ candidates. 
The published scores place our model in the same broad performance range 
but do not support a direct ranking of the architectures, because both the 
evaluation protocol and preprocessing differ. In both prior studies, each test window 
was positioned relative to a word onset, placing the onset at a fixed location within 
the window, whereas our test windows followed a fixed stride and were not aligned to 
linguistic events. This alignment may provide a consistent temporal cue absent from our 
evaluation, although the larger candidate pools used in those studies make retrieval 
more difficult. We also removed ocular and cardiac components, whereas the other 
pipelines did not report their explicit removal. Because such components may carry 
stimulus-correlated information unrelated to the neural activity of interest, this 
preprocessing difference may also affect retrieval performance. Taken together, our 
model achieves comparable accuracy under fixed-stride evaluation and explicit artifact 
removal, while using a compact decoder that exposes directly interpretable spatial and 
temporal components.

Our own ablation grid supports a controlled comparison because every
configuration in it shares the same data, preprocessing, and test set. The
$K=270$, five-block configuration is the closest tested model in branch count
and decoder depth. Despite containing 14.8 times more trainable parameters, it
scores 3.60 percentage points lower in Top-1 and 3.14 points lower in Top-10
than the main model (Appendix~\ref{app:model-size}). The remaining
architectural differences were examined separately.
The extended filter-length sweep, see
Figure~\ref{fig:temporal-filter-support}, shows that the one-tap condition,
which carries no temporal context, averages about one percentage point below
the 150\,ms default, while performance continues to improve over substantially
longer supports. Mean Top-1 is highest at 2.51\,s and mean Top-10 at 1.51\,s,
while both metrics decline at the longest tested support of 2.99\,s. The
results therefore indicate a broad long-support regime rather than saturation
around 150\,ms. The component-level front-end ablations in
Figure~\ref{fig:ablations} are evaluated separately across six matched
training seeds.

The capacity result can also be considered from a point of view more conventional in MEG neuroimaging.
Accuracy rises from a few branches to a plateau at $K \approx 10$--$25$,
and adding channels toward $K{=}270$ hurts rather than helps.
From a linear-algebra perspective, task-related M/EEG activity is known to occupy a low-dimensional subspace
of the sensor space. Dipole modeling uses this through the signal subspace~\cite{Mosher2002}, which is
defined by SVD and is therefore driven by power. Spatio-spectral
decomposition~\cite{Nikulin2011}, source-power comodulation
methods~\cite{Dahne2014}, and compact factorized networks such as EEGNet,
ShallowConvNet and LF-CNN~\cite{Lawhern2016,Schirrmeister2017,Zubarev2019}
instead find a subspace driven by information. Those networks were built
for simpler classification and regression tasks. 
Our results extend this low-dimensional view to continuous-speech
retrieval: near-maximal accuracy is obtained with only
\(K\approx10\)--\(25\) learned branch dimensions despite contrastive alignment
to a high-dimensional speech embedding.

The public MEG-MASC release contains participant-specific T1-weighted
MRI for 12 participants. Five scans had incomplete brain coverage and were
unsuitable for cortical reconstruction, leaving seven participants with usable
individual FreeSurfer anatomy. The full 27-participant analysis therefore uses
the common \texttt{fsaverage} cortical anatomy, while retaining
participant- and recording-specific sensor-to-head geometry, coregistration,
and forward solutions. As a sensitivity analysis, we compared the current
template-anatomy source estimates with estimates obtained on the individual
MRI anatomy of these seven participants and subsequently morphed to the same
\texttt{fsaverage} mesh (Appendix~\ref{app:individual-mri-validation}).
Across the original learned-filter branch maps, the participant-level mean
Pearson correlation was 0.633 (95\% CI 0.581--0.679) and the mean cosine
similarity was 0.875 (95\% CI 0.858--0.890). Peak hemisphere agreed in 83.4\%
of branches (95\% CI 80.6--86.3\%), whereas the mean displacement between
template- and individual-anatomy source-map peaks was 48.1\,mm
(95\% CI 42.4--52.6\,mm). These results support broad spatial and hemispheric
agreement, but not precise vertex- or parcel-level localization.
The cross-participant RAP-MUSIC analysis
(Figure~\ref{fig:rap-music-dipoles}) was consistent with the distributed
source-map analysis below, both emphasizing recurrent lateral cortical
structure in the learned representations.

The distributed source maps in Figure~\ref{fig:patterns} further detail the picture. 
The dominant foci lie along the superior and middle
temporal gyri, and several clusters carry additional foci anterior and superior
to these, within peri-Sylvian cortex. 
Several clusters show lateralized source distributions in either
hemisphere.
This distribution is consistent with the cortical network described for
auditory speech perception, with accounts of auditory--conceptual
integration~\cite{Bonilha2017} and predictive coding in
speech~\cite{Cope2023}, and with the dual-stream organization of speech
processing~\cite{Hickok2007}. We refrain from assigning individual clusters to
named cortical areas. 
The maps are common-template estimates built from participant-specific
sensor-to-head geometry and forward models.
Comparison with individual-MRI estimates supports the broad
lateral temporal and peri-Sylvian organization of these patterns,
but not the precise localization of individual cortical foci
(Appendix~\ref{app:individual-mri-validation}).
Across model initializations, spatial-subspace agreement is
stronger than correspondence between individual branch source maps,
while selected broad lateral temporal and peri-Sylvian motifs recur
across independently clustered models
(Appendix~\ref{app:interpretation-stability},
Figure~\ref{fig:cross-seed-motifs}). These findings support the
reproducibility of broader source-space organization rather than
the identity of individual cortical sources.
A parcellation-based assignment on individual anatomy is the natural next step.
Ciferri et al.~\cite{Ciferri2026} found a similar distribution when mapping
Whisper layers onto ECoG. The two pipelines differ in modality and in speech
model, yet they cover overlapping territory.

In the temporal patterns and spectra in Figure~\ref{fig:patterns}, low-frequency components dominate many clusters, consistent with temporal-response studies of speech tracking and with the role of slow cortical dynamics in parsing the speech envelope and phrase structure~\cite{Ding2012,Giraud2012,Gross2013,Cope2023,rogachev2024neural}. 
Several medoids also contain a distinct faster component in roughly the \(10\)--\(15\)\,Hz range. At the temporal resolution of the main 150\,ms model this is most naturally interpreted as a broad faster component rather than a narrowly localized oscillatory band. 
With 1.51\,s temporal support, the corresponding sensitivity
analysis resolves repeated secondary peaks across approximately
8--17\,Hz at about 0.66\,Hz Fourier-bin spacing, 
showing that the broad faster structure observed with the 150\,ms filters
persists at substantially finer spectral resolution
(Appendix~\ref{app:long-temporal-patterns}).
The coexistence of slow and faster temporal profiles is consistent with natural-speech studies separating envelope-related dynamics from faster activity associated with predictive and top-down processing~\cite{zioga2023naturalistic}.

\paragraph{What makes a speech segment identifiable?}
The occlusion results show that the decoder identifies speech segments from
MEG information spanning several levels of the stimulus rather than from a
single privileged linguistic representation. Retrieval depended on broad
acoustic state, rapid changes in the sound, phonetic content, and selected
contextual properties of words. Because the paired comparison held the
replaced intervals and substitution procedure constant within each feature,
these effects cannot be explained solely by generic damage caused by replacing
part of the MEG signal. They nevertheless reflect information associated with
each tested feature, rather than an independent causal contribution of that
feature, because several stimulus properties overlap and covary.

The prominent effects for silence, high loudness, and strong acoustic onsets
indicate that the decoder relies both on sustained acoustic state and on
changes that segment the continuous speech stream.
Speech pauses supply stable landmarks within a
continuous narrative, while loudness and spectral flux distinguish the
acoustic content surrounding them. Cortical tracking of speech edges and the
envelope is well established in invasive and non-invasive recordings
\cite{Oganian2019,Oganian2023,Nourski2009,Kubanek2013,Ding2012,Gross2013,Brodbeck2020},
and inserting pauses between words strengthens the relationship between the
speech envelope and neural activity~\cite{Deoisres2023}. Our intervention
results show that the decoder uses MEG information associated with both the
speech--silence state and changes within the audible signal. They should not
be read as showing that larger sound amplitude itself increases a neural
response or model sensitivity.

At the phonetic level, vowels produced the largest contrast, and positive
corrected effects were also found for stops, fricatives, sibilants, nasals,
liquids/glides, and schwa. For these analyses,
the feature-absent donor state consisted of other labelled phonemes, so the
contrast cannot be explained merely by replacing speech with silence. It
instead indicates that MEG distinctions associated with phoneme classes help
the model identify the matching audio segment. The effects nevertheless do
not form a calibrated phonetic hierarchy: the classes overlap (for example,
sibilants are a subset of fricatives and schwa is included among vowels), and
their masks occupy different fractions of the test windows.

The linguistic results were more selective. High-surprisal words produced a
clear positive effect, consistent with evidence that contextual prediction
statistics are represented during continuous speech
\cite{Heilbron2022,Gillis2023,Caucheteux2023,Goldstein2022,Schrimpf2021}.
Word onsets and pseudoword insertions also produced smaller corrected effects,
whereas predictive entropy was inconclusive after correction and lexical
rarity produced a small negative effect. 
The strongest qualitative
contrast within this group was between pseudowords and random word lists:
pseudowords produced a positive effect, whereas word-list intervals produced
a large negative effect. 
The word-list result is consistent with
sentence- and narrative-level context contributing to the neural
activity supporting retrieval, although this contrast cannot isolate
that contribution from accompanying acoustic, phonetic, and lexical
differences.

The feature-axis compression experiments complement, but do not calibrate,
these occlusion effects. Reducing the 768-dimensional wav2vec feature axis to
about a dozen dimensions with a learned linear map leaves retrieval accuracy
nearly unchanged, whereas power-driven PCA degrades much earlier. Because each reduced dimensionality was trained as a separate
model rather than obtained by projecting the reported one, the twelve-dimensional
configuration is itself a working decoder: it retains nearly all retrieval accuracy
while shrinking the convolutional head from $K\times768\times3$ to
$K\times12\times3$ parameters and reducing the
complete learned system to 436,879 trainable parameters
(Appendix~\ref{app:model-size}). The dominant reductions relative to the original
architecture nevertheless come from the branch count and the convolutional
depth rather than from the target dimension. This
indicates that MEG is aligned to a compact task-selected subspace rather than
to the full wav2vec geometry or its highest-variance directions, consistent
with reports that brain responses align to particular layers or subspaces of
self-supervised speech models~\cite{Millet2022,Vaidya2022,Ciferri2026}. The
temporal axis behaves differently: global pooling of the within-segment
trajectory reduces retrieval to near chance. Together, the occlusion and
compression analyses indicate that the
model combines temporally structured acoustic, phonetic, and contextual
information within a compact learned target subspace.

\paragraph{Interpretable architectural choices.}
The front-end ablations, see Figure~\ref{fig:ablations}, connect these
findings to design choices that D{\'e}fossez et al.~\cite{Defossez2023}
introduced for engineering reasons and that we kept in an interpretable
form. 
The six-seed ablations identify participant-specific spatial
adaptation as the dominant spatial component: removing it reduces Top-1 and
Top-10 accuracy by approximately 21.5 and 24.9 percentage points,
respectively, whereas retaining the participant-specific projection as the
only spatial stage incurs losses of only approximately 5.4 and 5.1 points.
This is consistent with the need to accommodate differences in anatomy, head
position, and sensor--source geometry across participants
\cite{mellot2024geodesic}. The shared spatial-attention stage provides a
smaller additional benefit, with its removal reducing retrieval by
approximately 3--4 percentage points.
The spherical and planar spatial parameterizations yield comparable
retrieval performance across seeds. At \(H=24\), however, the spherical
representation requires 576 basis coefficients per attention channel rather
than 1152 for the planar Fourier representation. The full sensor-space rank
of the spherical basis further shows that, at the order used here, this
parameterization does not reduce the dimensionality of the sampled spatial
filter space. Its practical role is therefore to provide an 
inductive bias by expressing the spatial weights
in coordinates adapted to the approximately spherical sensor
layout~\cite{Sivakumar2016}, while doing so with fewer basis coefficients than
the planar parameterization used for comparison.
Retrieval continues to improve over temporal supports substantially
longer than 150\,ms, with mean Top-1 highest at 2.51\,s and mean Top-10 at
1.51\,s. Both metrics decline at the longest tested support of 2.99\,s,
indicating a broad long-support regime rather than saturation around the
150\,ms default.
Their main value
here, though, is interpretive. They provide access to the
second-order temporal structure selected by each branch, complementing its
spatial topography and enabling the joint spatial--temporal characterization
used above.

\paragraph{What retrieval accuracy actually measures.}
Taken together, the results argue for treating retrieval accuracy as a
composite score rather than as a direct assay of linguistic content. A model
can reach the operating range of D{\'e}fossez et al. by combining a compact
peri-Sylvian spatial subspace with MEG distinctions associated with silence,
sound intensity, acoustic onsets, phoneme classes, and contextual surprisal,
all aligned to a thin task-selected subspace of a pretrained speech embedding
whose temporal trajectory must remain intact. This combination fits decades
of speech-envelope tracking and predictive-processing work, but it is not the
same as reading out phonemes one by one, or meaning alone, from single trials
of MEG.
Replacing MEG
recorded during random word lists with MEG from coherent narrative improved
retrieval (\(\Delta r=-17.77\)), whereas pseudoword intervals showed a
positive occlusion effect. Together, these results suggest that retrieval-relevant MEG
reflects not only local acoustic and lexical information, but
potentially also the broader linguistic context in which speech
unfolds.

The strong acoustic-state effects and the failure of heavy temporal
compression suggest one reason audio-aligned objectives may be advantageous
relative to text-aligned objectives on non-invasive data:
the audio target preserves pause structure and fine temporal variation that
MEG carries, whereas token-level representations remove much of that direct
alignment. Importantly, the interpretable architecture does not lower the
headline score here; it changes what that score can be taken to mean.

This matters because the two available alternatives leave a gap. Most
high-performance deep networks optimize accuracy and leave the decision
rule opaque; their weights do not translate into cortical sources and the dynamics of their activity. Classical neuroimaging does identify the
neural substrate of behavior-specific states, but usually in a non-ecological setting and with models too
simple to capture the subtle variations that identify segments in the natural
continuous speech. An interpretable front end closes that gap from both
sides. It keeps a compact decoder that matches current retrieval
benchmarks, while returning branch-wise topographies and spectra that can
be read as physiologically meaningful structure.

With individual anatomy and appropriate source-recovery validation,
the same pipeline could become a practical tool for time-resolved imaging
and knowledge discovery from multichannel neurophysiological data collected
during complex, ecologically valid tasks.

\section{Limitations}
\label{sec:limitations}

Our analyses used a single MEG corpus~\cite{Gwilliams2023} with 27
participants. The decoder is participant-dependent and was not
evaluated on previously unseen participants. The participant sign-flip test
provides population-level inference for the feature-use effects across
participants conditional on the fixed held-out audio material; it does not
establish that the same feature-use pattern will replicate for new narratives.
Repeating the paired occlusion analysis across six independently
trained instances of the same architecture showed that the principal
feature-use effects were robust to model initialization
(Appendix~\ref{app:occlusion-seed-robustness}). Population-level inference is
based on participant variability in the primary seed-42 analysis; the
additional seeds quantify variability across trained model solutions.

Interpretation of the front end assumes linear space--time factorized
processing in the first layer. This may miss more complex source dynamics,
such as space--time inseparable cortical waves. Although a plurality of
branches can represent some propagating activity, the present interpretation
cannot establish the role of cortical waves \textit{per se}.

The frequency profiles of learned temporal patterns
characterize the decoder's temporal selectivity. Their
correspondence with frequency ranges associated with speech
processing motivates the physiological interpretations discussed
above, although the spectra alone do not identify specific
neural oscillations or predictive-processing mechanisms.

The paired substitutions use real MEG from the same participant and session,
and the feature-present arm controls for generic replacement damage, but the
intervention does not isolate a strictly independent causal contribution of
each stimulus variable. Several annotations overlap or covary (Appendix~\ref{app:feature-mask-overlap}), their masks differ in duration and eligible-window coverage, and donor replacement may
alter correlated stimulus states. The reported rank contrasts should therefore be interpreted as evidence that
the trained decoder uses MEG information distinguishing the feature-present
and feature-absent states, rather than as directly comparable estimates of
neural encoding strength across features. 
The common 150\,ms MEG--audio alignment also does not resolve
feature-specific response latency, which is particularly relevant for lexical
and contextual responses that may occur several hundred milliseconds after
word onset~\cite{Broderick2018,Weissbart2020}.

Both the RAP-MUSIC analysis and the clustered spatial--temporal profiles use participant-specific sensor-to-head geometry and forward models on the common \texttt{fsaverage} anatomy. For RAP-MUSIC, the pooled locations summarize the localization of the same shared cross-participant SVD subspace under the 27 participant-specific forward models. An individual-MRI sensitivity analysis in the seven participants with usable structural scans showed substantial agreement in the broad cortical distributions (mean Pearson \(r=0.633\), mean cosine similarity \(=0.875\)) but appreciable variation in the precise source-map maxima (Appendix~\ref{app:individual-mri-validation}). 
Fine-grained anatomical localization should therefore be interpreted
as approximate. These inverse-model analyses provide anatomical summaries of
the learned sensor-space patterns rather than independent validation of
precise neuronal source identity.

\section*{Acknowledgements}
This work is an output of a research project 
HSE-BR-2025-26 implemented as part of the Basic Research Program at HSE University. 
This research was supported in part through computational resources of HPC facilities at HSE University. The authors thank the contributors of the MEG-MASC dataset~\cite{Gwilliams2023} for making the data publicly available. The authors are grateful to  Alexey Voskoboynikov for the original version of the code partly reproducing the results of \cite{Defossez2023}. 

\section*{Data and code availability}
The MEG-MASC dataset is publicly available~\cite{Gwilliams2023}. 
The code used in this study is publicly available at
\href{https://github.com/ivsemenkov/LISA/}{\nolinkurl{https://github.com/ivsemenkov/LISA/}}.
ICA components required to clean the MEG-MASC dataset are available at
\href{https://osf.io/3wrft/}{\nolinkurl{https://osf.io/3wrft/}}.

\section*{Ethics statement}
This study performed a secondary analysis of the publicly available
MEG-MASC dataset~\cite{Gwilliams2023} and involved no new participant
recruitment or data collection. As reported by the dataset authors, all
participants provided written informed consent, and the original data
collection was approved by the Institutional Review Board (IRB) of
New York University Abu Dhabi.
The original approval date (16 April 2012) and protocol
number (\#003-2012A) were subsequently confirmed through correspondence
with the IRB.
The authors of the current manuscript declare no competing interests.

\section*{Declaration of generative AI and AI-assisted technologies in the manuscript preparation process}

During the preparation of this work, the authors used LLMs for stylistic editing and computer code development assistance. The authors reviewed and edited the output as needed and take full responsibility for the content of the published article. 

\bibliographystyle{unsrtnat}
\bibliography{references} 

@article{Geirhos2020,
  author  = {Geirhos, Robert and Jacobsen, J{\"o}rn-Henrik and Michaelis, Claudio
             and Zemel, Richard and Brendel, Wieland and Bethge, Matthias
             and Wichmann, Felix A.},
  title   = {Shortcut learning in deep neural networks},
  journal = {Nature Machine Intelligence},
  volume  = {2},
  number  = {11},
  pages   = {665--673},
  year    = {2020},
  doi     = {10.1038/s42256-020-00257-z}
}

@article{Zhang2026,
  author  = {Zhang, Xinyu and Liu, Sichao and Lu, Runhao and
             Woolgar, Alexandra and Wang, Lihui},
  title   = {What Are We Actually Decoding? Source Attribution for
             Non-Invasive Brain-to-Language Retrieval},
  journal = {arXiv preprint arXiv:2605.24524},
  year    = {2026}
}

@article{dAscoli2025,
  author  = {d'Ascoli, St{\'e}phane and Bel, Corentin and Rapin, J{\'e}r{\'e}my
             and Banville, Hubert and Benchetrit, Yohann and Pallier, Christophe
             and King, Jean-R{\'e}mi},
  title   = {Towards decoding individual words from non-invasive brain recordings},
  journal = {Nature Communications},
  volume  = {16},
  pages   = {10521},
  year    = {2025},
  doi     = {10.1038/s41467-025-65499-0}
}

@article{Nourmohammadi2023,
  author  = {Nourmohammadi, Amin and Swift, James R. and de Pesters, Adriana and Guay, Christian S. and Adamo, Matthew A. and Dalfino, John C. and Ritaccio, Anthony L. and Schalk, Gerwin and Brunner, Peter},
  title   = {Passive functional mapping of receptive language cortex during general anesthesia using electrocorticography},
  journal = {Clinical Neurophysiology},
  year    = {2023},
  volume  = {147},
  pages   = {31--44},
  doi     = {10.1016/j.clinph.2022.11.021}
}

@article{Taplin2016,
  author  = {Taplin, AmiLyn M. and de Pesters, Adriana and Brunner, Peter and Hermes, Dora and Dalfino, John C. and Adamo, Matthew A. and Ritaccio, Anthony L. and Schalk, Gerwin},
  title   = {Intraoperative mapping of expressive language cortex using passive real-time electrocorticography},
  journal = {Epilepsy \& Behavior Case Reports},
  year    = {2016},
  volume  = {5},
  pages   = {46--51},
  doi     = {10.1016/j.ebcr.2016.03.003}
}

@incollection{Bisla2025,
title = {A comprehensive review on state-of-the-art imagined speech decoding techniques using electroencephalography},
editor = {M.A. Ansari and R.S. Anand and Pragati Tripathi and Rajat Mehrotra and Md Belal Bin Heyat},
booktitle = {Artificial Intelligence in Biomedical and Modern Healthcare Informatics},
publisher = {Academic Press},
pages = {101--126},
year = {2025},
isbn = {978-0-443-21870-5},
doi = {10.1016/B978-0-443-21870-5.00011-X},
author = {Meenakshi Bisla and R.S. Anand},
}

@article{Card2024,
  title={An accurate and rapidly calibrating speech neuroprosthesis},
  author={Card, Nicholas S and Wairagkar, Maitreyee and Iacobacci, Carrina and Hou, Xianda and Singer-Clark, Tyler and Willett, Francis R and Kunz, Erin M and Fan, Chaofei and Vahdati Nia, Maryam and Deo, Darrel R and others},
  journal={New England Journal of Medicine},
  volume={391},
  number={7},
  pages={609--618},
  year={2024},
  publisher={Mass Medical Soc}
}

@article{Friederici2011,
  title={The brain basis of language processing: from structure to function},
  author={Friederici, Angela D},
  journal={Physiological Reviews},
  volume={91},
  number={4},
  pages={1357--1392},
  year={2011},
  doi={10.1152/physrev.00006.2011}
}

@book{Penfield1959,
  title={Speech and Brain Mechanisms},
  author={Penfield, Wilder and Roberts, Lamar},
  year={1959},
  publisher={Princeton University Press},
  address={Princeton, NJ}
}

@article{Geschwind1970,
  title={The organization of language and the brain},
  author={Geschwind, Norman},
  journal={Science},
  volume={170},
  number={3961},
  pages={940--944},
  year={1970},
  doi={10.1126/science.170.3961.940}
}

@inproceedings{mellot2024geodesic,
  title={Geodesic optimization for predictive shift adaptation on {EEG} data},
  author={Mellot, Apolline and Collas, Antoine and Chevallier, Sylvain and Gramfort, Alexandre and Engemann, Denis A},
  booktitle={Advances in Neural Information Processing Systems},
  volume={37},
  pages={32828--32855},
  year={2024}
}

@article{kutas1980reading,
  title={Reading senseless sentences: Brain potentials reflect semantic incongruity},
  author={Kutas, Marta and Hillyard, Steven A},
  journal={Science},
  volume={207},
  number={4427},
  pages={203--205},
  year={1980},
  publisher={American Association for the Advancement of Science}
}

@article{lau2008cortical,
  title={A cortical network for semantics: (de)constructing the {N400}},
  author={Lau, Ellen F and Phillips, Colin and Poeppel, David},
  journal={Nature Reviews Neuroscience},
  volume={9},
  number={12},
  pages={920--933},
  year={2008},
}

@article{rogachev2024neural,
  title={Neural tracking of natural speech in children in relation to their receptive speech abilities},
  author={Rogachev, Anton and Sysoeva, Olga},
  journal={Cognitive Systems Research},
  volume={86},
  pages={101236},
  year={2024},
  publisher={Elsevier}
}

@article{zioga2023naturalistic,
  title={Naturalistic spoken language comprehension is supported by alpha and beta oscillations},
  author={Zioga, Ioanna and Weissbart, Hugo and Lewis, Ashley G and Haegens, Saskia and Martin, Andrea E},
  journal={Journal of Neuroscience},
  volume={43},
  number={20},
  pages={3718--3732},
  year={2023},
  publisher={Society for Neuroscience}
}

@article{Hickok2007,
  title={The cortical organization of speech processing},
  author={Hickok, Gregory and Poeppel, David},
  journal={Nature Reviews Neuroscience},
  volume={8},
  number={5},
  pages={393--402},
  year={2007}
}

@article{Ding2012,
  title={Emergence of neural encoding of auditory objects while listening to competing speakers},
  author={Ding, Nai and Simon, Jonathan Z},
  journal={Proceedings of the National Academy of Sciences},
  volume={109},
  number={29},
  pages={11854--11859},
  year={2012}
}

@article{Gross2013,
  title={Speech rhythms and multiplexed oscillatory sensory coding in the human brain},
  author={Gross, Joachim and Hoogenboom, Nienke and Thut, Gregor and Schyns, Philippe and Panzeri, Stefano and Belin, Pascal and Garrod, Simon},
  journal={PLoS Biology},
  volume={11},
  number={12},
  pages={e1001752},
  year={2013}
}

@article{Giraud2012,
  title={Cortical oscillations and speech processing: emerging computational principles and operations},
  author={Giraud, Anne-Lise and Poeppel, David},
  journal={Nature Neuroscience},
  volume={15},
  number={4},
  pages={511--517},
  year={2012}
}

@article{Oganian2019,
author = {Yulia Oganian  and Edward F. Chang },
title = {A speech envelope landmark for syllable encoding in human superior temporal gyrus},
journal = {Science Advances},
volume = {5},
number = {11},
pages = {eaay6279},
year = {2019},
doi = {10.1126/sciadv.aay6279},
}

@article {Oganian2023,
	author = {Oganian, Yulia and Kojima, Katsuaki and Breska, Assaf and Cai, Chang and Findlay, Anne and Chang, Edward F. and Nagarajan, Srikantan S.},
	title = {Phase Alignment of Low-Frequency Neural Activity to the Amplitude Envelope of Speech Reflects Evoked Responses to Acoustic Edges, Not Oscillatory Entrainment},
	volume = {43},
	number = {21},
	pages = {3909--3921},
	year = {2023},
	doi = {10.1523/JNEUROSCI.1663-22.2023},
	publisher = {Society for Neuroscience},
	issn = {0270-6474},
	journal = {Journal of Neuroscience}
}

@article{Nourski2009,
  title={Temporal envelope of time-compressed speech represented in the human auditory cortex},
  author={Nourski, Kirill V and Reale, Richard A and Oya, Hiroyuki and Kawasaki, Hiroto and Kovach, Christopher K and Chen, Haiming and Howard, Matthew A and Brugge, John F},
  journal={Journal of Neuroscience},
  volume={29},
  number={49},
  pages={15564--15574},
  year={2009}
}

@article{Kubanek2013,
  title={The tracking of speech envelope in the human cortex},
  author={Kubanek, Jan and Brunner, Peter and Gunduz, Aysegul and Poeppel, David and Schalk, Gerwin},
  journal={PLoS One},
  volume={8},
  number={1},
  pages={e53398},
  year={2013}
}

@article{Brodbeck2020,
  title={Continuous speech processing},
  author={Brodbeck, Christian and Simon, Jonathan Z},
  journal={Current Opinion in Physiology},
  volume={18},
  pages={25--31},
  year={2020}
}

@article{Deoisres2023,
    doi = {10.1371/journal.pone.0289288},
    author = {Deoisres, Suwijak AND Lu, Yuhan AND Vanheusden, Frederique J. AND Bell, Steven L. AND Simpson, David M.},
    journal = {PLOS ONE},
    publisher = {Public Library of Science},
    title = {Continuous speech with pauses inserted between words increases cortical tracking of speech envelope},
    year = {2023},
    month = {07},
    volume = {18},
    pages = {e0289288},
    number = {7},
}

@article{Heilbron2022,
  title={A hierarchy of linguistic predictions during natural language comprehension},
  author={Heilbron, Micha and Armeni, Kristijan and Schoffelen, Jan-Mathijs and Hagoort, Peter and de Lange, Floris P},
  journal={Proceedings of the National Academy of Sciences},
  volume={119},
  number={32},
  pages={e2201968119},
  year={2022}
}

@article {Gillis2023,
	author = {Gillis, Marlies and Vanthornhout, Jonas and Francart, Tom},
	title = {Heard or Understood? Neural Tracking of Language Features in a Comprehensible Story, an Incomprehensible Story and a Word List},
	volume = {10},
	number = {7},
	year = {2023},
    pages = {ENEURO.0075-23.2023},
	doi = {10.1523/ENEURO.0075-23.2023},
	publisher = {Society for Neuroscience},
	journal = {eNeuro}
}

@article{Schrimpf2021,
  title={The neural architecture of language: Integrative modeling converges on predictive processing},
  author={Schrimpf, Martin and Blank, Idan Asher and Tuckute, Greta and Kauf, Carina and Hosseini, Eghbal A and Kanwisher, Nancy and Tenenbaum, Joshua B and Fedorenko, Evelina},
  journal={Proceedings of the National Academy of Sciences},
  volume={118},
  number={45},
  pages={e2105646118},
  year={2021}
}

@inproceedings{Millet2022,
 author = {Millet, Juliette and Caucheteux, Charlotte and Orhan, Pierre and Boubenec, Yves and Gramfort, Alexandre and Dunbar, Ewan and Pallier, Christophe and King, Jean-R{\'e}mi},
 booktitle = {Advances in Neural Information Processing Systems},
 doi = {10.52202/068431-2422},
 editor = {S. Koyejo and S. Mohamed and A. Agarwal and D. Belgrave and K. Cho and A. Oh},
 pages = {33428--33443},
 publisher = {Curran Associates, Inc.},
 title = {Toward a realistic model of speech processing in the brain with self-supervised learning},
 volume = {35},
 year = {2022}
}

@InProceedings{Vaidya2022,
  title = 	 {Self-Supervised Models of Audio Effectively Explain Human Cortical Responses to Speech},
  author =       {Vaidya, Aditya R and Jain, Shailee and Huth, Alexander},
  booktitle = 	 {Proceedings of the 39th International Conference on Machine Learning},
  pages = 	 {21927--21944},
  year = 	 {2022},
  editor = 	 {Chaudhuri, Kamalika and Jegelka, Stefanie and Song, Le and Szepesvari, Csaba and Niu, Gang and Sabato, Sivan},
  volume = 	 {162},
  series = 	 {Proceedings of Machine Learning Research},
  month = 	 {17--23 Jul},
  publisher =    {PMLR},
}

@article{LehnSchioler2024,
      title={Mechanistic Interpretability of {EEG} Foundation Models via Sparse Autoencoders}, 
      author={William Lehn-Schi{\o}ler and Magnus Ruud Kj{\ae}r and Rahul Thapa and Magnus Guldberg Pedersen and Anton Mosquera Storgaard and Nick Williams and Radu Gatej and Tue Lehn-Schi{\o}ler and Andreas Brink-Kj{\ae}r and Sadasivan Puthusserypady and S{\'a}ndor Beniczky and James Zou and Lars Kai Hansen},
      year={2026},
      journal={arXiv preprint arXiv:2605.13930},
      eprint={2605.13930},
      archivePrefix={arXiv},
      primaryClass={cs.LG},
}

@article{Hammer2022,
doi = {10.1088/1741-2552/ac6770},
year = {2022},
month = {may},
publisher = {IOP Publishing},
volume = {19},
number = {3},
pages = {036006},
author = {Hammer, J and Schirrmeister, R T and Hartmann, K and Marusic, P and Schulze-Bonhage, A and Ball, T},
title = {Interpretable functional specialization emerges in deep convolutional networks trained on brain signals},
journal = {Journal of Neural Engineering}
}

@article{Maghsoudi2025,
      title={Mechanistic Interpretability of Brain-to-Speech Models Across Speech Modes}, 
      author={Maryam Maghsoudi and Ayushi Mishra},
      year={2026},
      journal={arXiv preprint arXiv:2602.01247},
      eprint={2602.01247},
      archivePrefix={arXiv},
      primaryClass={cs.LG}
}

@article{Ciferri2026,
      title={Mapping {Whisper} Representations to Human {ECoG} Responses with Interpretable Time-Resolved Neural Encoding}, 
      author={Matteo Ciferri and Tommaso Boccato and Michal Olak and Matteo Ferrante and Nicola Toschi},
      year={2026},
      journal={arXiv preprint arXiv:2606.02305},
      eprint={2606.02305},
      archivePrefix={arXiv},
      primaryClass={q-bio.NC}
}

@article{ossadtchi2024representational,
  title={Representational dissimilarity component analysis ({ReDisCA})},
  author={Ossadtchi, Alexei and Semenkov, Ilia and Zhuravleva, Anna and Kozunov, Vladimir and Serikov, Oleg and Voloshina, Ekaterina},
  journal={NeuroImage},
  volume={301},
  pages={120868},
  year={2024},
  publisher={Elsevier}
}

@article{hecker2026invertmeeg,
	author = {Hecker, Lukas},
	title = {invertmeeg: A Benchmark and Unified {Python} Library for {EEG} Inverse Solvers},
	year = {2026},
	doi = {10.64898/2026.03.06.710103},
	publisher = {Cold Spring Harbor Laboratory},
	journal = {bioRxiv}
}

@article{Lawhern2016,
  title={{EEGNet}: a compact convolutional neural network for {EEG}-based brain--computer interfaces},
  author={Lawhern, Vernon J and Solon, Amelia J and Waytowich, Nicholas R and Gordon, Stephen M and Hung, Chou P and Lance, Brent J},
  journal={Journal of Neural Engineering},
  volume={15},
  number={5},
  pages={056013},
  year={2018},
  publisher={IOP Publishing}
}

@article{Schirrmeister2017,
  title={Deep learning with convolutional neural networks for {EEG} decoding and visualization},
  author={Schirrmeister, Robin Tibor and Springenberg, Jost Tobias and Fiederer, Lukas Dominique Josef and Glasstetter, Martin and Eggensperger, Katharina and Tangermann, Michael and Hutter, Frank and Burgard, Wolfram and Ball, Tonio},
  journal={Human Brain Mapping},
  volume={38},
  number={11},
  pages={5391--5420},
  year={2017},
  publisher={Wiley Online Library}
}

@article{Waytowich2018,
  title={Compact convolutional neural networks for classification of asynchronous steady-state visual evoked potentials},
  author={Waytowich, Nicholas and Lawhern, Vernon J and Garcia, Javier O and Cummings, Jennifer and Faller, Josef and Sajda, Paul and Vettel, Jean M},
  journal={Journal of Neural Engineering},
  volume={15},
  number={6},
  pages={066031},
  year={2018},
  publisher={IOP Publishing}
}

@article{Zubarev2019,
title = {Adaptive neural network classifier for decoding {MEG} signals},
journal = {NeuroImage},
volume = {197},
pages = {425--434},
year = {2019},
issn = {1053-8119},
doi = {10.1016/j.neuroimage.2019.04.068},
author = {Ivan Zubarev and Rasmus Zetter and Hanna-Leena Halme and Lauri Parkkonen}
}

@inproceedings{Baevski2020,
  title     = {wav2vec 2.0: A framework for self-supervised learning of speech representations},
  author    = {Baevski, Alexei and Zhou, Yuhao and Mohamed, Abdelrahman and Auli, Michael},
  booktitle = {Advances in Neural Information Processing Systems},
  volume    = {33},
  pages     = {12449--12460},
  year      = {2020}
}

@article{Bonilha2017,
  title   = {Temporal lobe networks supporting the comprehension of spoken words},
  author  = {Bonilha, Leonardo and Hillis, Argye E and Hickok, Gregory and den Ouden, Dirk B and Rorden, Chris and Fridriksson, Julius},
  journal = {Brain},
  volume  = {140},
  number  = {9},
  pages   = {2370--2380},
  year    = {2017}
}

@article{Braga2016,
  title   = {Eye movements during auditory attention predict individual differences in dorsal attention network activity},
  author  = {Braga, Rodrigo M and Fu, Richard Z and Seemungal, Barry M and Wise, Richard JS and Leech, Robert},
  journal = {Frontiers in Human Neuroscience},
  volume  = {10},
  pages   = {164},
  year    = {2016}
}

@book{Buzsaki2006,
  title     = {Rhythms of the Brain},
  author    = {Buzs{\'a}ki, Gy{\"o}rgy},
  publisher = {Oxford University Press},
  year      = {2006}
}

@article{Caucheteux2023,
  title   = {Evidence of a predictive coding hierarchy in the human brain listening to speech},
  author  = {Caucheteux, Charlotte and Gramfort, Alexandre and King, Jean-R{\'e}mi},
  journal = {Nature Human Behaviour},
  volume  = {7},
  number  = {3},
  pages   = {430--441},
  year    = {2023}
}

@article{Cohen1968,
  title   = {Magnetoencephalography: evidence of magnetic fields produced by alpha-rhythm currents},
  author  = {Cohen, David},
  journal = {Science},
  volume  = {161},
  number  = {3843},
  pages   = {784--786},
  year    = {1968}
}

@article{Cope2023,
  title   = {Temporal lobe perceptual predictions for speech are instantiated in motor cortex and reconciled by inferior frontal cortex},
  author  = {Cope, Thomas E. and Sohoglu, Ediz and Peterson, Katie A. and Jones, P. Simon and Rua, Catarina and Passamonti, Luca and Sedley, William and Post, Brechtje and Coebergh, Jan and Butler, Christopher R. and Garrard, Peter and Abdel-Aziz, Khaled and Husain, Masud and Griffiths, Timothy D. and Patterson, Karalyn and Davis, Matthew H. and Rowe, James B.},
  journal = {Cell Reports},
  volume  = {42},
  number  = {5},
  pages   = {112422},
  year    = {2023},
  doi     = {10.1016/j.celrep.2023.112422}
}

@article{Dahne2014,
  title   = {{SPoC}: a novel framework for relating the amplitude of neuronal oscillations to behaviorally relevant parameters},
  author  = {D{\"a}hne, Sven and Meinecke, Frank C and Haufe, Stefan and H{\"o}hne, Johannes and Tangermann, Michael and M{\"u}ller, Klaus-Robert and Nikulin, Vadim V},
  journal = {NeuroImage},
  volume  = {86},
  pages   = {111--122},
  year    = {2014}
}

@article{Defossez2023,
  title   = {Decoding speech perception from non-invasive brain recordings},
  author  = {D{\'e}fossez, Alexandre and Caucheteux, Charlotte and Rapin, J{\'e}r{\'e}my and Kabeli, Ori and King, Jean-R{\'e}mi},
  journal = {Nature Machine Intelligence},
  volume  = {5},
  number  = {10},
  pages   = {1097--1107},
  year    = {2023}
}

@article{Eliades2008,
  title   = {Neural substrates of vocalization feedback monitoring in primate auditory cortex},
  author  = {Eliades, Steven J and Wang, Xiaoqin},
  journal = {Nature},
  volume  = {453},
  number  = {7198},
  pages   = {1102--1106},
  year    = {2008}
}

@article{Fadiga2002,
  title   = {Speech listening specifically modulates the excitability of tongue muscles: a {TMS} study},
  author  = {Fadiga, Luciano and Craighero, Laila and Buccino, Giovanni and Rizzolatti, Giacomo},
  journal = {European Journal of Neuroscience},
  volume  = {15},
  number  = {2},
  pages   = {399--402},
  year    = {2002}
}

@article{Fischl2012,
  title   = {{FreeSurfer}},
  author  = {Fischl, Bruce},
  journal = {NeuroImage},
  volume  = {62},
  number  = {2},
  pages   = {774--781},
  year    = {2012}
}

@article{Gehmacher2024,
  title   = {Eye movements track prioritized auditory features in selective attention to natural speech},
  author  = {Gehmacher, Quirin and Schubert, Juliane and Schmidt, Fabian and Hartmann, Thomas and Reisinger, Patrick and R{\"o}sch, Sebastian and Schwarz, Konrad and Popov, Tzvetan and Chait, Maria and Weisz, Nathan},
  journal = {Nature Communications},
  volume  = {15},
  number  = {1},
  pages   = {3692},
  year    = {2024}
}

@article{Goldstein2022,
  title   = {Shared computational principles for language processing in humans and deep language models},
  author  = {Goldstein, Ariel and Zada, Zaid and Buchnik, Eliav and Schain, Mariano and Price, Amy and Aubrey, Bobbi and Nastase, Samuel A and Feder, Amir and Emanuel, Dotan and Cohen, Alon and others},
  journal = {Nature Neuroscience},
  volume  = {25},
  number  = {3},
  pages   = {369--380},
  year    = {2022}
}

@book{GolubVanLoan2013,
  title     = {Matrix Computations},
  author    = {Golub, Gene H and Van Loan, Charles F},
  publisher = {Johns Hopkins University Press},
  year      = {2013}
}

@article{Gramfort2013,
  title   = {{MEG} and {EEG} data analysis with {MNE}-{P}ython},
  author  = {Gramfort, Alexandre and Luessi, Martin and Larson, Eric and Engemann, Denis A and Strohmeier, Daniel and Brodbeck, Christian and Goj, Roman and Jas, Mainak and Brooks, Teon and Parkkonen, Lauri and H{\"a}m{\"a}l{\"a}inen, Matti},
  journal = {Frontiers in Neuroscience},
  volume  = {7},
  pages   = {267},
  year    = {2013}
}

@article{Gwilliams2023,
  title   = {Introducing {MEG}-{MASC}: a high-quality magneto-encephalography dataset for evaluating natural speech processing},
  author  = {Gwilliams, Laura and Flick, Graham and Marantz, Alec and Pylkk{\"a}nen, Liina and Poeppel, David and King, Jean-R{\'e}mi},
  journal = {Scientific Data},
  volume  = {10},
  number  = {1},
  pages   = {862},
  year    = {2023}
}

@article{Haufe2014,
  title   = {On the interpretation of weight vectors of linear models in multivariate neuroimaging},
  author  = {Haufe, Stefan and Meinecke, Frank and G{\"o}rgen, Kai and D{\"a}hne, Sven and Haynes, John-Dylan and Blankertz, Benjamin and Bie{\ss}mann, Felix},
  journal = {NeuroImage},
  volume  = {87},
  pages   = {96--110},
  year    = {2014}
}

@article{Jin2018,
  title   = {Eye activity tracks task-relevant structures during speech and auditory sequence perception},
  author  = {Jin, Peiqing and Zou, Jiajie and Zhou, Tao and Ding, Nai},
  journal = {Nature Communications},
  volume  = {9},
  number  = {1},
  pages   = {5374},
  year    = {2018}
}

@book{Kay1993,
  title     = {Fundamentals of statistical signal processing: estimation theory},
  author    = {Kay, Steven M},
  publisher = {Prentice Hall},
  year      = {1993}
}

@article{Liu2023,
  title   = {Neural circuits underlying language control and modality control in bilinguals: An {fMRI} study},
  author  = {Liu, Huanhuan and Guo, Zibin and Jiang, Yishan and Schwieter, John W and Wang, Fenqi},
  journal = {Neuropsychologia},
  volume  = {178},
  pages   = {108430},
  year    = {2023}
}

@inproceedings{Loshchilov2019,
title={Decoupled Weight Decay Regularization},
author={Ilya Loshchilov and Frank Hutter},
booktitle={International Conference on Learning Representations},
year={2019},
}

@ARTICLE{Mosher2002,
  author={Mosher, J.C. and Leahy, R.M.},
  journal={IEEE Transactions on Signal Processing}, 
  title={Source localization using recursively applied and projected ({RAP}) {MUSIC}}, 
  year={1999},
  volume={47},
  number={2},
  pages={332--340},
  doi={10.1109/78.740118}}

@article{Nikulin2011,
  title   = {A novel method for reliable and fast extraction of neuronal {EEG}/{MEG} oscillations on the basis of spatio-spectral decomposition},
  author  = {Nikulin, Vadim V and Nolte, Guido and Curio, Gabriel},
  journal = {NeuroImage},
  volume  = {55},
  number  = {4},
  pages   = {1528--1535},
  year    = {2011}
}

@article{Petrosyan2021,
  title   = {Decoding and interpreting cortical signals with a compact convolutional neural network},
  author  = {Petrosyan, Artur and Sinkin, Mikhail and Lebedev, Mikhail and Ossadtchi, Alexei},
  journal = {Journal of Neural Engineering},
  volume  = {18},
  number  = {2},
  pages   = {026019},
  year    = {2021}
}

@article{Petrosyan2022,
  title   = {Speech decoding from a small set of spatially segregated minimally invasive intracranial {EEG} electrodes with a compact and interpretable neural network},
  author  = {Petrosyan, Artur and Voskoboinikov, Alexey and Sukhinin, Dmitrii and Makarova, Anna and Skalnaya, Anastasia and Arkhipova, Nastasia and Sinkin, Mikhail and Ossadtchi, Alexei},
  journal = {Journal of Neural Engineering},
  volume  = {19},
  number  = {6},
  pages   = {066016},
  year    = {2022}
}

@InProceedings{Radford2021,
  title = 	 {Learning Transferable Visual Models From Natural Language Supervision},
  author =       {Radford, Alec and Kim, Jong Wook and Hallacy, Chris and Ramesh, Aditya and Goh, Gabriel and Agarwal, Sandhini and Sastry, Girish and Askell, Amanda and Mishkin, Pamela and Clark, Jack and Krueger, Gretchen and Sutskever, Ilya},
  booktitle = 	 {Proceedings of the 38th International Conference on Machine Learning},
  pages = 	 {8748--8763},
  year = 	 {2021},
  editor = 	 {Meila, Marina and Zhang, Tong},
  volume = 	 {139},
  series = 	 {Proceedings of Machine Learning Research},
  month = 	 {18--24 Jul},
  publisher =    {PMLR},
}

@article{Sivakumar2016,
  title   = {Spherical harmonics reveal standing {EEG} waves and long-range neural synchronization during non-{REM} sleep},
  author  = {Sivakumar, Siddharth S and Namath, Amalia G and Gal{\'a}n, Roberto F},
  journal = {Frontiers in Computational Neuroscience},
  volume  = {10},
  pages   = {59},
  year    = {2016}
}

@article{Wallentin2011,
title = {Amygdala and heart rate variability responses from listening to emotionally intense parts of a story},
journal = {NeuroImage},
volume = {58},
number = {3},
pages = {963--973},
year = {2011},
issn = {1053-8119},
doi = {10.1016/j.neuroimage.2011.06.077},
author = {Mikkel Wallentin and Andreas Højlund Nielsen and Peter Vuust and Anders Dohn and Andreas Roepstorff and Torben Ellegaard Lund},
}

@article{Hamalainen1994,
  title   = {Interpreting magnetic fields of the brain: minimum norm estimates},
  author  = {H{\"a}m{\"a}l{\"a}inen, M. S. and Ilmoniemi, R. J.},
  journal = {Medical \& Biological Engineering \& Computing},
  volume  = {32},
  pages   = {35--42},
  year    = {1994}
}

@article{Baillet2017,
  title   = {Magnetoencephalography for brain electrophysiology and imaging},
  author  = {Baillet, Sylvain},
  journal = {Nature Neuroscience},
  volume  = {20},
  number  = {3},
  pages   = {327--339},
  year    = {2017},
  doi     = {10.1038/nn.4504}
}

@article{Wilson2004,
  title   = {Listening to speech activates motor areas involved in speech production},
  author  = {Wilson, Stephen M and Saygin, Ay{\c{s}}e Pinar and Sereno, Martin I and Iacoboni, Marco},
  journal = {Nature Neuroscience},
  volume  = {7},
  number  = {7},
  pages   = {701--702},
  year    = {2004},
  doi     = {10.1038/nn1263}
}

@article{Voskoboynikov_2025,
doi = {10.1088/1741-2552/adfc9c},
year = {2025},
month = {sep},
publisher = {IOP Publishing},
volume = {22},
number = {5},
pages = {056002},
author = {Voskoboynikov, Alexei and Aliverdiev, Magomed and Nekrasova, Yulia and Semenkov, Ilia and Skalnaya, Anastasia and Sinkin, Mikhail and Ossadtchi, Alexei},
title = {Towards stimulation-free automatic electrocorticographic speech mapping in neurosurgery patients},
journal = {Journal of Neural Engineering},
}

@article{Perez2021,
title = {Conscious processing of narrative stimuli synchronizes heart rate between individuals},
journal = {Cell Reports},
volume = {36},
number = {11},
pages = {109692},
year = {2021},
issn = {2211-1247},
doi = {10.1016/j.celrep.2021.109692},
author = {Pauline P{\'e}rez and Jens Madsen and Leah Banellis and Ba{\c{s}}ak T{\"u}rker and Federico Raimondo and Vincent Perlbarg and Melanie Valente and Marie-C{\'e}cile Ni{\'e}rat and Louis Puybasset and Lionel Naccache and Thomas Similowski and Damian Cruse and Lucas C. Parra and Jacobo D. Sitt},
}

@article{Murakami2006,
title = {Contributions of principal neocortical neurons to magnetoencephalography and electroencephalography signals},
journal = {The Journal of Physiology},
volume = {575},
number = {3},
pages = {925--936},
year = {2006},
doi = {10.1113/jphysiol.2006.105379},
author = {Murakami, Shingo and Okada, Yoshio},
}

@article{DingSimon2012Continuous,
author = {Ding, Nai and Simon, Jonathan Z.},
title = {Neural coding of continuous speech in auditory cortex during monaural and dichotic listening},
journal = {Journal of Neurophysiology},
volume = {107},
number = {1},
pages = {78-89},
year = {2012},
doi = {10.1152/jn.00297.2011},
note ={PMID: 21975452},
}

@article{LuoPoeppel2007,
title = {Phase Patterns of Neuronal Responses Reliably Discriminate Speech in Human Auditory Cortex},
journal = {Neuron},
volume = {54},
number = {6},
pages = {1001--1010},
year = {2007},
doi = {10.1016/j.neuron.2007.06.004},
author = {Luo, Huan and Poeppel, David},
}

@article{BrodbeckPresaccoSimon2018,
title = {Neural source dynamics of brain responses to continuous stimuli: Speech processing from acoustics to comprehension},
journal = {NeuroImage},
volume = {172},
pages = {162-174},
year = {2018},
issn = {1053-8119},
doi = {10.1016/j.neuroimage.2018.01.042},
author = {Christian Brodbeck and Alessandro Presacco and Jonathan Z. Simon},
}

@article{BrodbeckHongSimon2018,
title = {Rapid Transformation from Auditory to Linguistic Representations of Continuous Speech},
journal = {Current Biology},
volume = {28},
number = {24},
pages = {3976--3983.e5},
year = {2018},
doi = {10.1016/j.cub.2018.10.042},
author = {Brodbeck, Christian and Hong, L. Elliot and Simon, Jonathan Z.},
}

@article {Gwilliams2022,
	author = {Gwilliams, Laura and King, Jean-Remi and Marantz, Alec and Poeppel, David},
	title = {Neural dynamics of phoneme sequences reveal position-invariant code for content and order},
	volume = {13},
	number = {1},
	year = {2022},
	doi = {10.1038/s41467-022-34326-1},
	journal = {Nature Communications}
}

@article{Koskinen2013SpeechFragments,
author = {Koskinen, Miika and Viinikanoja, Jaakko and Kurimo, Mikko and Klami, Arto and Kaski, Samuel and Hari, Riitta},
title = {Identifying fragments of natural speech from the listener's MEG signals},
journal = {Human Brain Mapping},
volume = {34},
number = {6},
pages = {1477-1489},
doi = {10.1002/hbm.22004},
year = {2013}
}

@article{Broderick2018,
author = {Broderick, Michael P. and Anderson, Andrew J. and Di Liberto, Giovanni M. and Crosse, Michael J. and Lalor, Edmund C.},
title = {Electrophysiological Correlates of Semantic Dissimilarity Reflect the Comprehension of Natural, Narrative Speech},
journal = {Current Biology},
volume = {28},
number = {5},
pages = {803-809.e3},
doi = {10.1016/j.cub.2018.01.080},
year = {2018}
}

@article{Weissbart2020,
    author = {Weissbart, Hugo and Kandylaki, Katerina D. and Reichenbach, Tobias},
    title = {Cortical Tracking of Surprisal during Continuous Speech Comprehension},
    journal = {Journal of Cognitive Neuroscience},
    volume = {32},
    number = {1},
    pages = {155-166},
    year = {2020},
    month = {01},
    issn = {0898-929X},
    doi = {10.1162/jocn_a_01467}
}

\newpage
\appendix
\section*{Appendix}

\section{Additional background on source estimation and task-driven spatial filtering}
\label{app:spatial-filter-background}

For completeness, we provide here additional background motivating
the transition from classical source estimation to the task-driven
interpretable front end used in this work.

In practice, we don't know the pivotal sources and their topographies and it is the goal of a typical neuroscientific study to find them. We are often interested in 
sources whose activity predicts a specific behavioral variable, such as limb
kinematics or a perceived or produced audio signal, or reflects differences
between experimental conditions.
In doing so we can opt for three main strategies: exhaustive search, optimal filtering, or deep learning. The exhaustive search approach relies on the rows of an EEG or MEG inverse operator to define spatial filters $\mathbf{W}$. The primary advantage of this method is its high interpretability, as the resulting filters correspond directly to specific neural source locations. However, this approach requires a highly accurate inverse operator, suffers from low statistical sensitivity due to the multiple comparisons problem, and is generally not designed to handle complex decoding tasks but rather to perform binary contrasts of several experimental conditions.

To circumvent the computational burden of the exhaustive search strategy, optimal spatial filtering provides a task-driven alternative. In this framework, spatial filters are learned directly from the data according to method-specific objectives \cite{Nikulin2011,Dahne2014,ossadtchi2024representational}, making the process highly efficient --- source localization via an inverse operator is only required as an optional final step. Nevertheless, this approach is still not well suited to complex decoding tasks. By constraining the behavioral variable to be a simple function of the filter output or its envelope, it lacks the expressivity required to model intricate, non-linear mappings between neural activity and behavior.

Finally, deep learning represents a fully adaptive, data-driven paradigm where the  spatial filters are learned from the data as a part of a more general decoder tuned to solve a classification or regression task. This approach is suited to complex tasks as it connects source activity estimates to the behavioral variable via a complex learnable transformation. However, these powerful capabilities come with significant trade-offs: deep learning models require massive amounts of training data and act somewhat like a "black box," making it notoriously difficult to link the learned features back to specific anatomical sources and their underlying dynamic properties. 

This shortcoming can be overcome by using an interpretable, physiologically grounded front end that allows us to relate the parameters of the trained network to the neural sources whose activity is pivotal for solving the decoding task. 

\subsection{Linear-estimation relation for spatial filters}

For the linear estimator
\[
\hat{\mathbf s}(t)=\mathbf W^\top\mathbf x(t),
\]
the orthogonality principle gives
\[
\mathbb E\!\left[
\mathbf x(t)
\left(\mathbf s(t)-\mathbf W^\top\mathbf x(t)\right)^\top
\right]=\mathbf 0.
\]
Hence
\[
\mathbf R_{\mathbf x}\mathbf W
=
\mathbf R_{\mathbf x\mathbf s},
\qquad
\mathbf R_{\mathbf x\mathbf s}
=
\mathbb E[\mathbf x(t)\mathbf s(t)^\top].
\]
Using the forward model in Eq.~\eqref{eq:meg-forward}, and assuming that
task-unrelated activity and additive sensor noise are uncorrelated with the
task-related sources,
\[
\mathbf R_{\mathbf x\mathbf s}
=
\mathbf G\mathbf R_{\mathbf s}.
\]
Therefore
\[
\mathbf W
=
\mathbf R_{\mathbf x}^{-1}
\mathbf G
\mathbf R_{\mathbf s},
\]
which is Eq.~\eqref{eq:w_vs_g}.

\section{Mapping the signal subspace to cortical sources via recursive subspace correlation scan}
\label{app:rap-music}

\begin{algorithm}[H]
\caption{Mapping the signal subspace $\mathcal{S}$ to cortical sources via a
recursive subspace-correlation scan.}
\label{alg:rap-music}
\begin{algorithmic}[1]
\State \textbf{Input:} data-subspace basis
       $G_D\in\mathbb{R}^{M\times R}$, forward-model matrix $G_M$,
       and acceptance threshold $\theta$
\State Initialise: $k\gets0$, $A_0\gets[]$, $P_0\gets I$,
       $G_{D,0}\gets G_D$, $G_{M,0}\gets G_M$
\While{$k<R$}
  \For{$i=1,\ldots,Q$}
     \State $\rho_1^k(i)\gets
       \max\subcorr(\mathcal{S}_{D,k},\mathcal{S}_{M_i,k})$
  \EndFor
  \State $i_{k+1}\gets\arg\max_i\rho_1^k(i)$
  \If{$\rho_1^k(i_{k+1})\leq\theta$}
     \State \textbf{break}
  \EndIf
  \State $k\gets k+1$
  \State $A_k\gets[A_{k-1},G_M^{i_k}]$
  \State $P_k\gets I-A_kA_k^{\dagger}$
  \State $G_{D,k}\gets P_kG_{D,k-1}$,
         $G_{M,k}\gets P_kG_{M,k-1}$
\EndWhile
\State \textbf{Output:} selected source sites
       $i_1,\ldots,i_k$ and their subspace correlations
\end{algorithmic}
\end{algorithm}

The forward-model matrix
$\mathbf{G}_M$ is an $M\times (Q\times 3)$ matrix whose $i$-th triplet of
columns is the topography triplet
$[\mathbf{g}_{ix}^M,\mathbf{g}_{iy}^M,\mathbf{g}_{iz}^M]$ of three ECDs
oriented along the $x,y,z$ axes and located at the $i$-th vertex of the
cortical mesh, $i=1,\ldots,Q$, extracted via FreeSurfer~\cite{Fischl2012}.
The data-derived signal-subspace matrix
$\mathbf{G}_D=[\mathbf{v}_1,\ldots,\mathbf{v}_R]$ contains an orthonormal
basis of the analyzed spatial-pattern subspace. For the reported analysis,
we used the leading $R=10$ right singular vectors of the
row-$L_2$-normalized spatial-pattern matrix.
These singular vectors were estimated once from the pooled cross-participant spatial-pattern matrix; the recursive source scan was then performed separately with each participant-specific forward model, and the resulting coordinates were transformed to \texttt{fsaverage} for group visualization.
At recursion $k$, $\mathcal{S}_{D,k}$ is the column space of the projected
data matrix $\mathbf{G}_{D,k}$. For each cortical site $i$, the projected
three-orientation lead-field triplet is reduced by singular value
decomposition to two normalized tangential directions, whose span defines
$\mathcal{S}_{M_i,k}$. Once a site is selected, its full
three-orientation lead-field triplet is used to construct the recursive
projector.
The subspace-correlation function $\subcorr$ is
the vector of cosines of the principal angles between two
subspaces~\cite{GolubVanLoan2013,Mosher2002}; 
we use its largest element $\rho_1$ as the candidate-site score.
The recursive projection
step is inspired by RAP-MUSIC~\cite{Mosher2002} and improves the dynamic
range of $\rho_1$ when several pivotal sources are present. We use $Q=5124$ mesh vertices and $\theta=0.98$ for the analysis reported in
Figure~\ref{fig:rap-music-dipoles}.

\section{Sensitivity of template-anatomy localization to individual MRI}
\label{app:individual-mri-validation}

The public MEG-MASC release contains individual T1-weighted MRI for
12 participants. Five scans had incomplete brain coverage and were unsuitable
for cortical reconstruction, leaving seven participants (11, 14, 15, 18, 24,
26, and 27) with usable individual FreeSurfer anatomy. We quantified the
effect of using the common \texttt{fsaverage} anatomy in this seven-participant
subset. For each participant, the same 25 branch sensor topographies from the
seed-42 model, computed using the learned temporal filters, were
localized on the participant's individual cortical anatomy using the same
inverse-model settings as in the main analysis. The resulting source-magnitude
maps were morphed to the \texttt{fsaverage} \texttt{ico4} mesh and re-normalized
to unit \(L_2\) norm, and were compared with the current template-anatomy
localization of the same participant--branch pair.

\begin{figure*}[t]
    \centering
    \includegraphics[width=\textwidth]{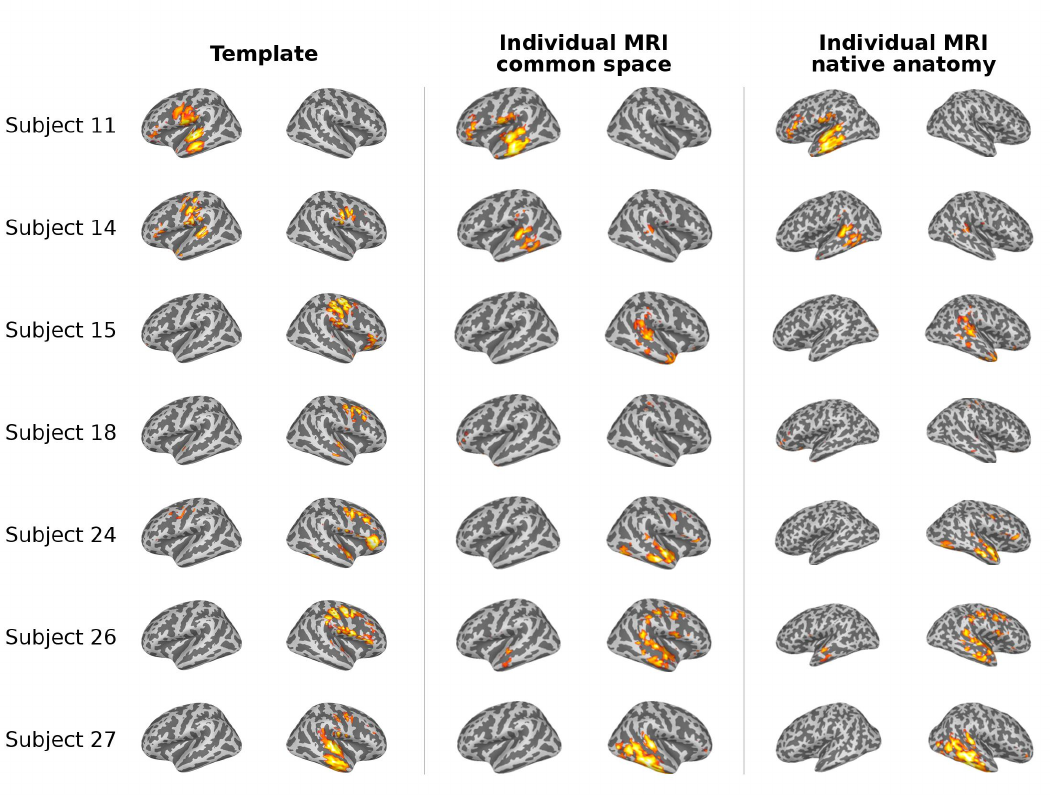}
    \caption{Representative comparison of template- and
    individual-anatomy source localization for the seven participants with
    usable structural MRI. For each participant, the displayed branch was
    selected deterministically as the non-rough branch whose
    template-versus-individual source-map Pearson correlation was closest to
    that participant's median correlation. The left column shows the current
    main-analysis estimate on the common \texttt{fsaverage} anatomy; the
    middle column shows the corresponding estimate obtained on the
    participant's individual anatomy and subsequently morphed to
    \texttt{fsaverage}; the right column shows the same individual-anatomy
    estimate on the participant's native cortical surface. Color scaling is
    panel-specific and is used only to visualise spatial distributions;
    quantitative agreement was evaluated from the normalized source maps over
    all 25 branches per participant.}
    \label{fig:individual-mri-examples}
\end{figure*}

\begin{figure*}[t]
    \centering
    \includegraphics[width=\textwidth]{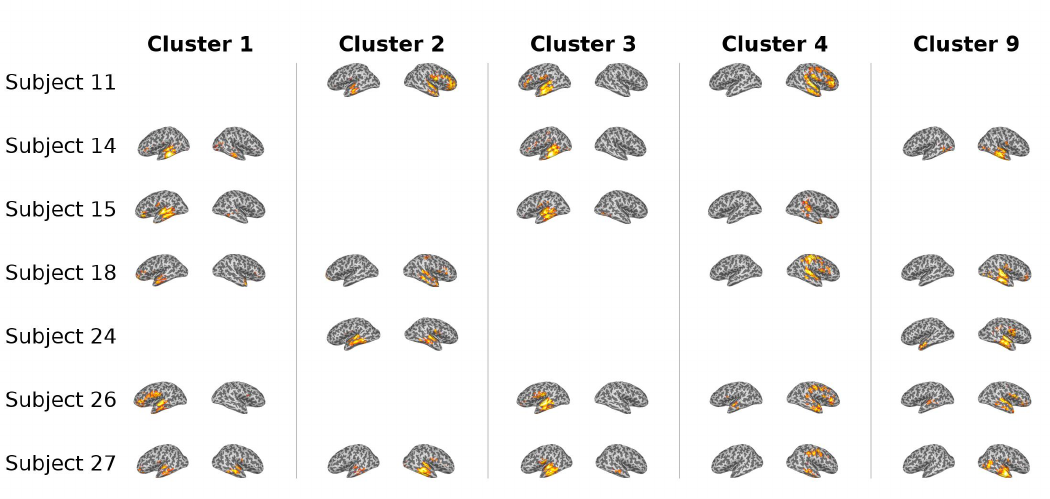}
    \caption{Native-anatomy localisation of representative branches
    from five major current branch-pattern clusters (C1, C2, C3, C4, and C9)
    in the seven-participant individual-MRI subset. Rows correspond to
    participants and columns to clusters; blank cells indicate that the
    participant had no branch assigned to that cluster. For each represented
    participant--cluster pair, the displayed branch is the participant's
    cluster member closest to the current canonical cluster medoid under the
    same combined source--temporal similarity used for the main clustering
    analysis. These panels show the individual-anatomy localization of branches
    assigned by the main analysis and are not an independent re-clustering of
    the individual-MRI maps. Clusters were chosen from the twelve largest 
    clusters to span most of the seven participants with individual MRI scans.}
    \label{fig:individual-mri-cluster-examples}
\end{figure*}

For each branch we measured Pearson correlation and cosine similarity
between the two source-magnitude maps, agreement of the hemisphere containing
the map maximum, and the Euclidean distance in MNI coordinates between their
peaks. Because branches from the same participant are not independent
observations, metrics were first averaged within participant and 95\%
confidence intervals were obtained by bootstrap resampling the seven
participants (10,000 resamples). 
After first averaging each metric within participant, the mean across the seven
participants was 0.633 for Pearson correlation (95\% CI 0.581--0.679),
0.875 for cosine similarity (0.858--0.890), and 83.4\% for
peak-hemisphere agreement (80.6--86.3\%); mean peak displacement was
48.1\,mm (42.4--52.6\,mm).
The 48.1\,mm peak displacement quantifies disagreement between the peaks
of two estimated inverse solutions, not localization error relative to known
neuronal generators. Neither estimate is ground truth; this comparison is
therefore a sensitivity analysis rather than a source-recovery validation.
The quantitative comparison, together with the representative maps in
Figures~\ref{fig:individual-mri-examples} and
\ref{fig:individual-mri-cluster-examples}, indicates partial agreement in
broad spatial distribution and hemispheric localization, but substantial
uncertainty in the precise location of source-map maxima.

The representative maps provide a complementary qualitative view of
this sensitivity. Across template- and individual-anatomy estimates, the broad
lateral temporal and peri-Sylvian distributions and their dominant hemisphere
are generally preserved, whereas the precise location and spatial extent of
the maxima vary. The individual-anatomy cluster examples show the same
large-scale organization as the corresponding medoids in
Figure~\ref{fig:patterns}: Clusters~1 and 3 are predominantly left-lateralized,
whereas Clusters~2, 4, and 9 are predominantly right-lateralized. Across
participants, these motifs repeatedly involve lateral temporal cortex, with
Cluster~4 showing a comparatively more dorsal peri-Sylvian extension. This
recurrence on individual cortical anatomy supports interpreting the clusters
as broad lateralized spatial motifs rather than artifacts of the common
template anatomy. Their distribution is also consistent with the lateral
temporal regions implicated in continuous speech tracking
\cite{Ding2012,Gross2013,Nourski2009,Kubanek2013,Brodbeck2020} and with the
broader peri-Sylvian organization described in dual-stream accounts of speech
processing~\cite{Hickok2007}. In light of the peak-displacement analysis above,
however, these comparisons support regional and hemispheric organization
rather than precise assignment of individual clusters to cortical parcels.

\section{Stability of branch interpretations}
\label{app:interpretation-stability}

\subsection{Stability across stories, sessions, and model initializations}

We assessed the reproducibility of the spatial and temporal branch
interpretations across stimulus stories, repeated recording sessions, and
independent model initializations. For story-level stability of the seed-42
model, we compared the same participant--branch item between every pair of
the four stories within each session. Across the 12 resulting story-pair and
session comparisons, the median same-item Pearson correlation ranged from
0.872 to 0.921 for sensor-space spatial patterns and from 0.831 to 0.894 for
cortical source-magnitude maps. Temporal representations were still more
stable: median correlations ranged from 0.998 to 0.999 for temporal patterns
and from 0.986 to 0.991 for their log-magnitude spectra.

Repeated-session stability was evaluated in the 22 participants with
two recording sessions by comparing the same branch and story across
sessions. Across the four stories, median correlations ranged from 0.758 to
0.826 for sensor-space spatial patterns, from 0.563 to 0.610 for cortical
source-magnitude maps, from 0.997 to 0.998 for temporal patterns, and from
0.978 to 0.985 for log-magnitude spectra. Thus, the temporal structure is
particularly stable across recordings, while the spatial representation
shows greater session-to-session variability.

Across independently trained models, individual branch indices need
not retain a one-to-one identity because equivalent representations can be
permuted or mixed across branches. We therefore first compared the spatial
representations at the subspace level. For each participant and each of the
15 pairs of model initializations, the 25 sensor-space spatial patterns from
each model were arranged as a \(25\times208\) matrix. We compared the row
spaces of the two matrices using their principal angles and summarized the
result as the root-mean-square cosine of these angles. Averaged over all
\(27\times15=405\) participant--seed-pair comparisons, this spatial subspace
overlap was 0.887.

We additionally performed a descriptive branch-correspondence
analysis separately for each representation. For each pair of model
initializations, we formed a \(25\times25\) matrix containing the similarity
between every pair of branches, averaged equally across participants, and
used the Hungarian algorithm to obtain the one-to-one assignment maximizing
the total similarity. The assignment was performed independently for each
representation rather than imposing a common branch correspondence across
views. We used absolute Pearson correlation for the sign-ambiguous
sensor-space spatial and temporal patterns, and signed Pearson correlation
for source-magnitude maps and temporal log-magnitude spectra. For each seed
pair, we then averaged the correlation over its 25 matched branch pairs.
Across the 15 seed pairs, these mean matched correlations ranged from
0.465 to 0.493 for sensor-space spatial patterns, from 0.511 to 0.540 for
source-magnitude maps, from 0.721 to 0.777 for temporal patterns, and from
0.742 to 0.787 for temporal log-magnitude spectra. Thus, cross-initialization
reproducibility is stronger at the representation level than at the level of
uniquely identifiable individual branches.

\begin{figure}[p]
    \centering

    \begin{subfigure}{\linewidth}
        \centering
        \includegraphics[width=\linewidth]{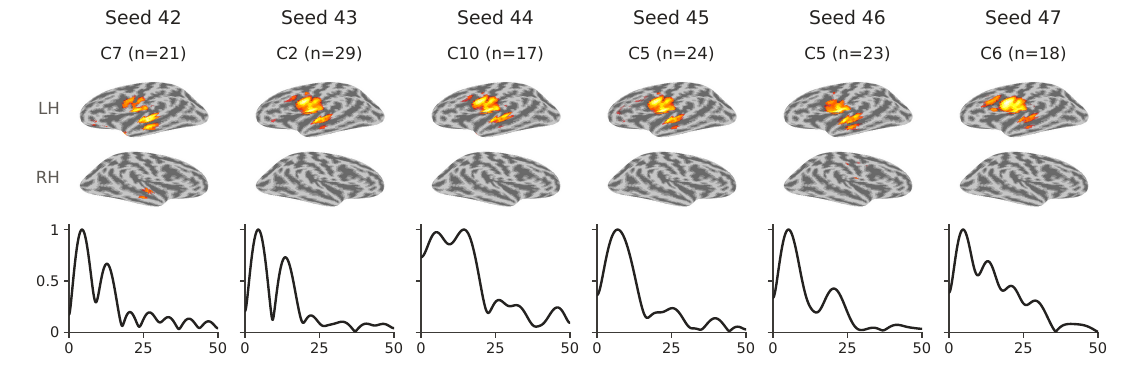}
        \caption{Left lateral/peri-Sylvian pattern.}
        \label{fig:cross-seed-motif-left}
    \end{subfigure}

    \smallskip

    \begin{subfigure}{\linewidth}
        \centering
        \includegraphics[width=\linewidth]{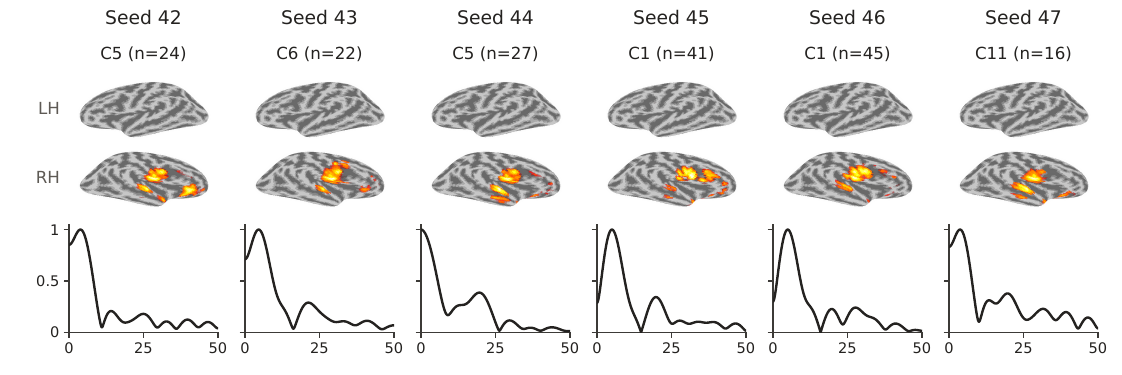}
        \caption{Right posterior-lateral temporal pattern.}
        \label{fig:cross-seed-motif-posterior}
    \end{subfigure}

    \smallskip

    \begin{subfigure}{\linewidth}
        \centering
        \includegraphics[width=\linewidth]{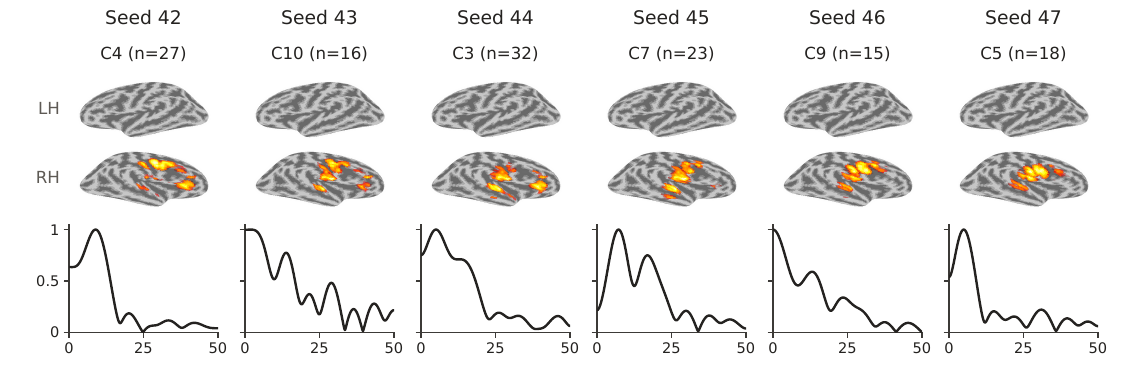}
        \caption{Right superior/peri-Sylvian pattern.}
        \label{fig:cross-seed-motif-superior}
    \end{subfigure}

    \caption{Representative spatial--temporal motifs across six
    independently trained models (seeds 42--47). Each subfigure shows
    one seed-specific cluster medoid per model, selected by visual
    comparison among its twelve largest principal spatial--spectral
    clusters. LH and RH show approximate cortical inverse-model
    source-magnitude projections on the common \texttt{fsaverage}
    surface in lateral views. The corresponding learned temporal-pattern
    magnitude spectrum (frequency in Hz) is shown below each pair,
    normalized to its own maximum. Cluster identifiers (C1, C2, etc.)
    are specific to each seed, and $n$ denotes the number of
    participant--branch items in that cluster after spatial-roughness
    screening. The examples illustrate recurrence of broad regional patterns, 
    not one-to-one reproducibility of individual inverse-model source projections.}
    \label{fig:cross-seed-motifs}
\end{figure}

To examine whether the broad anatomical motifs identified in
the seed-42 analysis recur across model initializations, we compared
the spatial--spectral cluster medoids of six independently trained
models (seeds 42--47). The participant--branch patterns were clustered
separately for each seed using the first-session \emph{Cable Spool Fort}
recordings and the procedure described in
Section~\ref{ssec:clustering}. Among the twelve largest principal
clusters of each model, we identified representative examples of
three qualitatively recurring spatial motifs: a left lateral/peri-Sylvian
pattern, a right posterior-lateral temporal pattern, and a right
superior/peri-Sylvian pattern
(Figure~\ref{fig:cross-seed-motifs}). These broad hemispheric and
regional distributions are recognizable across all six initializations,
although the extent and relative prominence of individual cortical
foci, as well as the temporal spectral profiles, vary. This descriptive
comparison illustrates recurrence of broad anatomical motifs rather
than quantitative reproducibility of individual clusters or precise
correspondence of cortical sources across models.

\begin{figure}[htbp]
\centering
\includegraphics[width=\linewidth]{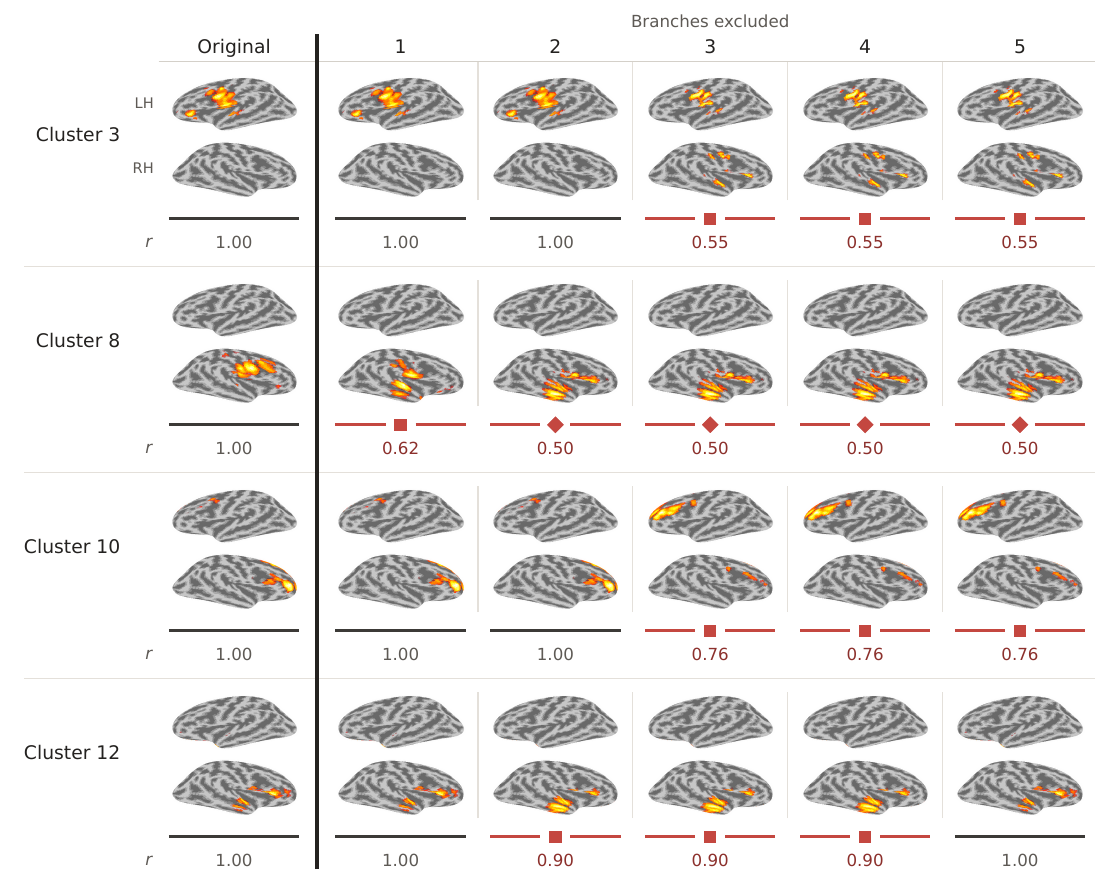}
\caption{Medoid evolution under branch exclusion in seed 42.
The four rows show all of the twelve largest principal clusters whose
medoids change at least once (C3, C8, C10, and C12); the remaining eight
retain their original medoids throughout. Columns display the original
medoid and the medoid after excluding one through five branches.
Left- and right-hemisphere panels show approximate cortical inverse-model
source-magnitude projections; the values below each pair are signed
Pearson correlations with the original medoid's source map. Red markers
indicate a replacement medoid, and C12 returns to its original medoid
after five exclusions.}
\label{fig:medoid-exclusion}
\end{figure}

\subsection{Sensitivity of cluster medoids to exclusion of low-importance branches}
\label{app:medoid-exclusion}

We performed a descriptive sensitivity analysis of medoid
selection under exclusion of branches with low retrieval importance.
Within each of six independently
trained models (seeds 42--47), we ranked the 25 branches by the held-out
Top-1 accuracy decrease when each branch output was zeroed
(Section~\ref{ssec:training}; Figure~\ref{fig:branch-importance}).
We progressively excluded participant--branch items corresponding to the
one through five lowest-ranked branches per seed. This range was chosen
descriptively, not as an optimized threshold. We retained the original
cluster assignments and recomputed medoids from the surviving members
using the same joint cortical source-magnitude-map and
log-temporal-spectrum similarity as in Section~\ref{ssec:clustering}.
Neither the model nor the clusters were retrained or reclustered.

\begin{table}[htbp]
\centering
\footnotesize
\setlength{\tabcolsep}{3pt}
\caption{Sensitivity of cluster medoids to exclusion of
low-importance branches across six model initializations. The top-12
subset contains the twelve largest original principal clusters per
seed ($n=72$); all principal clusters total $n=333$. Changed and
vanished clusters are counted separately relative to these original
totals. Median signed Pearson correlations compare original and
replacement medoids only, using cortical source-magnitude maps and
log temporal spectra.}
\label{tab:medoid-exclusion}
\begin{tabular}{lccccc}
\toprule
Subset & Excluded & Changed & Source $r$ & Spectrum $r$ & Vanished \\
\midrule
\multirow{5}{*}{Top 12 per seed}
& 1 & 4/72 (5.6\%) & 0.683 & 0.778 & 0 \\
& 2 & 8/72 (11.1\%) & 0.811 & 0.796 & 0 \\
& 3 & 10/72 (13.9\%) & 0.781 & 0.748 & 0 \\
& 4 & 12/72 (16.7\%) & 0.753 & 0.775 & 0 \\
& 5 & 13/72 (18.1\%) & 0.760 & 0.795 & 0 \\
\midrule
\multirow{5}{*}{All principal}
& 1 & 19/333 (5.7\%) & 0.625 & 0.730 & 2 \\
& 2 & 38/333 (11.4\%) & 0.628 & 0.714 & 4 \\
& 3 & 65/333 (19.5\%) & 0.584 & 0.705 & 12 \\
& 4 & 80/333 (24.0\%) & 0.579 & 0.705 & 16 \\
& 5 & 94/333 (28.2\%) & 0.579 & 0.709 & 18 \\
\bottomrule
\end{tabular}
\end{table}

After five exclusions, 59 of the 72 original top-12 medoids
(81.9\%) were retained, 13 differed from their original representatives,
and none of these clusters vanished. Across all exclusion levels, 57
of the 72 representatives never changed. Among changed medoids at the
five-branch exclusion, median source-map correlation was $0.760$ and
median log-spectrum correlation was $0.795$. Across all 333 principal
clusters, 221 (66.4\%) retained their medoids, 94 changed, and 18
vanished; median replacement correlations were $0.579$ for source maps
and $0.709$ for log spectra
(Table~\ref{tab:medoid-exclusion}).
These preservation rates are descriptive measures under the
original cluster assignments, rather than independent validation of
source-space pattern stability.

The seed-42 examples in Figure~\ref{fig:medoid-exclusion} illustrate
how the remaining changes affect the spatial motifs. Cluster~8
retains a predominantly right-lateralized temporal distribution,
although its strongest foci shift ($r\approx0.50$), whereas
Cluster~12 shows a closely similar pattern during replacement
($r\approx0.90$) and ultimately returns to its original medoid.
Cluster~3 retains a left-hemisphere focus but develops additional
right-hemisphere foci ($r\approx0.55$), while Cluster~10 shows
a change in the relative prominence of its hemispheric projections
($r\approx0.76$). Thus, 
when the original cluster assignments are held fixed, 
the major cluster representatives are
largely preserved under exclusion of low-importance branches,
with some replacements retaining their broad spatial organization
despite changes in the location, extent, or lateralization of
individual foci.

\section{Long-support temporal patterns and spectral resolution}
\label{app:long-temporal-patterns}

\begin{figure}[!ht]
    \centering
    \includegraphics[width=0.95\linewidth]
    {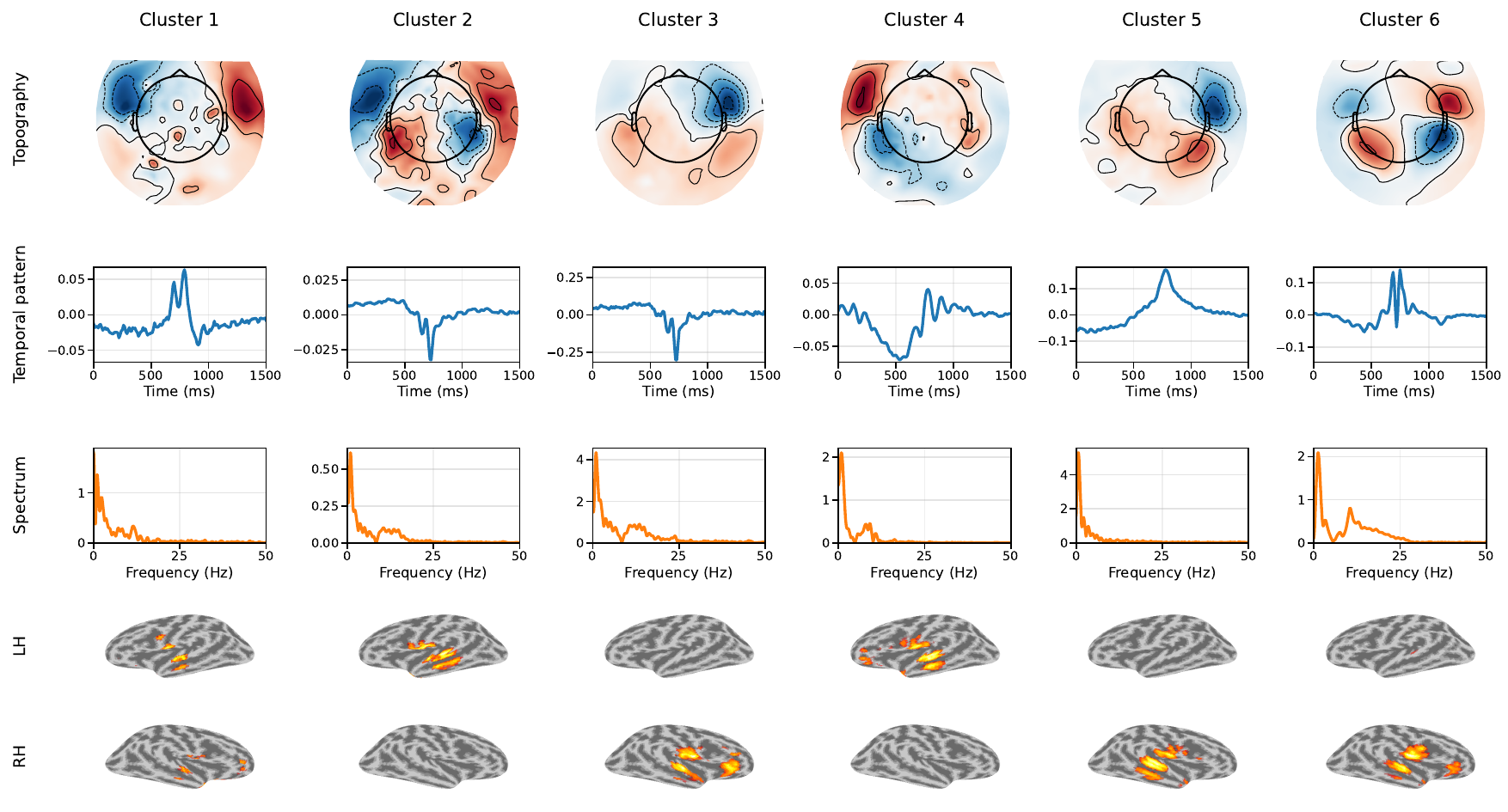}
    \vspace{0.5em}
    \includegraphics[width=0.95\linewidth]
    {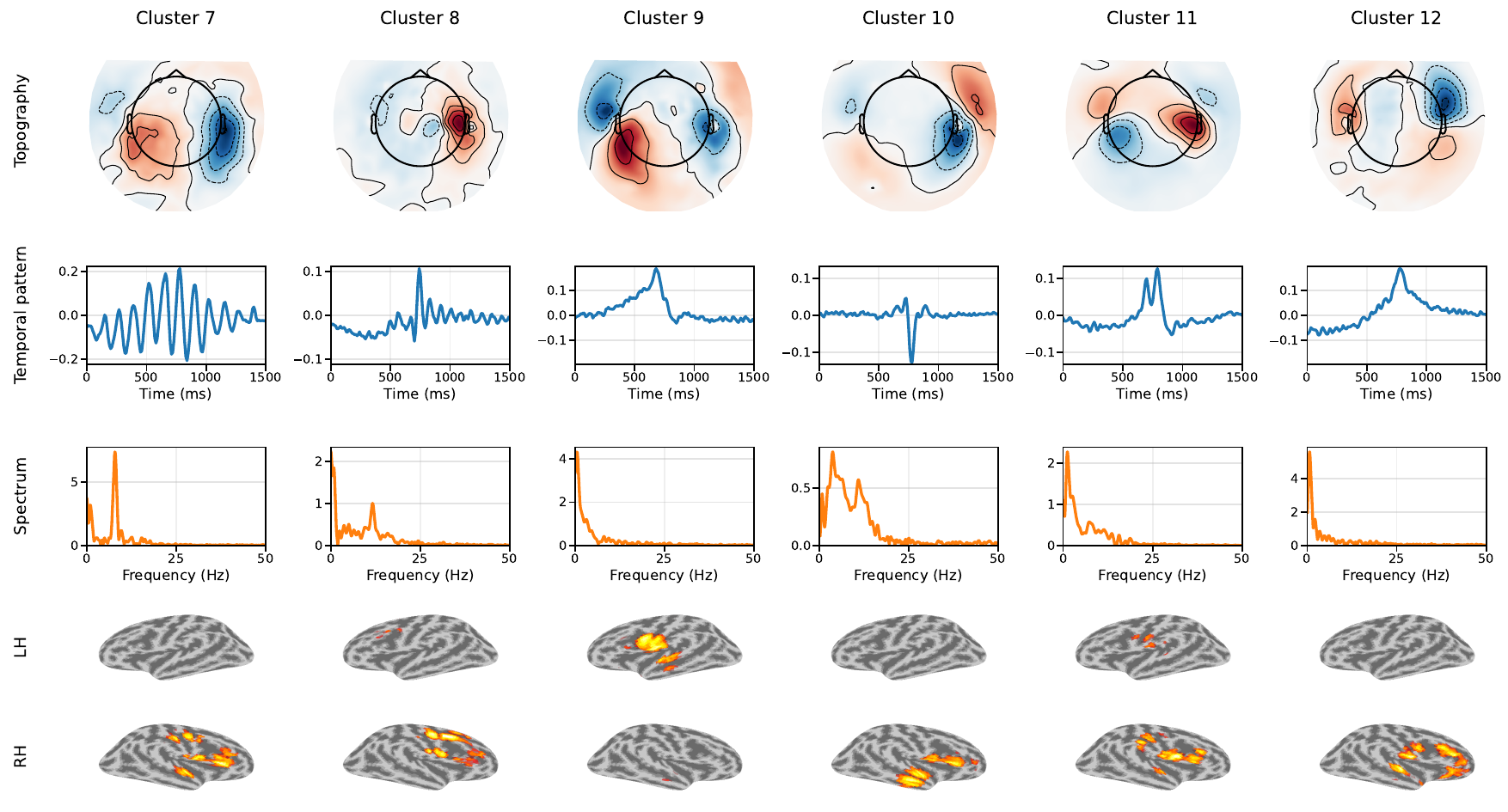}
    \caption{The 12 largest spatial--spectral clusters for the \(K=25\) model with 151-sample (1.51\,s) temporal filters, computed from the first-session \emph{Cable Spool Fort} recordings after spatial-roughness screening. Each column shows one cluster medoid. Rows show, from top to bottom, the sensor-space spatial pattern, the 1.51\,s temporal pattern, its magnitude spectrum, and the corresponding participant-specific MNE-Python~\cite{Gramfort2013} source-magnitude estimate on the common \texttt{fsaverage} surface in left- and right-hemisphere lateral views. The 151-sample support provides an independent Fourier-bin spacing of approximately 0.66\,Hz, compared with 6.67\,Hz for the main 150\,ms analysis.}
    \label{fig:patterns-t151}
\end{figure}

To examine the learned temporal structure at substantially finer
frequency resolution, we repeated the complete branch-interpretation analysis
for the seed-42 model with a 151-sample temporal filter, corresponding to
1.51\,s at the 100\,Hz MEG sampling rate. The architecture, data split,
training procedure, physical-unit reconstruction, participant-specific source
localization, spatial-roughness screening, and spatial--spectral clustering
procedure were otherwise the same as for the main 150\,ms model. This
long-support model achieved 41.14\% Top-1 and 71.63\% Top-10 retrieval
accuracy.

A 151-sample temporal pattern has an independent Fourier-bin spacing
of
\[
\Delta f = \frac{f_s}{T} = \frac{100}{151} \simeq 0.66~\mathrm{Hz},
\]
compared with approximately 6.67\,Hz for the 15-sample patterns in the main
analysis. The spectra in Figure~\ref{fig:patterns-t151} are evaluated on a zero-padded frequency grid
for smooth visualization, but the approximately tenfold improvement in
spectral resolution follows from the longer temporal support itself.

The same spatial-roughness criterion excluded 91 of the 675
participant--branch items, leaving 584 items for clustering. The twelve
largest long-support clusters each contained branches from 8--16
participants. Their source estimates again form predominantly lateral
temporal motifs with additional dorsal and anterior extensions, and many are
strongly lateralized. Thus, increasing temporal support does not replace the
broad spatial organization recovered by the main model.

To compare the two decompositions without assuming that branch
indices are preserved across independently trained models, we matched the
eleven multi-participant medoids displayed in the main 150\,ms analysis
one-to-one to the twelve long-support medoids by maximizing cortical
source-map correlation. The matched source maps had Pearson correlations of
0.58--0.94 (median 0.77), while the corresponding sensor topographies had
absolute correlations of 0.59--0.93 (median 0.70). 
Correlations between the displayed magnitude spectra were more variable, ranging from 0.20 to 0.90 (median 0.55).
The spatial and source-level motifs therefore recur more strongly than the
exact temporal profiles, which are reorganized when the available temporal
support increases by an order of magnitude.

At the finer spectral resolution, the temporal patterns resolve
secondary components that were only coarsely represented by the 150\,ms
analysis. Several medoids show resolved secondary maxima within 
approximately 8--13\,Hz, with additional peaks extending into the lower-beta range near 15--17\,Hz.
These faster components occur across both left- and right-lateralized source
motifs. The long-support analysis therefore identifies
alpha/lower-beta-range spectral components in learned temporal
patterns associated with spatial motifs in both hemispheres.

\section{Occlusion design and complete feature-wise statistics}
\label{app:occlusion-design-details}

\begin{table}[!ht]
\centering
\scriptsize
\setlength{\tabcolsep}{3pt}
\caption{Feature-wise audit of the paired MEG substitution analysis.
Retrieval ranks are participant-balanced means for the primary seed-42 model;
\(p_{\mathrm{FWER}}\) denotes the two-sided single-step
max-\(|T|\) familywise-error-corrected \(p\)-value.}
\label{tab:occlusion-design-details}

\resizebox{\textwidth}{!}{%
\begin{tabular}{@{}lrrrrrrrrr@{}}
\toprule
Feature &
\shortstack{Eligible\\queries} &
\shortstack{Distinct feature\\events} &
\shortstack{MEG\\replaced (\%)} &
\shortstack{Minimum present\\donor-pool size} &
\shortstack{Minimum absent\\donor-pool size} &
\shortstack{Rank with\\present donor} &
\shortstack{Rank with\\absent donor} &
\shortstack{Rank\\difference} &
\(p_{\mathrm{FWER}}\) \\
\midrule
Silence
    & 857 & 401 & 19.0 & 5 & 288 & 38.1 & 113.7 & 75.62 & \(<10^{-4}\) \\
High loudness
    & 993 & 6105 & 43.8 & 5 & 85 & 85.6 & 146.4 & 60.77 & \(<10^{-4}\) \\
Vowels
    & 659 & 3921 & 52.5 & 110 & 6 & 104.2 & 142.9 & 38.70 & \(<10^{-4}\) \\
Strong acoustic onset
    & 997 & 5026 & 44.5 & 9 & 337 & 99.5 & 135.7 & 36.19 & \(<10^{-4}\) \\
Fricatives
    & 938 & 1730 & 20.4 & 9 & 1043 & 55.0 & 67.8 & 12.80 & \(<10^{-4}\) \\
Stops
    & 1000 & 2069 & 21.6 & 5 & 651 & 56.9 & 66.2 & 9.27 & \(<10^{-4}\) \\
Sibilants
    & 957 & 929 & 12.9 & 5 & 1463 & 41.4 & 49.3 & 7.94 & \(<10^{-4}\) \\
High surprisal
    & 654 & 551 & 28.9 & 42 & 5 & 72.3 & 79.6 & 7.34 & \(<10^{-4}\) \\
Pseudowords
    & 114 & 46 & 12.0 & 7 & 722 & 41.1 & 42.8 & 1.71 & 0.039 \\
Silence offsets
    & 805 & 401 & 2.4 & 269 & 12661 & 23.2 & 24.8 & 1.60 & \(<10^{-4}\) \\
Liquids/glides
    & 942 & 1171 & 13.0 & 7 & 1419 & 39.0 & 40.2 & 1.21 & \(<10^{-4}\) \\
Nasals
    & 970 & 1149 & 12.4 & 6 & 1526 & 38.0 & 39.2 & 1.18 & \(<10^{-4}\) \\
Word onsets
    & 1005 & 2620 & 12.8 & 17 & 4899 & 37.1 & 38.2 & 1.02 & \(<10^{-4}\) \\
Schwa
    & 713 & 467 & 6.7 & 8 & 2133 & 33.8 & 34.7 & 0.96 & \(<10^{-4}\) \\
Silence edges
    & 845 & 795 & 4.4 & 7 & 6717 & 28.6 & 29.5 & 0.91 & \(<10^{-4}\) \\
High predictive entropy
    & 684 & 594 & 30.5 & 23 & 5 & 77.8 & 79.2 & 1.40 & 0.119 \\
Silence onsets
    & 822 & 401 & 2.4 & 274 & 12653 & 25.9 & 26.0 & 0.05 & 0.494 \\
Rare words
    & 172 & 519 & 15.6 & 256 & 5 & 41.5 & 38.5 & -2.98 & 0.011 \\
Random word lists
    & 115 & 283 & 51.6 & 10 & 46 & 153.5 & 135.7 & -17.77 & \(<10^{-4}\) \\
\bottomrule
\end{tabular}
}
\end{table}

Table~\ref{tab:occlusion-design-details} summarizes three aspects of
the paired MEG substitution analysis for each feature: the amount of test data
contributing to the analysis, the availability of duration-matched donor
intervals, and retrieval performance under the two substitution conditions.
``Eligible queries'' is the number of 3\,s test windows used for a feature
after excluding windows in which the feature was absent, covered the complete
query, or could not be supplied with the required donor pairs. ``Distinct
feature events'' counts each feature-defining event once in the held-out test
material before the overlapping 3\,s windows are formed. Depending on the
feature, these are silence events, phoneme intervals, word occurrences, or
connected components of a thresholded acoustic mask. For random word lists,
the 283 events are inserted-word occurrences belonging to 19 contiguous
word-list blocks. ``MEG replaced'' is the mean percentage of each eligible
3\,s query covered by the substituted intervals. Each substituted component
is replaced by a donor interval of exactly the same duration, so donor
availability depends on component duration. The donor-pool columns therefore
report, separately for feature-present and feature-absent states, the smallest
number of non-overlapping candidate intervals available across the component
durations that actually occur for that feature. The two rank columns report
the participant-balanced mean rank of the correct audio candidate after
substitution with feature-present or feature-absent donor MEG for the primary
seed-42 model, with rank 1 being best. ``Rank difference'' is the
feature-absent-donor rank minus the feature-present-donor rank, so a positive
value indicates better preservation of retrieval with feature-present donor
MEG.

\section{Robustness of paired occlusion effects across model initializations}
\label{app:occlusion-seed-robustness}

\begin{figure}[!ht]
    \centering
    \begin{subfigure}[t]{0.75\linewidth}
        \centering
        \includegraphics[width=\linewidth]
        {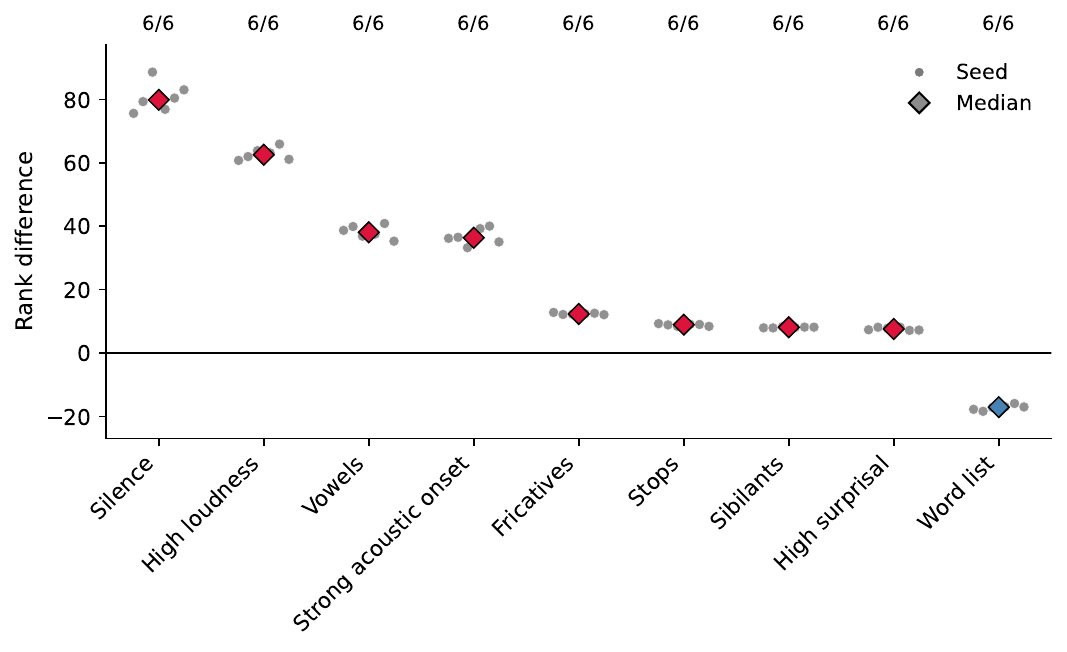}
        \caption{Features with a large absolute across-seed median rank difference.}
    \end{subfigure}

    \begin{subfigure}[t]{0.75\linewidth}
        \centering
        \includegraphics[width=\linewidth]
        {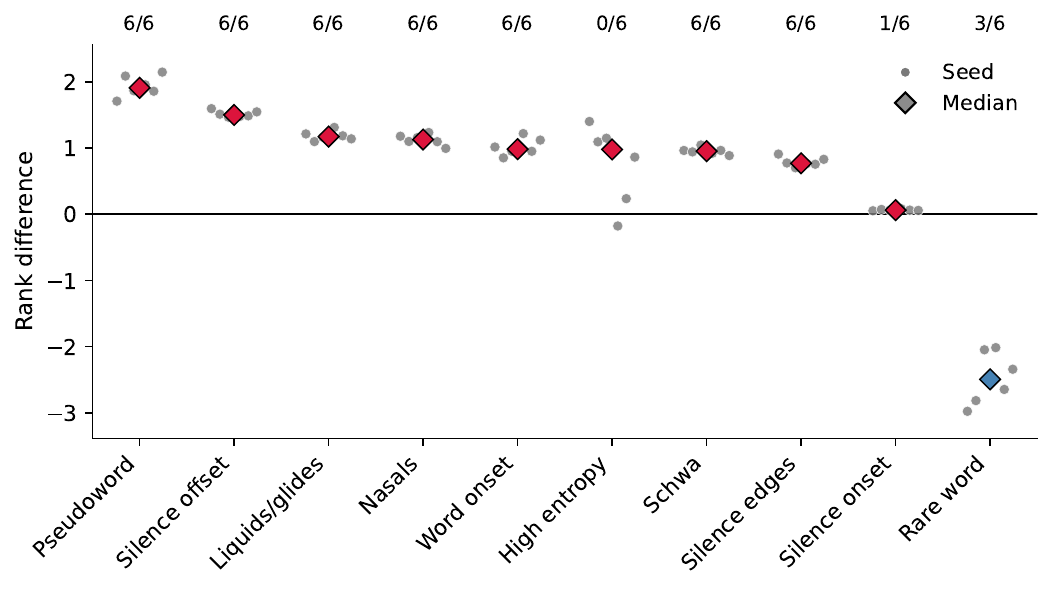}
        \caption{The remaining features, shown on an expanded vertical scale.}
    \end{subfigure}
    \caption{
    Robustness of paired MEG occlusion effects across six model
    initializations. Small grey points show the participant-balanced group mean
    rank difference for each trained model, and diamonds show the across-seed
    median. Numbers above each feature indicate how many models reached
    two-sided single-step max-$|T|$ familywise-error-corrected significance at
    $p<0.05$. Effect direction is given by the sign of the rank difference, so
    these counts report corrected significance irrespective of whether the effect
    is positive or negative. The division between panels is solely for visualization and has no
    inferential meaning.
    }
    \label{fig:occlusion-seed-robustness}
\end{figure}

The main paired occlusion analysis used the model trained with seed 42. To test
whether its conclusions depended on that particular trained solution, we
repeated the complete analysis for five additional instances of the same
two-block, $K=25$ architecture trained with seeds 43--47. The feature
definitions, eligible test queries, donor pool, five-pair substitution
procedure, participant-level aggregation, and two-sided max-$|T|$ inference were
identical to those described in Section~\ref{ssec:occlusion-features}. Donor
selection and sign-flip permutations were held fixed across checkpoints using
analysis seed 42, so that the comparison varied the trained model rather than
the occlusion randomization. 
These model-wise repetitions assess robustness to initialization;
statistical inference for the primary analysis is based on the 27 participant
effects from the seed-42 model.

All 15 features that showed a corrected positive effect in the seed-42 model
remained positive and reached corrected significance in all six trained models
(Figure~\ref{fig:occlusion-seed-robustness}). The largest effects were also
stable in magnitude: the across-seed medians were 79.89 rank positions for
silence (range 75.62--88.64), 62.57 for high loudness (60.77--65.92), 38.10 for
vowels (35.28--40.86), and 36.35 for strong acoustic onset (33.23--40.04).

The random-word-list contrast was negative and reached corrected
significance in all six models, with an across-seed median of $-17.10$ and a
range of $-18.40$ to $-15.94$. Rare words were also negative in all six models
and reached corrected significance in three of six, whereas the high-entropy
effect was small and variable and did not pass correction in any model. The
silence-onset contrast remained close to zero and reached corrected
significance in one of six models.
Thus, the main conclusion of the occlusion analysis---that retrieval
uses MEG information associated with acoustic state, acoustic change, phonetic
classes, and selected contextual properties---does not depend on the seed-42
trained solution. The cross-feature ordering of effect magnitudes remains
subject to the mask-duration and eligible-window differences discussed in
Section~\ref{sec:limitations}.

\section{Overlap between stimulus-feature masks}
\label{app:feature-mask-overlap}

\begin{figure}[!ht]
    \centering
    \includegraphics[width=0.75\linewidth]
    {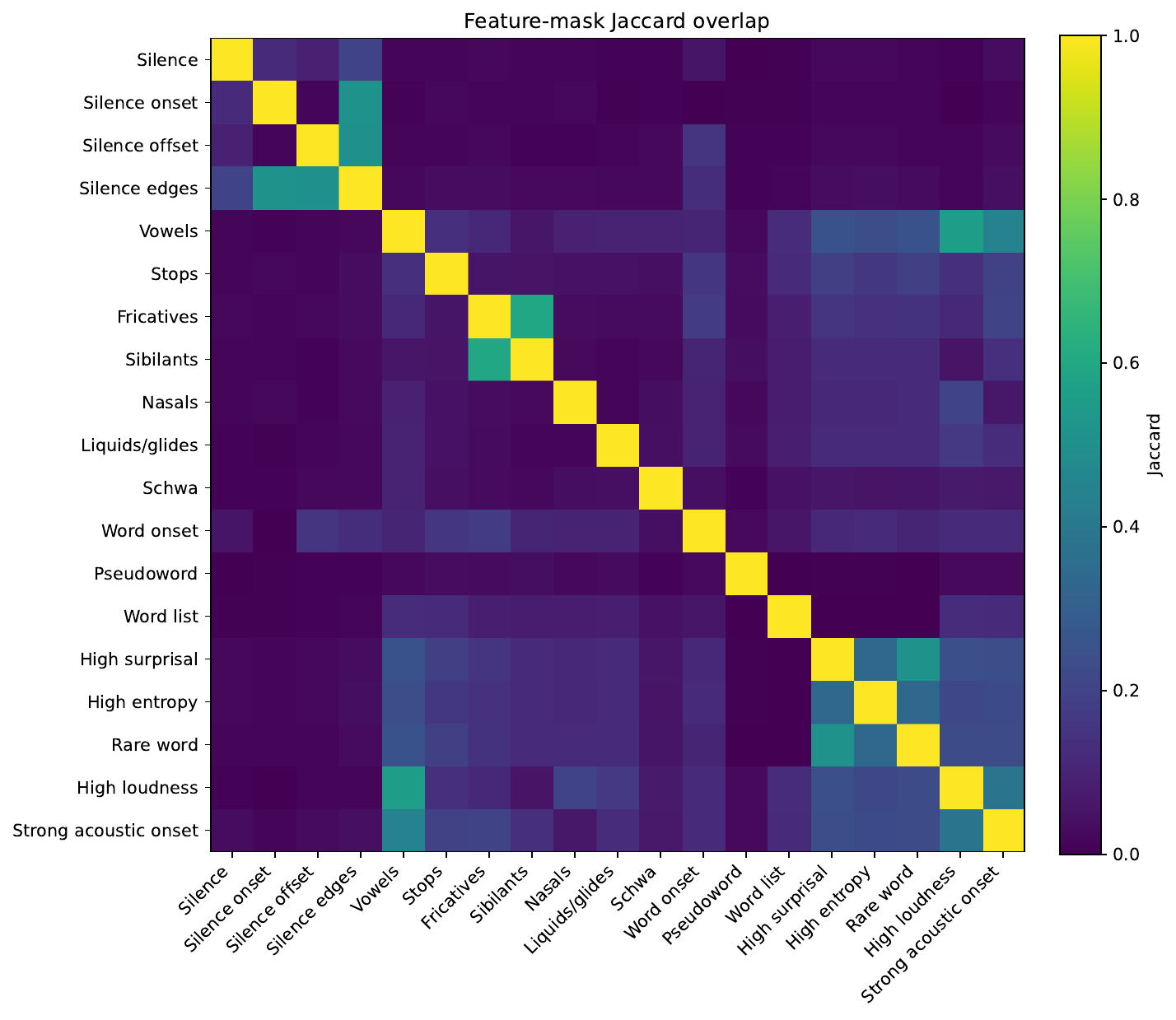}
    \caption{Pairwise overlap of the 19 feature-present masks on the
    test material, quantified by the Jaccard index. The diagonal denotes
    self-overlap. The matrix shows substantial heterogeneity in dependence
    between stimulus features, including overlap induced by nested phonetic
    categories and by correlated acoustic and linguistic variables.}
    \label{fig:feature-mask-overlap}
\end{figure}

To quantify dependence among the occlusion targets, we computed the
pairwise Jaccard overlap of all 19 feature-present masks on the test material.
Figure~\ref{fig:feature-mask-overlap} reports the complete matrix. Overlap was
heterogeneous rather than uniformly high. High loudness and strong acoustic
onset overlapped moderately (Jaccard \(=0.387\)), while silence overlapped
minimally with high loudness (\(0.008\)) and strong acoustic onset (\(0.032\)).
The largest overlaps included sibilants with fricatives (\(0.596\)), vowels
with high loudness (\(0.556\)), and surprisal with word frequency (\(0.511\)).
The phonetic hierarchy also creates asymmetric containment not fully conveyed
by Jaccard overlap: all sibilant frames are contained within the fricative
mask, while sibilants account for 59.6\% of the fricative mask; all schwa
frames are contained within the vowel mask, while schwa accounts for only
9.7\% of the vowel mask.

\section{Validation performance across the architecture sweep}
\label{app:architecture-validation}

\begin{figure}[!ht]
    \centering
    \includegraphics[width=\linewidth]{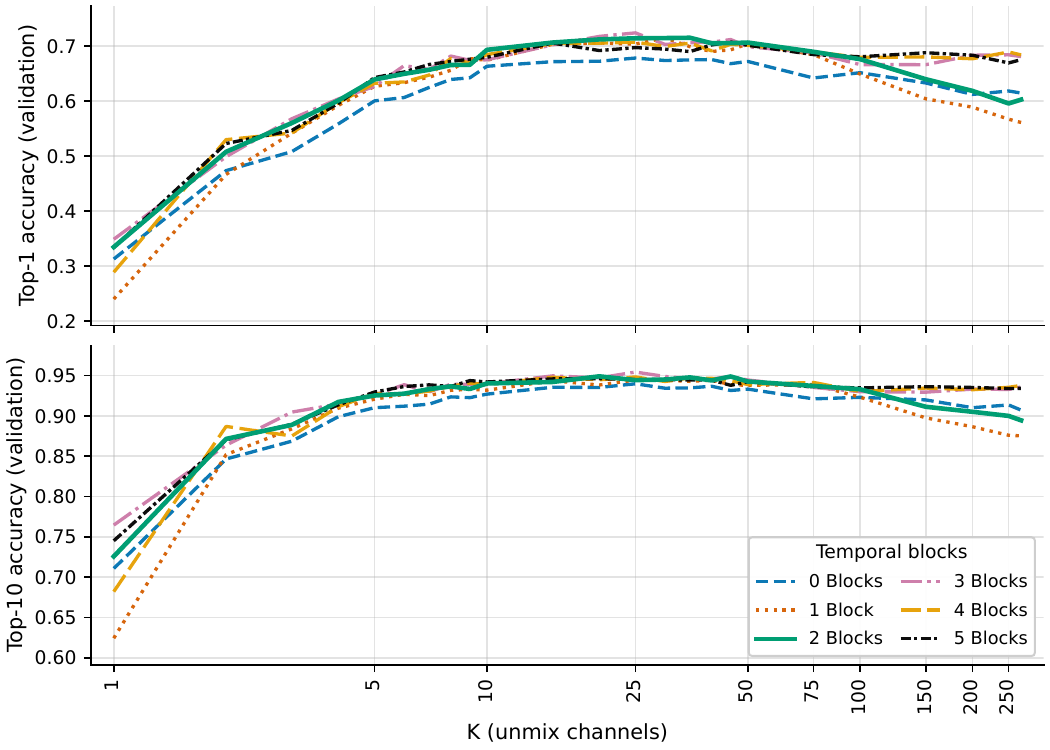}
    \caption{Validation retrieval accuracy across the architecture
    sweep. Retrieval is evaluated against 108 candidate audio segments from
    the held-out validation piece, compared with 1005 candidates in the
    final-test evaluation; absolute validation and final-test accuracies are
    therefore not directly comparable. Top-1 and Top-10 validation accuracy
    are shown as functions of the number of interpretable branches \(K\) and
    convolutional decoder blocks \(B\), using the checkpoint with the lowest
    validation loss for each configuration.}
    \label{fig:k_conv_blocks-val}
\end{figure}

Figure~\ref{fig:k_conv_blocks-val} shows the validation landscape.
Validation retrieval is evaluated against 108 candidate audio segments from
the held-out validation piece, compared with 1005 candidates in the final-test
evaluation. The absolute validation and final-test accuracies are therefore
not directly comparable while the saturation trends are comparable. Top-1 and Top-10 validation accuracy rise sharply
from strongly bottlenecked branch spaces and then form a broad high-performing
regime at intermediate \(K\). Within this regime, additional decoder depth
beyond a shallow nonlinear decoder provides little systematic benefit. The
\(K=25\), \(B=2\) reference configuration lies within this broad regime.

\section{Model size and retrieval performance}
\label{app:model-size}

Table~\ref{tab:model-size} reports trainable parameter counts of the complete 
MEG decoder for
representative points in the architecture and feature-compression
experiments. The frozen wav2vec feature extractor and
the single learned contrastive-temperature scalar are excluded. 
For the LinearDR-12 configuration, the table does not include the trainable 
audio-side projection for consistency. 
This projection contains 9,228 additional parameters which are not a part 
of the MEG decoder. All LISA
rows use the seed-42 checkpoint and the same 1005-candidate final-test
evaluation, and are therefore directly comparable.

\begin{table}[H]
\centering
\small
\setlength{\tabcolsep}{4pt}
\caption{
Trainable model size and retrieval performance for representative
configurations. $F$ denotes the dimensionality of the target audio
representation. The $K=270$, five-block LISA model is the closest tested configuration to D{\'e}fossez et al.~\cite{Defossez2023} in branch count and decoder depth, but is not a reproduction: it retains our 3D attention, temporal filters, preprocessing, and training procedure. The published D{\'e}fossez et al. scores use a different preprocessing and evaluation protocol (word-aligned test segments and without ocular or cardiac component removal) and are included only to ground the external parameter comparison.
}
\label{tab:model-size}
\begin{tabular}{@{}lrrrrrrr@{}}
\toprule
Model & $K$ & $B$ & $F$ & Parameters & Candidates & Top-1 (\%) & Top-10 (\%) \\
\midrule
LISA, smaller branch space
    & 15 & 2 & 768 & 380,769 & 1005 & 39.51 & 70.23 \\
LISA, LinearDR-12
    & 25 & 2 & 12 & 427,651 & 1005 & 39.95 & 70.54 \\
LISA, no convolutional blocks
    & 25 & 0 & 768 & 471,219 & 1005 & 36.76 & 67.37 \\
\textbf{LISA, main}
    & \textbf{25} & \textbf{2} & \textbf{768} & \textbf{486,619}
    & \textbf{1005} & \textbf{40.01} & \textbf{70.60} \\
LISA, five convolutional blocks
    & 25 & 5 & 768 & 509,719 & 1005 & 39.06 & 69.82 \\
LISA, closest tested capacity
    & 270 & 5 & 768 & 7,210,224 & 1005 & 36.41 & 67.46 \\
\midrule
D{\'e}fossez et al.~\cite{Defossez2023} (no ocular/cardiac removal)
    & 270 & 5 & 1024 & 9,565,054 & 1363 & 41.30 & 70.70 \\
\bottomrule
\end{tabular}
\end{table}

The main LISA decoder contains 486,619 trainable parameters: 426,715 in
the spatial--temporal and nonlinear decoder core and 59,904 in the final
projection to the wav2vec target. For the brain decoder of D{\'e}fossez et
al.~\cite{Defossez2023}, we count 9,565,054 trainable parameters for the
27-participant MEG-MASC setting. This count includes its 2D Fourier spatial
attention, shared $1\times1$ projection, participant-specific
$270\times270$ matrices, five convolutional blocks, and final projection to
the 1024-dimensional XLSR-53 representation. The resulting ratio is 19.66,
reported in the main text as approximately $20\times$.

The controlled LISA comparison leads to the same conclusion independently
of differences between published evaluation protocols. The $K=270$,
five-block model contains 7,210,224 parameters---14.8 times more than the
main model---but performs 3.60 percentage points worse in Top-1 and 3.14
points worse in Top-10 accuracy. Conversely, the $K=15$ model retains nearly 
the full retrieval score with 380,769 parameters, while the learned 12-dimensional 
target retains the main performance with a 427,651-parameter MEG decoder and 
a separate 9,228-parameter audio-side projection.
The main result therefore does not arise from increased generic model
capacity: the high-performing regime is reached by a compact branch space
and a shallow nonlinear decoder.

\end{document}